\documentclass{article}

\usepackage[numbers]{natbib}
 \usepackage[final]{nips_2017}
\usepackage{times}
\usepackage{graphicx} % more modern
\usepackage{subcaption}

\usepackage{algorithm}
\usepackage{algorithmic}
\usepackage{diagbox}

\usepackage{hyperref}

\usepackage{enumitem}

\usepackage{Definitions}
\graphicspath{{./figs/}}
\usepackage{color}

\usepackage{mathtools}

\usepackage{booktabs}

\usepackage{wrapfig}
\usepackage{lipsum}

\newcounter{inlineenum}
\renewcommand{\theinlineenum}{\alph{inlineenum}}
\newenvironment{inlineenum}
  {\unskip\ignorespaces\setcounter{inlineenum}{0}%
   \renewcommand{\item}{\refstepcounter{inlineenum}{\textit{\theinlineenum})~}}}
  {\ignorespacesafterend}
  
\makeatletter
\newcommand\footnoteref[1]{\protected@xdef\@thefnmark{\ref{#1}}\@footnotemark}
\makeatother

\usepackage[utf8]{inputenc} % allow utf-8 input
\usepackage[T1]{fontenc}    % use 8-bit T1 fonts
\usepackage{hyperref}       % hyperlinks
\usepackage{url}            % simple URL typesetting
\usepackage{booktabs}       % professional-quality tables
\usepackage{amsfonts}       % blackboard math symbols
\usepackage{nicefrac}       % compact symbols for 1/2, etc.
\usepackage{microtype}      % microtypography

\usepackage{chngcntr}

\newcommand{\ssymbol}[1]{^{\@fnsymbol{#1}}}

\title{Learning Combinatorial Optimization Algorithms over Graphs}

	\author{Hanjun Dai$^{\dagger}$$^*$, Elias B. Khalil$^{\dagger}$\thanks{Both authors contributed equally to the paper.}~, Yuyu Zhang$^{\dagger}$, Bistra Dilkina$^{\dagger}$, Le Song$^{\dagger \mathsection}$ \\
		$^{\dagger}$ College of Computing, Georgia Institute of Technology \\
		$^{\mathsection}$ Ant Financial \\
		\{hanjun.dai, elias.khalil, yuyu.zhang, bdilkina, lsong\}@cc.gatech.edu}
		
\begin{document}

\maketitle

\begin{abstract}
		The design of good heuristics or approximation algorithms for NP-hard combinatorial optimization problems often requires significant specialized knowledge and trial-and-error.
		Can we automate this challenging, tedious process, and learn the algorithms instead? In many real-world applications, it is typically the case that the same optimization problem is solved again and again on a regular basis, maintaining the same problem structure but differing in the data. This provides an opportunity for learning heuristic algorithms that exploit the structure of such recurring problems. 
		In this paper, we propose a unique combination of reinforcement learning and graph embedding to address this challenge. The learned greedy policy behaves like a meta-algorithm that incrementally constructs a solution, and the action is determined by the output of a graph embedding network capturing the current state of the solution. We show that our framework can be applied to a diverse range of optimization problems over graphs, and learns effective algorithms for the Minimum Vertex Cover, Maximum Cut and Traveling Salesman problems.
\end{abstract}

	\vspace{-5mm}
	\section{Introduction}
	\label{sec:introduction}
	\vspace{-3mm}
	
        \setlength{\abovedisplayskip}{4pt}
        \setlength{\abovedisplayshortskip}{1pt}
        \setlength{\belowdisplayskip}{4pt}
        \setlength{\belowdisplayshortskip}{1pt}
        \setlength{\jot}{3pt}
        \setlength{\textfloatsep}{6pt}	
		
	Combinatorial optimization problems over graphs arising from numerous application domains, such as social networks, transportation, telecommunications and scheduling, are NP-hard, and have thus attracted considerable interest from the theory and algorithm design communities over the years. 
 	In fact, of Karp's 21 problems in the seminal paper on reducibility~\citep{Karp72}, 10 are decision versions of graph optimization problems, while most of the other 11 problems, such as set covering, can be naturally formulated on graphs. 	
	Traditional approaches to tackling an NP-hard graph optimization problem have three main flavors: exact algorithms, approximation algorithms and heuristics. Exact algorithms are based on enumeration or branch-and-bound with an integer programming formulation, but may be prohibitive for large instances. 
	On the other hand, polynomial-time approximation algorithms are desirable, but may suffer from weak optimality guarantees or empirical performance, or may not even exist for inapproximable problems. Heuristics are often fast, effective algorithms that lack theoretical guarantees, and may also require substantial problem-specific research and trial-and-error on the part of algorithm designers. 
	
	All three paradigms seldom exploit a common trait of real-world optimization problems: 
	instances of the same type of problem are solved again and again on a regular
	basis, maintaining the same combinatorial structure, but differing mainly in their data. That is, in many applications, values of the coefficients in the objective function or constraints can be thought of as being sampled from the same underlying distribution. For instance, an advertiser on a social network targets a limited set of users with ads, in the hope that they spread them to their neighbors; such covering instances need to be solved repeatedly, since the influence pattern between neighbors may be different each time.  Alternatively, a package delivery company routes trucks on a daily basis in a given city; thousands of similar optimizations need to be solved, since the underlying demand locations can differ.
	
	\begin{figure*}[t!]
	\centering
	\includegraphics[width=0.95\textwidth,page=3]{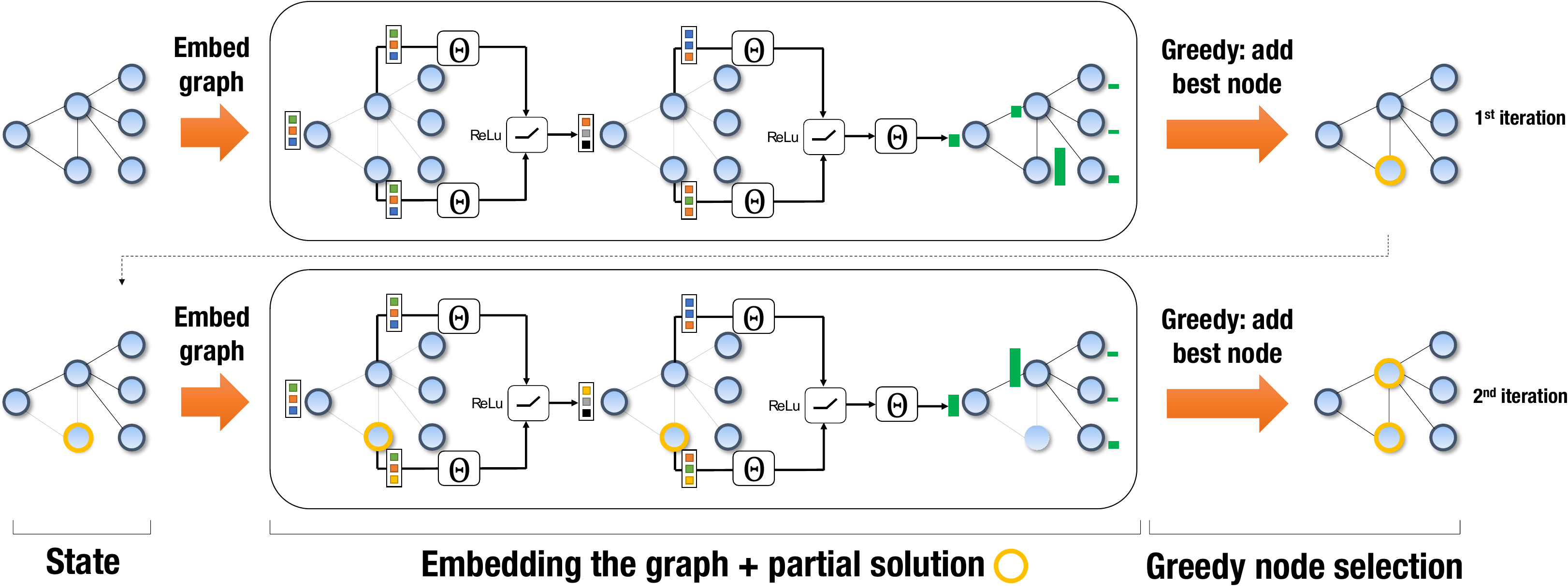}
	\vspace{-2mm}
	\caption{\small Illustration of the proposed framework as applied to an instance of Minimum Vertex Cover. The middle part illustrates two iterations of the graph embedding, which results in node scores (green bars).}
	\label{fig:framework}
	\vspace{-1mm}
\end{figure*}	

	Despite the inherent similarity between problem instances arising in the same domain, classical algorithms do not systematically exploit this fact. However, in industrial settings, a company may be willing to invest in upfront, offline computation and learning if such a process can speed up its real-time decision-making and improve its quality. This motivates the main problem we address:%[-2mm]
	
			\advance\leftmargini -1.8em
			\begin{quotation}
				\vspace{-2mm}
				\noindent {\bf Problem Statement:} Given a graph optimization problem $G$ and a distribution $\mathbb{D}$ of problem instances, can we learn better heuristics that generalize to unseen instances from $\mathbb{D}$?
 				\vspace{-2mm}	
			\end{quotation}
	
	\noindent Recently, there has been some seminal work on using deep architectures to learn heuristics for combinatorial problems, including the Traveling Salesman Problem ~\citep{VinForJai15,BelPhaLeNoretal16,GraWayReyHaretal16}. However, the architectures used in these works are generic, not yet effectively reflecting the combinatorial structure of graph problems. As we show later, these architectures often require a huge number of instances in order to learn to generalize to new ones. Furthermore, existing works typically use the policy gradient for training~\citep{BelPhaLeNoretal16}, a method that is not particularly sample-efficient. While the methods in~\citep{VinForJai15,BelPhaLeNoretal16} can be used on graphs with different sizes -- a desirable trait -- they require manual, ad-hoc input/output engineering to do so  (e.g. padding with zeros). 
	
	In this paper, we address the challenge of learning algorithms for graph problems using a unique combination of reinforcement learning and graph embedding. The learned policy behaves like a meta-algorithm that incrementally constructs a solution, with the action being determined by a graph embedding network over the current state of the solution. More specifically, our proposed solution framework is different from previous work in the following aspects: 

		{\bf 1. Algorithm design pattern.} We will adopt a \textit{greedy} meta-algorithm design, whereby a feasible solution is constructed by successive addition of nodes based on the graph structure, and is maintained so as to satisfy the problem's graph constraints. Greedy algorithms are a popular pattern for designing approximation and heuristic algorithms for graph problems. As such, the same high-level design can be seamlessly used for different graph optimization problems.
		
		{\bf 2. Algorithm representation.} We will use a \textit{graph embedding} network, called \texttt{structure2vec} (S2V)~\cite{DaiDaiSon16}, to represent the policy in the greedy algorithm. 
		This novel deep learning architecture over the instance graph ``featurizes'' the nodes in the graph, capturing the properties of a node in the context of its graph neighborhood. This allows the policy to discriminate among nodes based on their usefulness, and generalizes to problem instances of different sizes. This contrasts with recent approaches~\citep{VinForJai15,BelPhaLeNoretal16} that adopt a graph-agnostic sequence-to-sequence mapping that does not fully exploit graph structure.
		
		{\bf 3. Algorithm training.} We will use fitted $Q$-learning to learn a greedy policy that is parametrized by the graph embedding network. The framework is set up in such a way that the policy will aim to optimize the objective function of the original problem instance~\textit{directly}. The main advantage of this approach is that it can deal with delayed rewards, which here represent the remaining increase in objective function value obtained by the greedy algorithm, in a data-efficient way; in each step of the greedy algorithm, the graph embeddings are updated according to the partial solution to reflect new knowledge of the benefit of \textit{each node} to the final objective value. In contrast, the policy gradient approach of~\citep{BelPhaLeNoretal16} updates the model parameters only once w.r.t. the whole solution (e.g. the tour in TSP).

	The application of a greedy heuristic learned with our framework is illustrated in Figure~\ref{fig:framework}. 
	To demonstrate the effectiveness of the proposed framework, we apply it to three extensively studied graph optimization problems. Experimental results show that our framework, a single meta-learning algorithm, efficiently learns effective heuristics for each of the problems. Furthermore, we show that our learned heuristics preserve their effectiveness even when used on graphs much larger than the ones they were trained on. Since many combinatorial optimization problems, such as the set covering problem, can be explicitly or implicitly formulated on graphs, we believe that our work opens up a new avenue for graph algorithm design and discovery with deep learning.  

	\vspace{-3mm}
	\section{Common Formulation for Greedy Algorithms on Graphs} 
	\label{sec:proposed_framework}
	\vspace{-3mm}
	
        We will illustrate our framework using three optimization problems over weighted graphs. Let ${G}({V},{E}, {w})$ denote a weighted graph, where ${V}$ is the set of nodes, ${E}$ the set of edges and ${w: E\rightarrow \mathbb{R}^{+}}$ the edge weight function, i.e. $w(u,v)$ is the weight of edge $(u,v) \in E$. These problems are:
        \begin{itemize}[leftmargin=*,nosep,nolistsep]
          \item \textbf{Minimum Vertex Cover (MVC):} Given a graph $G$, find a subset of nodes $S\subseteq V$ such that every edge is covered, i.e. $(u,v)\in E \Leftrightarrow u\in S \text{ or } v\in S$, and $|S|$ is minimized. 
          \item \textbf{Maximum Cut (MAXCUT):} Given a graph $G$, find a subset of nodes $S\subseteq V$ such that the weight of the cut-set $\sum_{(u,v)\in C}{w(u,v)}$ is maximized, where cut-set $C\subseteq E$ is the set of edges with one end in $S$ and the other end in $V\setminus S$.
          \item \textbf{Traveling Salesman Problem (TSP):} Given a set of points in 2-dimensional space, find a tour of minimum total weight, where the corresponding graph $G$ has the points as nodes and is fully connected with edge weights corresponding to distances between points; a tour is a cycle that visits each node of the graph \textit{exactly} once.
        \end{itemize}
	
        We will focus on a popular pattern for designing approximation and heuristic algorithms,~namely a greedy algorithm. A greedy algorithm will construct a solution by sequentially adding nodes to a partial solution $S$, based on maximizing some \textit{evaluation function} $Q$ that measures the quality of a node in the context of the current partial solution. 
        We will show that, despite the diversity of the combinatorial problems above, greedy algorithms for them can be expressed using a common formulation. Specifically: 
	\begin{enumerate}[leftmargin=*,nosep,nolistsep]
		\item A problem instance ${G}$ of a given optimization problem is sampled from a distribution $\mathbb{D}$, i.e. the $V$, $E$ and $w$ of the instance graph $G$ are generated according to a model or real-world data. 
		
                \item A partial solution is represented as an ordered list ${S} = (v_1,v_2,\dots,v_{|{S}|})$, $v_i \in {V}$, and $\overline S = {V} \setminus S$ the set of candidate nodes for addition, conditional on ${S}$. Furthermore, we use a vector of binary decision variables $x$, with each dimension $x_v$ corresponding to a node $v\in V$, 
                $x_v=1$ if $v \in S$ and 0 otherwise. One can also view $x_v$ as a tag or extra feature on $v$. 	
		\item A maintenance (or helper) procedure $h(S)$ will be needed, which maps an ordered list $S$ to a combinatorial structure satisfying the specific constraints of a problem.  
		\item The quality of a partial solution $S$ is given by an objective function $c(h(S), {G})$ based on the combinatorial structure $h$ of $S$.   		
                \item A generic greedy algorithm selects a node $v$ to add next such that $v$ maximizes an evaluation function, $Q(h(S),v) \in \mathbb{R}$, which depends on the combinatorial structure $h(S)$ of the current partial solution. Then, the partial solution $S$ will be extended as 
                \begin{align}
                  {S} \coloneqq ({S}, v^*),~~\text{where}~~v^*\coloneqq\argmax\nolimits_{v\in \overline S}~{Q(h({S}),v)},
                  \label{eq:q_argmax}
                \end{align}
                and $({S}, v^*)$ denotes appending $v^*$ to the end of a list ${S}$. This step is repeated until a termination criterion $t(h(S))$ is satisfied.
	\end{enumerate}
        In our formulation, we assume that the distribution $\mathbb{D}$, the helper function $h$, the termination criterion $t$ and the cost function $c$ are all given. Given the above abstract model, various optimization problems can be expressed by using different helper functions, cost functions and termination criteria: 
	\begin{itemize}[leftmargin=*,nosep,nolistsep]
		\item {\bf MVC:} The helper function does not need to do any work, and $c(h(S),{G}) = -\abr{S}$. The termination criterion checks whether all edges have been covered. 
		\item {\bf MAXCUT:} The helper function divides $V$ into two sets, $S$ and its complement $\overline S = V\setminus S$, and maintains a cut-set $C=\{(u,v) \,|\, (u,v) \in E, u \in S, v \in \overline S \}$. Then, the cost is ${c(h(S), {G}) = \sum_{(u,v) \in C} w(u,v)}$, and the termination criterion does nothing. 
		\item {\bf TSP:} The helper function will maintain a tour according to the order of the nodes in $S$. The simplest way is to append nodes to the end of partial tour in the same order as $S$. Then the cost $c(h(S), {G})=-\sum_{i =1}^{|S|-1} w(S(i), S(i+1)) - w(S(|S|),S(1))$, and the termination criterion is activated when $S=V$. 
		Empirically, inserting a node $u$ in the partial tour at the position which increases the tour length the least is a better choice. We adopt this insertion procedure as a helper function for TSP.

	\end{itemize}
	An estimate of the quality of the solution resulting from adding a node to partial solution $S$ will be determined by the \textit{evaluation function} $Q$, which will be learned using a collection of problem instances. This is in contrast with traditional greedy algorithm design, where the \textit{evaluation function} $Q$ is typically hand-crafted, and requires substantial problem-specific research and trial-and-error. In the following, we will design a powerful deep learning parameterization for the evaluation function, $\widehat{Q}(h({S}), v; \Theta)$, with parameters $\Theta$.
	
        \vspace{-3mm}
	\section{Representation: Graph Embedding}
	\label{sec:representation} 
	\vspace{-3mm}
	
	Since we are optimizing over a graph $G$, we expect that the evaluation function $\widehat{Q}$ should take into account the current partial solution $S$ as it maps to the graph. That is, $x_v=1$ for all nodes $v \in S$, and the nodes are connected according to the graph structure. Intuitively, $\widehat{Q}$ should summarize the state of such a ``tagged" graph $G$, and figure out the value of a new node if it is to be added in the context of such a graph. Here, both the state of the graph and the context of a node $v$ can be very complex, hard to describe in closed form, and may depend on complicated statistics such as global/local degree distribution, triangle counts, distance to tagged nodes, etc. In order to represent such complex phenomena over combinatorial structures, we will leverage a deep learning architecture over graphs, in particular the \texttt{structure2vec} of~\cite{DaiDaiSon16}, to parameterize $\widehat{Q}(h({S}), v; \Theta)$. 
	
	\vspace{-3mm}
	\subsection{Structure2Vec} 
	\label{sec:structure2vec}
	\vspace{-2mm}
	
	We first provide an introduction to \texttt{structure2vec}. This graph embedding network will compute a $p$-dimensional feature embedding $\mu_v$ for each node $v\in {V}$, given the current partial solution $S$. More specifically, \texttt{structure2vec} defines the network architecture recursively according to an input graph structure $G$, and the computation graph of \texttt{structure2vec} is inspired by graphical model inference algorithms, where node-specific tags or features $x_v$ are aggregated recursively according to ${G}$'s graph topology. After a few steps of recursion, the network will produce a new embedding for each node, taking into account both graph characteristics and long-range interactions between these node features. One variant of the \texttt{structure2vec} architecture will initialize the embedding $\mu_v^{(0)}$ at each node as $0$, and for all $v \in {V}$ update the embeddings synchronously at each iteration as
	\begin{align}
	\mu_v^{(t+1)} \leftarrow F \rbr{x_v, \{\mu_u^{(t)}\}_{u\in\Ncal(v)}, \cbr{w(v,u)}_{u\in\Ncal(v)}; \Theta},
	\label{eq:fix-point}
	\end{align}
	where $\Ncal(v)$ is the set of neighbors of node $v$ in graph ${G}$, and $F$ is a generic nonlinear mapping such as a neural network or kernel function. 
	
	Based on the update formula, one can see that the embedding update process is carried out based on the graph topology. A new round of embedding sweeping across the nodes will start only after the embedding update for all nodes from the previous round has finished. It is easy to see that the update also defines a process where the node features $x_v$ are propagated to other nodes via the nonlinear propagation function $F$. Furthermore, the more update iterations one carries out, the farther away the node features will propagate and get aggregated nonlinearly at distant nodes. In the end, if one terminates after $T$ iterations, each node embedding $\mu_v^{(T)}$ will contain information about its $T$-hop neighborhood as determined by graph topology, the involved node features and the propagation function $F$. An illustration of two iterations of graph embedding can be found in Figure~\ref{fig:framework}. 
	
	\vspace{-3mm}
	\subsection{Parameterizing $\widehat{Q}(h({S}), v; \Theta)$}
	\label{sec:parametrize_q}
	\vspace{-3mm}

	We now discuss the parameterization of $\widehat{Q}(h({S}), v; \Theta)$ using the embeddings from \texttt{structure2vec}. In particular, we design $F$ to update a $p$-dimensional embedding $\mu_v$ as:
	\begin{align}
	  \mu_v^{(t+1)} \leftarrow \mathrm{relu}\big(\theta_1 x_v + \theta_2 \sum\nolimits_{u\in\Ncal(v)} \mu_u^{(t)} + \theta_3 \sum\nolimits_{u \in \Ncal(v)} \mathrm{relu}(\theta_4\, w(v,u))\big), 
	  \label{eq:actual_f} 
	\end{align}
	where $\theta_1 \in \RR^p$, $\theta_2,\theta_3 \in \RR^{p\times p}$ and $\theta_4 \in \RR^p$ are the model parameters, and $\mathrm{relu}$ is the rectified linear unit ({$\mathrm{relu}(z) = \max(0, z)$}) applied elementwise to its input. The summation over neighbors is one way of aggregating neighborhood information that is invariant to permutations over neighbors. For simplicity of exposition, $x_v$ here is a binary scalar as described earlier; it is straightforward to extend $x_v$ to a vector representation by incorporating any additional useful node information.
	To make the nonlinear transformations more powerful, we can add some more layers of $\mathrm{relu}$ before we pool over the neighboring embeddings $\mu_u$. 
		
	Once the embedding for each node is computed after $T$ iterations, we will use these embeddings to define the $\widehat{Q}(h({S}), v; \Theta)$ function. More specifically, we will use the embedding $\mu_v^{(T)}$ for node $v$ and the pooled embedding over the entire graph, $\sum_{u \in {V}} \mu_{u}^{(T)}$, as the surrogates for $v$ and $h({S})$, respectively, i.e.
	\begin{align}
	  {\widehat{Q}(h({S}), v; \Theta) = \theta_5^\top \, \mathrm{relu}([\theta_6 \sum\nolimits_{u \in {V}} \mu_{u}^{(T)}, \theta_7\, \mu_v^{(T)}])}
	\end{align}
	where $\theta_5 \in \RR^{2p}$, $\theta_6,\theta_7 \in \RR^{p\times p}$ and $[\cdot,\cdot]$ is the concatenation operator. Since the embedding $\mu_u^{(T)}$ is computed based on the parameters from the graph embedding network, $\widehat{Q}(h({S}), v)$ will depend on a collection of 7 parameters $\Theta = \{\theta_i\}_{i=1}^{7}$. The number of iterations $T$ for the graph embedding computation is usually small, such as $T=4$.
	
	The parameters $\Theta$ will be learned. Previously, \cite{DaiDaiSon16} required a ground truth label for every input graph ${G}$ in order to train the \texttt{structure2vec} architecture. There, the output of the embedding is linked with a softmax-layer, so that the parameters can by trained end-to-end by minimizing the cross-entropy loss. This approach is not applicable to our case due to the lack of training labels. Instead, we train these parameters together \textit{end-to-end} using reinforcement learning.
	
	\vspace{-3mm}
	\section{Training: Q-learning} 
	\label{sec:training}
	\vspace{-3mm}
	
	We show how reinforcement learning is a natural framework for learning the evaluation function $\widehat{Q}$. The definition of the evaluation function $\widehat{Q}$ naturally lends itself to a \textit{reinforcement learning} (RL) formulation~\citep{SutBar98}, and we will use $\widehat{Q}$ as a model for the state-value function in RL. We note that we would like to learn a function $\widehat{Q}$ \textit{across a set of $m$ graphs from distribution $\mathbb{D}$}, $\Dcal = \{{G}_i\}_{i=1}^{m}$, with potentially different sizes. The advantage of the graph embedding parameterization in our previous section is that we can deal with different graph instances and sizes seamlessly.  
	
	\vspace{-3mm}
	\subsection{Reinforcement learning formulation}
	\vspace{-2mm}

	We define the states, actions and rewards in the reinforcement learning framework as follows:
	%\vspace{-2mm}
	\begin{enumerate}[leftmargin=*,nosep,nolistsep]
		\item \textit{States}: a state ${S}$ is a sequence of actions (nodes) on a graph ${G}$. Since we have already represented nodes in the tagged graph with their embeddings, the state is a vector in $p$-dimensional space, $\sum_{v\in V} \mu_v$. It is easy to see that this embedding representation of the state can be used across different graphs. 	
		The terminal state $\widehat{S}$ will depend on the problem at hand;
		\item \textit{Transition}: transition is deterministic here, and corresponds to tagging the node $v\in G$ that was selected as the last action with feature $x_v=1$; 
		\item \textit{Actions}: an action $v$ is a node of ${G}$ that is not part of the current state ${S}$. Similarly, we will represent actions as their corresponding $p$-dimensional node embedding $\mu_v$, and such a definition is applicable across graphs of various sizes;	
		\item \textit{Rewards}: the reward function $r(S,v)$ at state $S$ is defined as the change in the cost function after taking action $v$ and transitioning to a new state ${S}'\coloneqq ({S}, v)$. That is, 
		\begin{align}
		r({S}, v) = c(h(S'), G) - c(h(S), G),               
		\end{align}
		and $c(h(\emptyset),G)=0$.
		As such, the \textit{cumulative reward} $R$ of a terminal state $\widehat{S}$ coincides exactly with the objective function value of the $\widehat{S}$, i.e. $R(\widehat{S}) = \sum_{i=1}^{|\widehat{S}|}{r({S}_i,v_i)}$ is equal to $c(h(\widehat{S}), G)$;
		\item \textit{Policy}: based on $\widehat{Q}$, a deterministic greedy policy $\pi(v|{S}):=\argmax_{v' \in \overline S} \widehat{Q}(h({S}),v')$ will be used. 
		Selecting action $v$ corresponds to adding a node of ${G}$ to the current partial solution, which results in collecting a reward $r({S},v)$.
	\end{enumerate}
	Table~\ref{tab:rlopt} shows the instantiations of the reinforcement learning framework for the three optimization problems considered herein.
	We let $Q^{*}$ denote the \textit{optimal} Q-function for each RL problem. Our graph embedding parameterization $\widehat{Q}(h(S),v;\Theta)$ from Section~\ref{sec:representation} will then be a function approximation model for it, which will be learned via $n$-step Q-learning.  
	
	\begin{table*}[t]
%                \vspace{-3mm}
		\centering
		\caption{Definition of reinforcement learning components for each of the three problems considered.}
		\label{tab:rlopt}%
		\vspace{-1mm}
		\resizebox{\textwidth}{!}{%
			\begin{tabular}{l|l|l|l|l|l}
				\toprule
				\textbf{Problem} & \textbf{State} & \textbf{Action} & \textbf{Helper function} & \textbf{Reward} & \textbf{Termination} \\
				\midrule
				MVC & subset of nodes selected so far & add node to subset & None  & -1 & all edges are covered \\
				MAXCUT & subset of nodes selected so far & add node to subset & None  & change in cut weight & cut weight cannot be improved \\
				TSP & partial tour & grow tour by one node & Insertion operation & change in tour cost & tour includes all nodes \\
				\bottomrule
			\end{tabular}%
		}
		\vspace{-3mm}
	\end{table*}%
	
	\vspace{-3mm}
	\subsection{Learning algorithm} 
	\vspace{-3mm}
        	
	In order to perform end-to-end learning of the parameters in $\widehat{Q}(h(S),v;\Theta)$, we use a combination of $n$-step Q-learning~\citep{SutBar98} and \textit{fitted Q-iteration}~\citep{Riedmiller05}, as illustrated in Algorithm~\ref{alg:rl}. We use the term \textit{episode} to refer to a complete sequence of node additions starting from an empty solution, and until termination; a \textit{step} within an episode is a single action (node addition).
	
	Standard (1-step) Q-learning updates the function approximator's parameters \textit{at each step} of an episode by performing a gradient step to minimize the squared loss:
	\begin{align}
	(y - \widehat{Q}(h({S}_t),v_t; \Theta))^2,
	\label{eq:sqloss}
	\end{align}
	where $y = \gamma \max_{v'}{\widehat{Q}(h({S}_{t+1}),v'; \Theta)} + r({S}_t,v_t)$ for a non-terminal state ${S}_t$.
	The $n$-step Q-learning helps deal with the issue of \textit{delayed rewards}, where the final reward of interest to the agent is only received far in the future during an episode. In our setting, the final objective value of a solution is only revealed after many node additions. As such, the 1-step update may be too myopic. A natural extension of 1-step Q-learning is to wait $n$ steps before updating the approximator's parameters, so as to collect a more accurate estimate of the future rewards. Formally, the update is over the same squared loss~\eqref{eq:sqloss}, but with a different target,
$
	y = \sum_{i=0}^{n-1}{r({S}_{t+i},v_{t+i})} + \gamma \max_{v'}{\widehat{Q}(h({S}_{t+n}),v'; \Theta)}.
$
	The fitted Q-iteration approach has been shown to result in faster learning convergence when using a neural network as a function approximator~\citep{Riedmiller05,MniKavSilGraetal13}, a property that also applies in our setting, as we use the embedding defined in Section~\ref{sec:parametrize_q}. Instead of updating the Q-function sample-by-sample as in Equation~\eqref{eq:sqloss}, the fitted Q-iteration approach uses~\textit{experience replay} to update the function approximator with a batch of samples from a dataset ${E}$, rather than the single sample being currently experienced. The dataset ${E}$ is populated during previous episodes, such that at step $t+n$, the tuple $({S}_t,a_t,R_{t,t+n},{S}_{t+n})$ is added to ${E}$, with $R_{t,t+n}=\sum_{i=0}^{n-1}{r({S}_{t+i},a_{t+i})}$. Instead of performing a gradient step in the loss of the current sample as in~\eqref{eq:sqloss}, stochastic gradient descent updates are performed on a random sample of tuples drawn from ${E}$. 
	
	It is known that \textit{off-policy} reinforcement learning algorithms such as Q-learning can be more sample efficient than their policy gradient counterparts~\citep{GuLilGhaTuetal16}. This is largely due to the fact that policy gradient methods require \textit{on-policy} samples for the new policy obtained after each parameter update of the function approximator.	

%        \begin{wrapfigure}{R}{0.55\textwidth}
\begin{minipage}{.96\linewidth}
	\begin{algorithm}[H]
		\caption{\textbf{Q-learning for the Greedy Algorithm}}\label{alg:rl}
		\begin{algorithmic}[1] 
			%               \STATE {\bf Input:} ${G}=( {V}, {E}, {W})$ and $x$. 
			\STATE Initialize experience replay memory $\Mcal$ to capacity $N$
			\FOR{episode $e=1$ {\bfseries to} $L$}
			\STATE Draw graph ${G}$ from distribution $\mathbb{D}$
			\STATE Initialize the state to empty ${S}_1=()$
			\FOR{step $t=1$ {\bfseries to} $T$}
			\STATE $v_t = \begin{cases} \text{random node } v\in \overline S_t, \qquad\text{w.p. }\epsilon \\ \argmax_{v\in \overline S_t}{\widehat{Q}(h({S}_{t}),v; \Theta)}  , \text{otherwise}  \end{cases}$
			\STATE Add $v_t$ to partial solution: ${S}_{t+1}\coloneqq ({S}_t,v_t)$
			\IF{$t\geq n$}
			\STATE Add tuple $({S}_{t-n},v_{t-n},R_{t-n,t},{S}_{t})$ to $\Mcal$
			\STATE Sample random batch from $B \overset{iid.}{\sim} \Mcal$
			\STATE Update $\Theta$ by SGD over~\eqref{eq:sqloss} for $B$
			\ENDIF
			\ENDFOR 
			\ENDFOR
			\STATE return $\Theta$
		\end{algorithmic}
	\end{algorithm}
\end{minipage}	
	\vspace{-3mm}
	\section{Experimental Evaluation}
	\vspace{-3mm}

	{\bf Instance generation.} To evaluate the proposed method against other approximation/heuristic algorithms and deep learning approaches, we generate graph instances for each of the three problems. For the MVC and MAXCUT problems, we generate Erd\H{o}s-Renyi (ER)~\citep{ErdRen60} and Barabasi-Albert (BA)~\citep{AlbertBarabasi02} graphs which have been used to model many real-world networks. For a given range on the number of nodes, e.g. 50-100, we first sample the number of nodes uniformly at random from that range, then generate a graph according to either ER or BA. For the two-dimensional TSP problem, we use an instance generator from the DIMACS TSP Challenge~\cite{JohnsonMcgeoch07} to generate uniformly random or clustered points in the 2-D grid. 
	We refer the reader to the Appendix~\ref{app:instances} for complete details on instance generation. 
	We have also tackled the Set Covering Problem, for which the description and results are deferred to Appendix~\ref{app:scp}. 
	
	{\bf Structure2Vec Deep Q-learning. } For our method, S2V-DQN, we use the graph representations and hyperparameters described in Appendix~\ref{app:s2v_config}. The hyperparameters are selected via preliminary results on small graphs, and then fixed for large ones. Note that for TSP, where the graph is fully-connected, we build the $K$-nearest neighbor graph ($K=10$) to scale up to large graphs. For MVC, where we train the model on graphs with up to 500 nodes, we use the model trained on small graphs as initialization for training on larger ones. We refer to this trick as ``pre-training", which is illustrated in Figure~\ref{fig:convergence}. 
	
	{\bf Pointer Networks with Actor-Critic.}
	We compare our method to a method, based on Recurrent Neural Networks (RNNs), which does not make full use of graph structure~\cite{BelPhaLeNoretal16}. 
	We implement and train their algorithm (PN-AC) for all three problems. The original model only works on the Euclidian TSP problem, where each node is represented by its $(x, y)$ coordinates, and is not designed for problems with graph structure. To handle other graph problems, we describe each node by its adjacency vector instead of coordinates. To handle different graph sizes, we use a singular value decomposition (SVD) to obtain a rank-8 approximation for the adjacency matrix, and use the low-rank embeddings as inputs to the pointer network.
	
	{\bf Baseline Algorithms.} Besides the PN-AC, we also include powerful approximation or heuristic algorithms from the literature. These algorithms are specifically designed for each type of problem: 
	%\vspace{-2mm}	
	\begin{itemize}[leftmargin=*,nosep,nolistsep]
		\item \textbf{MVC:} \emph{MVCApprox} iteratively selects an uncovered edge and adds both of its endpoints~\cite{PapSte82}. We designed a stronger variant, called \emph{MVCApprox-Greedy}, that greedily picks the uncovered edge with maximum sum of degrees of its endpoints. Both algorithms are 2-approximations. 
		\item \textbf{MAXCUT:} We include \emph{MaxcutApprox}, which maintains the cut set $(S, V \setminus S)$ and moves a node from one side to the other side of the cut if that operation results in cut weight improvement~\cite{KleinbergTardos06}. To make \emph{MaxcutApprox} stronger, we greedily move the node that results in the largest improvement in cut weight. A randomized, non-greedy algorithm, referred to as SDP, is also implemented based on~\cite{GoeWil95}; 100 solutions are generated for each graph, and the best one is taken.
		\item \textbf{TSP:} We include the following approximation algorithms: Minimum Spanning Tree (MST), Farthest insertion (Farthest), Cheapest insertion (Cheapest), Closest insertion (Closest), Christofides and 2-opt. We also add the Nearest Neighbor heuristic (Nearest); see~\cite{AppBixChvCoo11} for algorithmic details.

	\end{itemize}
	
	{\bf Details on Validation and Testing.} For S2V-DQN and PN-AC, we use a CUDA K80-enabled cluster for training and testing. Training convergence for S2V-DQN is discussed in Appendix~\ref{app:convergence}. S2V-DQN and PN-AC use 100 held-out graphs for validation, and we report the test results on another 1000 graphs.  We use CPLEX\cite{CPLEX1261} to get optimal solutions for MVC and MAXCUT, and Concorde~\cite{AppBixChvCoo06} for TSP (details in Appendix~\ref{app:instances}). All approximation ratios reported in the paper are with respect to the best (possibly optimal) solution found by the solvers within 1 hour. 
	For MVC, we vary the training and test graph sizes in the ranges $\{$15--20, 40--50, 50--100, 100--200, 400--500$\}$. For MAXCUT and TSP, which involve edge weights, we train up to 200--300 nodes due to the limited computation resource. For all problems, we test on graphs of size up to {1000--1200}.
	
	During testing, instead of using Active Search as in~\cite{BelPhaLeNoretal16}, we simply use the greedy policy. This gives us much faster inference, while still being powerful enough. 
	We modify existing open-source code to implement both S2V-DQN~\footnote{\tiny{\url{https://github.com/Hanjun-Dai/graphnn}}} and PN-AC~\footnote{\tiny{\url{https://github.com/devsisters/pointer-network-tensorflow}}}. Our code is publicly available~\footnote{\tiny{\url{https://github.com/Hanjun-Dai/graph\_comb\_opt}}}. 
	
	\vspace{-2mm}
	\subsection{Comparison of solution quality} 
	\vspace{-2mm}
	
	To evaluate the solution quality on test instances, we use the \textit{approximation ratio} of each method relative to the optimal solution, averaged over the set of test instances. The approximation ratio of a solution $S$ to a problem instance $G$ is defined as
	$\Rcal(S,G)=\max(  \smallfrac{OPT(G)}{c(h(S))}, \smallfrac{c(h(S))}{OPT(G)} )$, where $c(h(S))$ is the objective value of solution $S$, and $OPT(G)$ is the best-known solution value of instance $G$.
	
\begin{figure*}
%	\vspace{-2mm}
	\hspace{-4mm}	
	\setlength{\tabcolsep}{3pt}
	\begin{tabular}{ccc}
		\includegraphics[width=0.33\textwidth, trim=-40 0 0 0]{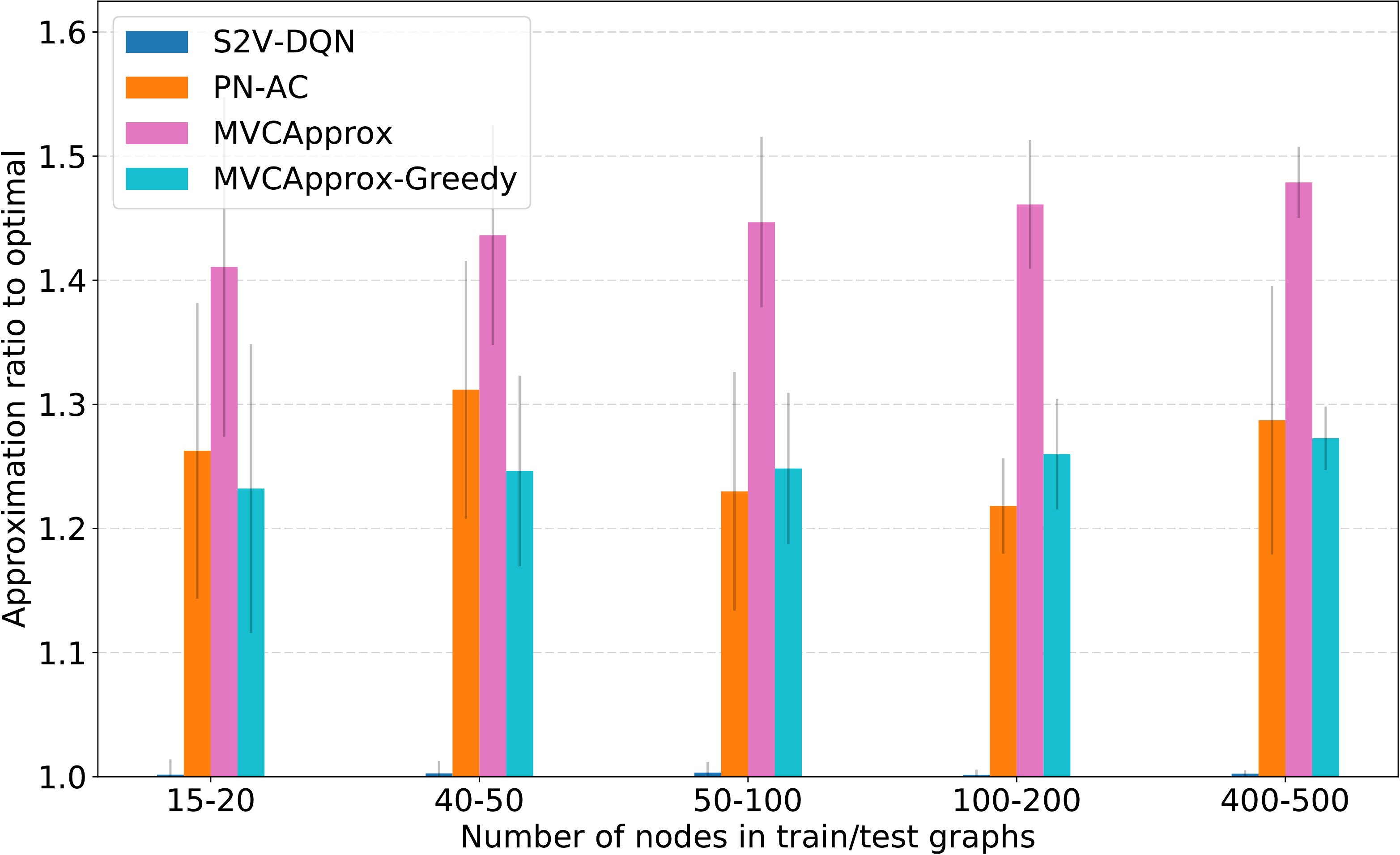} &
		
		\includegraphics[width=0.33\textwidth, trim=-40 0 0 0]{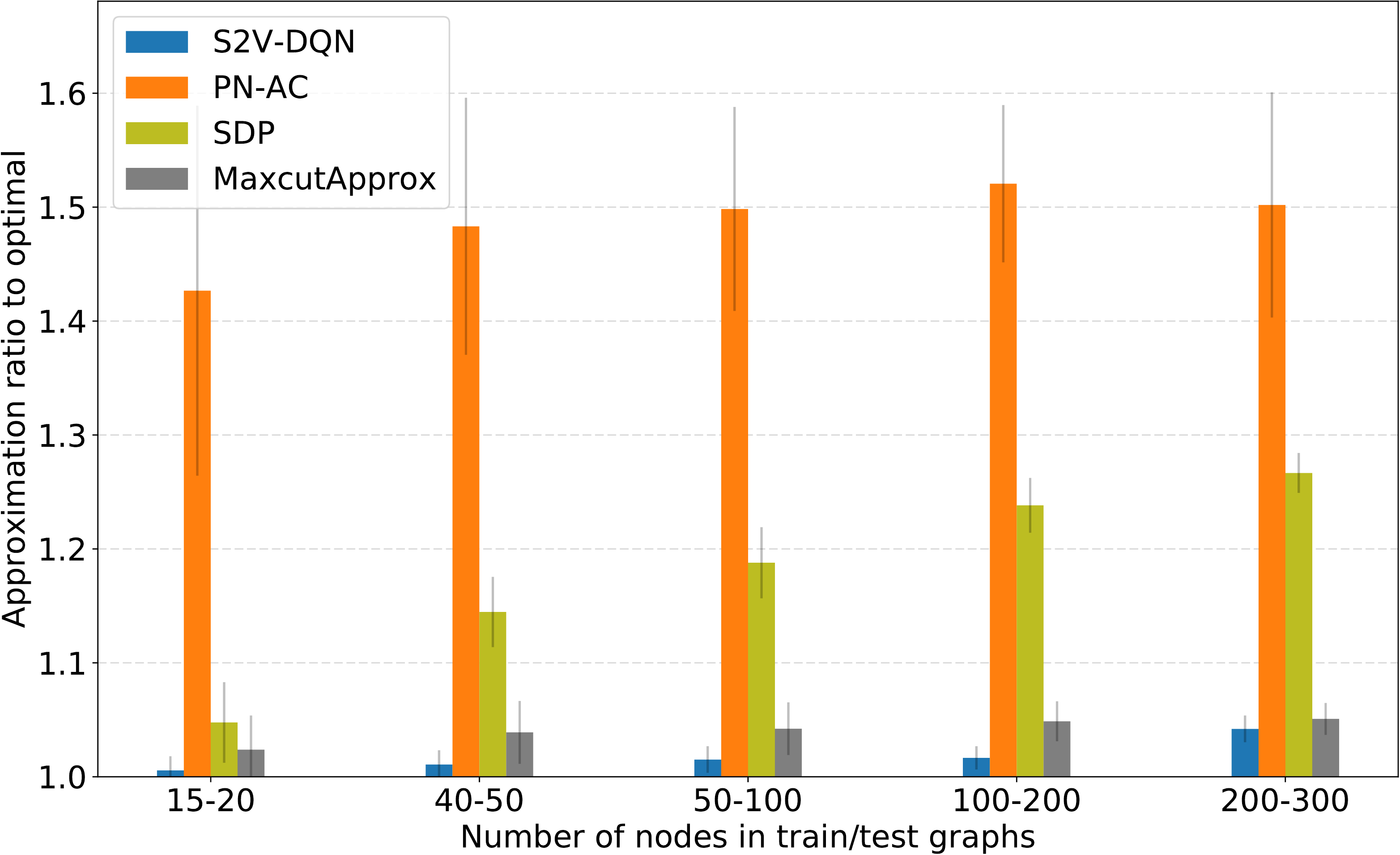} &
		\includegraphics[width=0.33\textwidth, trim=-40 0 0 0]{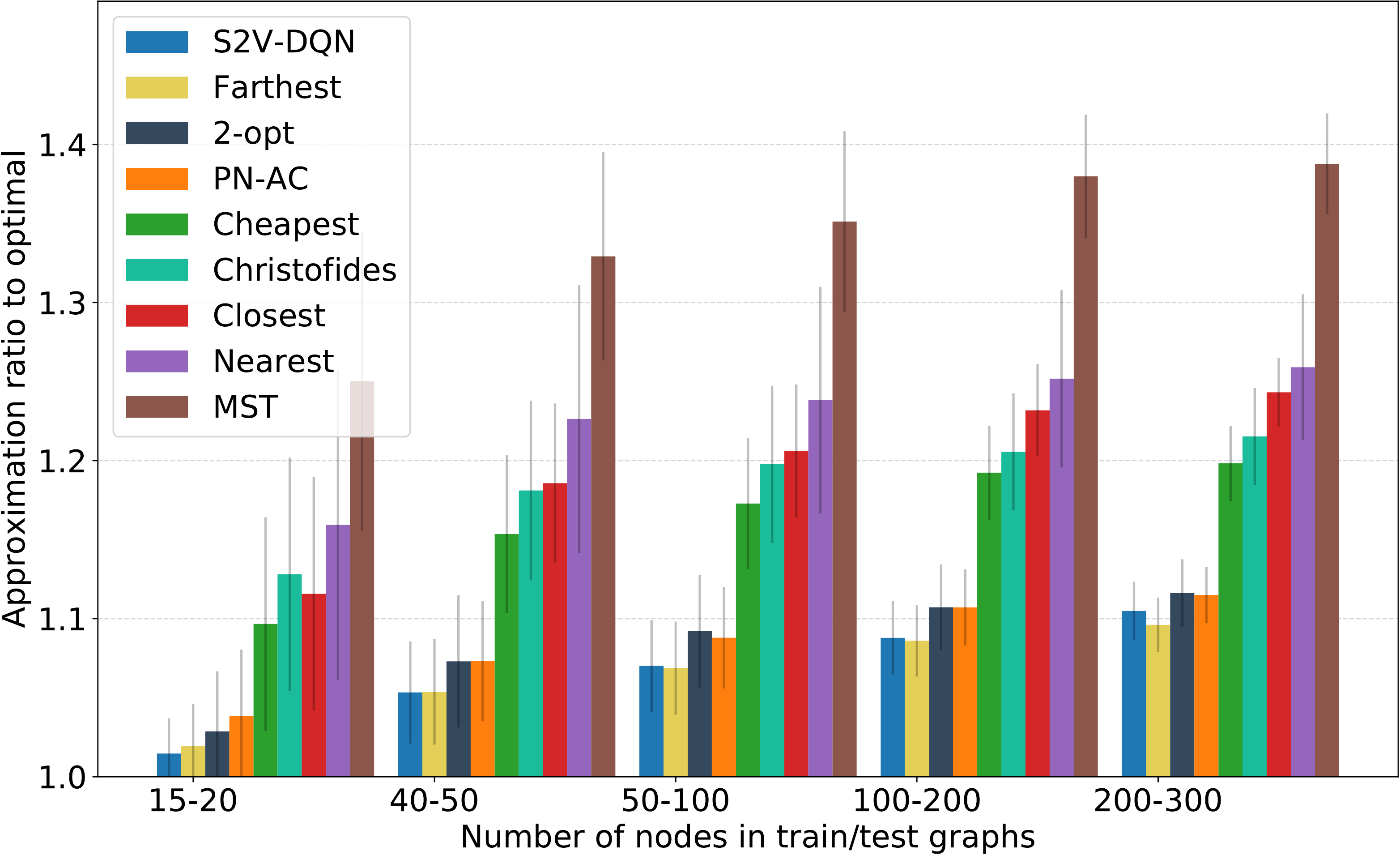} \\
		{\small (a) MVC BA} & {\small (b) MAXCUT BA} & (c) {\small TSP random}
	\end{tabular}   
%	\vspace{-3mm}
	\caption{\small Approximation ratio on 1000 test graphs. Note that on MVC, our performance is pretty close to optimal. In this figure, training and testing graphs are generated according to the same distribution. 
	}
	\label{fig:sol_qual}
	\vspace{-2mm}
\end{figure*}

	Figure~\ref{fig:sol_qual} shows the average approximation ratio across the three problems; other graph types are in Figure~\ref{fig:sol_qual_full} in the appendix. In all of these figures, a lower approximation ratio is better. Overall, our proposed method, S2V-DQN, performs significantly better than other methods. In MVC, the performance of S2V-DQN is particularly good, as the approximation ratio is roughly 1 and the bar is barely visible.
	
    The PN-AC algorithm performs well on TSP, as expected. Since the TSP graph is essentially fully-connected, graph structure is not as important. On problems such as MVC and MAXCUT, where graph information is more crucial, our algorithm performs significantly better than PN-AC. For TSP, The Farthest and 2-opt algorithm perform as well as S2V-DQN, and slightly better in some cases. However, we will show later that in real-world TSP data, our algorithm still performs better.  
    %an intuitive result given the sophistication of this algorithm, which exchanges pairs of edges that can give a smaller tour. 

	\vspace{-2mm}
	\subsection{Generalization to larger instances} 
	\vspace{-2mm}
	
	The graph embedding framework enables us to train and test on graphs of different sizes, since the same set of model parameters are used. How does the performance of the learned algorithm using small graphs generalize to test graphs of larger sizes? 
	To investigate this, we train S2V-DQN on graphs with 50--100 nodes, and test its generalization ability on graphs with up to 1200 nodes. Table~\ref{tab:generalize_sample} summarizes the results, and full results are in Appendix~\ref{app:full_result}. 
	
	\begin{table}[h]
		\vspace{-2mm}
		\centering
		\caption{\small S2V-DQN's generalization ability. Values are average approximation ratios over 1000 test instances. These test results are produced by S2V-DQN algorithms trained on graphs with 50-100 nodes.} 
		\label{tab:generalize_sample}
    \resizebox{\columnwidth}{!}{
		\begin{tabular}{c|c|c|c|c|c|c|c}
			\toprule 
			Test Size& 50-100 & 100-200 & 200-300 & 300-400 & 400-500 & 500-600 & 1000-1200\\
			\hline
			MVC (BA)& 1.0033 & 1.0041 & 1.0045 & 1.0040 & 1.0045 & 1.0048 & 1.0062\\	
			\hline	
			MAXCUT (BA)& 1.0150 & 1.0181 & 1.0202 & 1.0188 & 1.0123 & 1.0177 & 1.0038\\
			\hline
			TSP (clustered)& 1.0730 & 1.0895 & 1.0869 & 1.0918 & 1.0944 & 1.0975 & 1.1065\\
			\bottomrule
		\end{tabular}
	}
		\vspace{-2mm}
	\end{table}

	We can see that S2V-DQN achieves a very good approximation ratio. Note that the ``optimal" value used in the computation of approximation ratios may not be truly optimal (due to the solver time cutoff at 1 hour), and so CPLEX's solutions do typically get worse as problem size grows. This is why sometimes we can even get better approximation ratio on larger graphs.
	
	\vspace{-2mm}
	\subsection{Scalability \& Trade-off between running time and approximation ratio} 
	\vspace{-2mm}
	
	To construct a solution on a test graph, our algorithm has polynomial complexity of $O(k|E|)$ where $k$ is number of greedy steps (at most the number of nodes $|V|$) and $|E|$ is number of edges. 
	For instance, on graphs with 1200 nodes, we can find the solution of MVC within 11 seconds using a single GPU, while getting an approximation ratio of $1.0062$. 
	For dense graphs, we can also sample the edges for the graph embedding computation to save time, a measure we will investigate  in the future. 
	
\begin{figure*}[ht!]
	  \vspace{-2mm}
	  \centering
	  \begin{minipage}[c]{.68\textwidth}
		\setlength{\tabcolsep}{1pt}
		\begin{tabular}{cc}
			\includegraphics[width=0.48\textwidth]{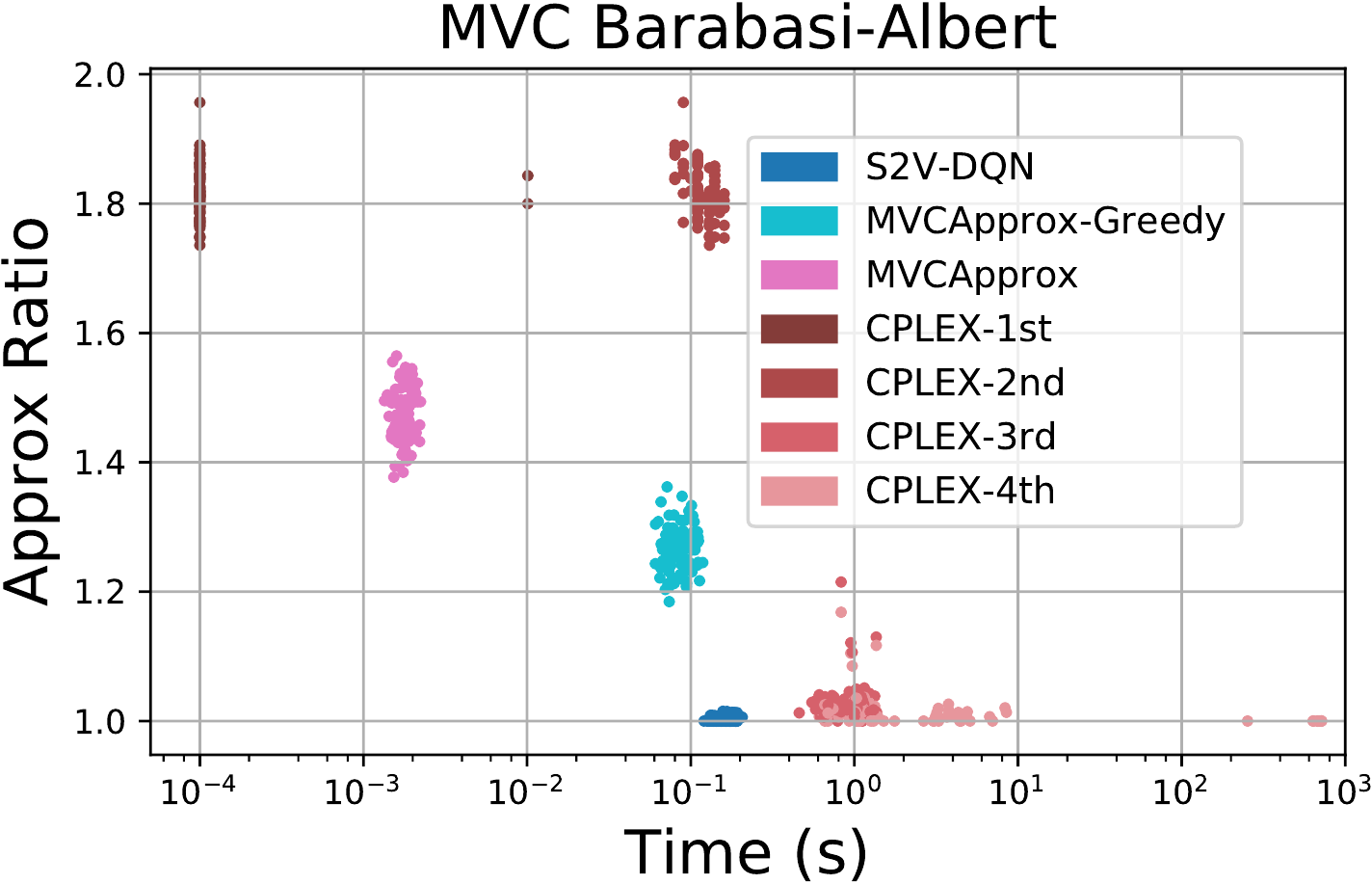} &
			\includegraphics[width=0.48\textwidth]{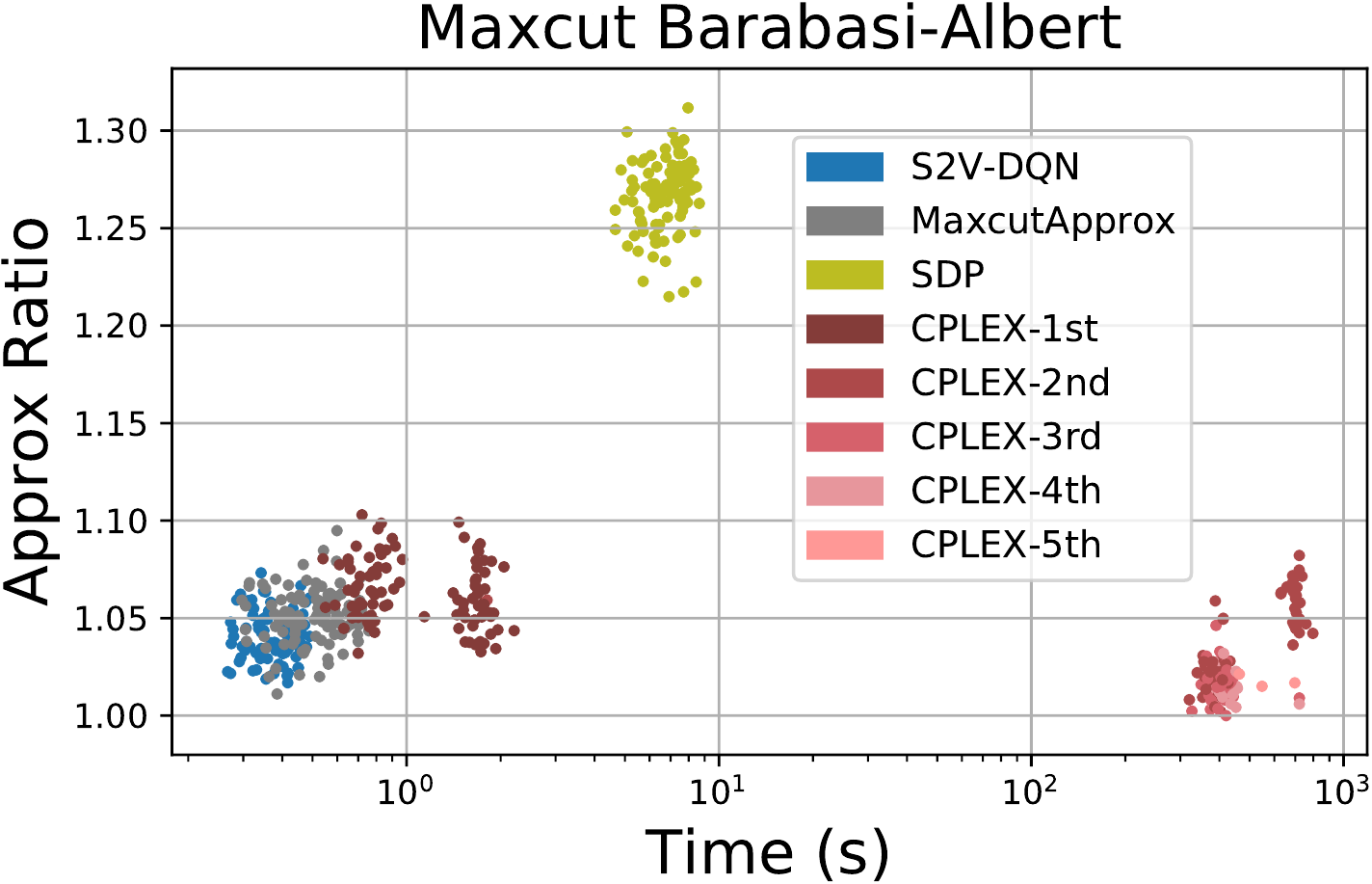}
			\\
			{\small (a) MVC BA 200-300} & {\small (b) MAXCUT BA 200-300}	
		\end{tabular}	  
	  \end{minipage}
	  \begin{minipage}[c]{.29\textwidth}
	    \centering
	%     \captionsetup{font=default, textfont={scriptsize}}         
	    \caption{\small Time-approximation trade-off for MVC and MAXCUT. In this figure, each dot represents a solution found for a single problem instance, for 100 instances. For CPLEX, we also record the time and quality of each solution it finds, e.g. CPLEX-1st means the first feasible solution found by CPLEX.}
	    \label{fig:tradeoff_sub}
	  \end{minipage}
	  \vspace{-2mm}
	\end{figure*}

	Figure~\ref{fig:tradeoff_sub} illustrates the approximation ratios of various approaches as a function of running time. 
	All algorithms report a single solution at termination, whereas CPLEX reports multiple improving solutions, for which we recorded the corresponding running time and approximation ratio. Figure~\ref{fig:tradeoff} (Appendix~\ref{app:fulltradeoff}) includes other graph sizes and types, where the results are consistent with Figure~\ref{fig:tradeoff_sub}.
	
	Figure~\ref{fig:tradeoff_sub} shows that, for MVC, we are slightly slower than the approximation algorithms but enjoy a much better approximation ratio. Also note that although CPLEX found the first feasible solution quickly, it also has much worse ratio; the second improved solution found by CPLEX takes similar or longer time than our S2V-DQN, but is still of worse quality.
	For MAXCUT, the observations are still consistent. One should be aware that sometimes our algorithm can obtain better results than 1-hour CPLEX, which gives ratios below $1.0$. Furthermore, sometimes S2V-DQN is even faster than the \textit{MaxcutApprox}, although this comparison is not exactly fair, since we use GPUs; however, we can still see that our algorithm is efficient. 
	
	\vspace{-2mm}
	\subsection{Experiments on real-world datasets}
	\vspace{-2mm}
	
	In addition to the experiments for synthetic data, we identified sets of publicly available benchmark or real-world instances for each problem, and performed experiments on them. A summary of results is in Table~\ref{tab:realdatasummary}, and details are given in Appendix~\ref{app:real}. S2V-DQN significantly outperforms all competing methods for MVC, MAXCUT and TSP.

% Table generated by Excel2LaTeX from sheet 'Sheet1'
\begin{table}[htbp]
  \vspace{-2mm}
  \centering
  \caption{\small Realistic data experiments, results summary. Values are average approximation ratios.}
    \resizebox{\columnwidth}{!}{
    \begin{tabular}{l|l|lll}
    \toprule
    \textbf{Problem} & \textbf{Dataset} & \multicolumn{1}{l}{\textbf{S2V-DQN}} & \multicolumn{1}{l}{\textbf{Best Competitor}} & \multicolumn{1}{l}{\textbf{2\textsuperscript{nd} Best Competitor}} \\
    \midrule
    MVC   & MemeTracker & \textbf{1.0021} & 1.2220 (MVCApprox-Greedy) & 1.4080  (MVCApprox) \\
    MAXCUT & Physics & \textbf{1.0223} & 1.2825 (MaxcutApprox) & 1.8996 (SDP) \\
    TSP   & TSPLIB & \textbf{1.0475} & 1.0800 (Farthest) & 1.0947 (2-opt)  \\
    \bottomrule
    \end{tabular}%
}
  \label{tab:realdatasummary}%
  \vspace{-2mm}
\end{table}%

	\vspace{-3mm}
	\subsection{Discovery of interesting new algorithms}
	\vspace{-2mm}
	
	We further examined the algorithms learned by S2V-DQN, and tried to interpret what greedy heuristics have been learned. We found that S2V-DQN is able to discover new and interesting algorithms which intuitively make sense but have not been analyzed before. For instance, S2V-DQN discovers an algorithm for MVC where nodes are selected to balance between their degrees and the connectivity of the remaining graph~(Appendix Figures~\ref{fig:viz_mvc} and~\ref{fig:mvc_comp_viz}). For MAXCUT, S2V-DQN discovers an algorithm where nodes are picked to avoid cancelling out existing edges in the cut set (Appendix Figure~\ref{fig:viz_maxcut}). These results suggest that S2V-DQN may also be a good assistive tool for discovering new algorithms, especially in cases when the graph optimization problems are new and less well-studied. 
	
	\vspace{-3mm}
	\section{Conclusions}
	\vspace{-2mm} 
	
	We presented an end-to-end machine learning framework for automatically designing greedy heuristics for hard combinatorial optimization problems on graphs. Central to our approach is the combination of a deep graph embedding  with reinforcement learning. Through extensive experimental evaluation, we demonstrate the effectiveness of the proposed framework in learning greedy heuristics as compared to manually-designed greedy algorithms. The excellent performance of the learned heuristics is consistent across multiple different problems, graph types, and graph sizes, suggesting that the framework is a promising new tool for designing algorithms for graph problems.  

\vspace{-2mm} 
\subsubsection*{Acknowledgments}
\vspace{-2mm} 
This project was supported in part by NSF IIS-1218749, NIH BIGDATA 1R01GM108341, NSF CAREER IIS-1350983, NSF IIS-1639792 EAGER, NSF CNS-1704701, ONR N00014-15-1-2340, Intel ISTC, NVIDIA and Amazon AWS. Dilkina is supported by NSF grant CCF-1522054 and ExxonMobil.

 \bibliographystyle{unsrtnat}
% \bibliography{../../bibfile/bibfile} % Hanjun
%\bibliography{../../../../bibfile2013/bibfile} %le

\clearpage
	\newpage
	\onecolumn
	
	\begin{center}
		{\LARGE\bf Appendix}
	\end{center}
	
	\begin{appendix}
		\counterwithin{figure}{section}
		\counterwithin{table}{section}		
		
			\section{Related Work}
		
		\noindent\textbf{Machine learning for combinatorial optimization.}
		Reinforcement learning is used to solve a job-shop flow scheduling problem in~\citep{ZhaDie00}. Boyan and Moore~\citep{BoyMoo00} use regression to learn good restart rules for local search algorithms. Both of these methods require hand-designed, problem-specific features, a limitation with the learned graph embedding. 
		%More recently, Cornish and Torr used deep reinforcement learning to optimize submodular functions -- a specific class of discrete functions~\citep{CornishTorr16}.
		%Baluja and Davies~\citep{BalDav97} learn a probabilistic tree of the dependencies between the variables of a combinatorial problem, and use the tree to generate candidate solutions; in contrast, we learn a constructive algorithm using a vastly different methodology.
		
		\noindent\textbf{Machine learning for branch-and-bound.}
		\textit{Learning to search} in branch-and-bound is another related research thread. This thread includes machine learning methods for branching~\citep{LagoudakisLittman01,KhaLebSonNemetal16}, tree node selection~\citep{HeDauEis14,SabSamRed12}, and heuristic selection~\citep{SamMem07,KhaDilNemetal17}. In comparison, our work promotes an even tighter integration of learning and optimization.
		
		\noindent\textbf{Deep learning for continuous optimization.}
		In continuous optimization, methods have been proposed for learning an update rule for gradient descent~\citep{AndDenGomHofetal16,LiMalik16} and solving black-box optimization problems~\citep{CheHofColDenetal16}; these are very interesting ideas that highlight the possibilities for better algorithm design through learning.
		
		\section{Set Covering Problem}
		\label{app:scp}
		We also applied our framework to the classical Set Covering Problem (SCP). SCP is interesting because it is not a graph problem, but can be formulated as one. Our framework is capable of addressing such problems seamlessly, as we will show in the coming sections of the appendix which detail the performance of S2V-DQN as compared to other methods.

\textbf{Set Covering Problem (SCP)}: Given a bipartite graph $G$ with node set $V\coloneqq \mathcal{U}\cup\mathcal{C}$, find a subset of nodes $S\subseteq\mathcal{C}$ such that every node in $\mathcal{U}$ is covered, i.e. $u\in\mathcal{U} \Leftrightarrow \exists s\in{S} $ s.t. $(u,s)\in E$, and $|S|$ is minimized. Note that an edge $(u,s), u\in\mathcal{U}, s\in\mathcal{C}$, exists whenever subset $s$ includes element $u$.

{\bf Meta-algorithm:} Same as MVC; the termination criterion checks whether all nodes in $\mathcal{U}$ have been covered. 		

{\bf RL formulation:} In SCP, the state is a function of the subset of nodes of $\mathcal{C}$ selected so far; an action is to add node of $\mathcal{C}$ to the partial solution; the reward is -1; the termination criterion is met when all nodes of $\mathcal{U}$ are covered; no helper function is needed.

\textbf{Baselines for SCP:} We include \emph{Greedy}, which iteratively selects the node of $\mathcal{C}$ that is not in the current partial solution and that has the most uncovered neighbors in $\mathcal{U}$~\cite{KleinbergTardos06}. We also used \emph{LP}, another $O(\log{|\mathcal{U}|)}$-approximation that solves a linear programming relaxation of SCP, and rounds the resulting fractional solution in decreasing order of variable values (SortLP-1 in~\cite{PelSchWoo93}).

		\section{Experimental Results on Realistic Data}
		\label{app:real}
In this section, we show results on realistic nstances for all four problems. In particular, for MVC and SCP, we used the MemeTracker graph to formulate network diffusion optimization problems. For MAXCUT and TSP, we used benchmark instances that arise in physics and transportation, respectively.
          
\subsection{Minimum Vertex Cover}
As mentioned in the introduction, the MVC problem is related to the efficient spreading of information in networks, where one wants to cover as few nodes as possible such that all nodes have at least one neighbor in the cover.
The MemeTracker graph~\footnote{\url{{http://snap.stanford.edu/netinf/\#data}}} is a network of who-copies-whom, where nodes represent news sites or blogs, and a (directed) edge from $u$ to $v$ means that $v$ frequently copies phrases (or memes) from $u$. The network is learned from real traces in~\cite{GomLesKra10}, having 960 nodes and 5000 edges. The dataset also provides the average transmission time $\Delta_{u,v}$ between a pair of nodes, i.e. how much later $v$ copies $u$'s phrases after their publication online, on average. As done in~\cite{KhaDilSon14}, we use these average transmission times to compute a diffusion probability $P(u,v)$ on the edge, such that $P(u,v)=\alpha\cdot \dfrac{1}{\Delta_{u,v}}$, where $\alpha$ is a parameter of the diffusion model. In both MVC and SCP, we use $\alpha=0.1$, but results are consistent for other values we have considered. For pairs of nodes that have edges in both directions, i.e. $(u,v)$ and $(v,u)$, we take the average probability to obtain an undirected 
version of the graph, as MVC is defined for undirected graphs. 

Following the widely-adopted Independent Cascade model (see~\cite{DuSonRodZha13} for example), we sample a diffusion cascade from the full graph by independently keeping an edge with probability $P(u,v)$. We then consider the largest connected component in the graph as a single training instance, and train S2V-DQN on a set of such sampled diffusion graphs. The aim is to test the learned model on the (undirected version of the) \textit{full} MemeTracker graph.

Experimentally, an optimal cover has 473 nodes, whereas S2V-DQN finds a cover with 474 nodes, only one more than the optimum, at an approximation ratio of $1.002$. In comparison, MVCApprox and MVCApprox-Greedy find much larger covers with 666 and 578 nodes, at approximation ratios of $1.408$ and $1.222$, respectively.

\subsection{Maximum Cut}
A library of Maximum Cut instances is publicly available~\footnote{\url{http://www.optsicom.es/maxcut/\#instances}}, and includes synthetic and realistic instances that are widely used in the optimization community (see references at library website). We perform experiments on a subset of the instances available, namely ten problems from Ising spin glass models in physics, given that they are realistic and manageable in size (the first 10 instances in Set2 of the library). All ten instances have 125 nodes and 375 edges, with edge weights in $\{-1,0,1\}$.

To train our S2V-DQN model, we constructed a training dataset by perturbing the instances, adding random Gaussian noise with mean 0 and standard deviation 0.01 to the edge weights. After training, the learned model is used to construct a cut-set greedily on each of the ten instances, as before.

Table~\ref{tab:real-maxcut} shows that S2V-DQN finds near-optimal solutions (optimal in 3/10 instances) that are much better than those found by competing methods. 
%We note that the SDP approach of~\cite{GoeWil95} loses its theoretical guarantees when the edge weights may be negative, which may explain its very poor performance.

\label{app:real-maxcut}
% Table generated by Excel2LaTeX from sheet 'Sheet1'
\begin{table}[htbp]
  \centering
  \caption{MAXCUT results on the ten instances described in~\ref{app:real-maxcut}; values reported are cut weights of the solution returned by each method, where larger values are better (best in bold). Bottom row is the average approximation ratio (lower is better).}
    \begin{tabular}{l|c|ccc}
    \toprule
    \textbf{Instance} & \textbf{OPT} & \textbf{S2V-DQN} & \textbf{MaxcutApprox} & \textbf{SDP} \\
    \midrule
    G54100 & 110   & \textbf{108} & 80    & 54 \\
    G54200 & 112   & \textbf{108} & 90    & 58 \\
    G54300 & 106   & \textbf{104} & 86    & 60 \\
    G54400 & 114   & \textbf{108} & 96    & 56 \\
    G54500 & 112   & \textbf{112} & 94    & 56 \\
    G54600 & 110   & \textbf{110} & 88    & 66 \\
    G54700 & 112   & \textbf{108} & 88    & 60 \\
    G54800 & 108   & \textbf{108} & 76    & 54 \\
    G54900 & 110   & \textbf{108} & 88    & 68 \\
    G5410000 & 112   & \textbf{108} & 80    & 54 \\
    \midrule
    Approx. ratio & 1     & \textbf{1.02} & 1.28  & 1.90 \\
    \bottomrule
    \end{tabular}%
  \label{tab:real-maxcut}%
\end{table}%

\subsection{Traveling Salesman Problem}
We use the standard TSPLIB library~\cite{Reinelt91} which is publicly available~\footnote{\url{http://elib.zib.de/pub/mp-testdata/tsp/tsplib/tsp/index.html}}. We target 38 TSPLIB instances with sizes ranging from 51 to 318 cities (or nodes). We do not tackle larger instances as we are limited by the memory of a single graphics card. Nevertheless, most of the instances addressed here are larger than the largest instance used in~\cite{BelPhaLeNoretal16}.

We apply S2V-DQN in ``Active Search" mode, similarly to~\cite{BelPhaLeNoretal16}: no upfront training phase is required, and the reinforcement learning algorithm~\ref{alg:rl} is applied on-the-fly on each instance. The best tour encountered over the episodes of the RL algorithm is stored.

Table~\ref{tab:real-tsp} shows the results of our method and six other TSP algorithms. On all but 6 instances, S2V-DQN finds the best tour among all methods. The average approximation ratio of S2V-DQN is also the smallest at $1.05$.

\begin{table}[htbp]
  \centering
  \caption{TSPLIB results: Instances are sorted by increasing size, with the number at the end of an instance's name indicating its size. Values reported are the cost of the tour found by each method (lower is better, best in bold). Bottom row is the average approximation ratio (lower is better).}
  \resizebox{\columnwidth}{!}{
    \begin{tabular}{l|r|rrrrrrrr}
	\toprule
	Instance & OPT   & S2V-DQN & Farthest & 2-opt & Cheapest & Christofides & Closest & Nearest & MST \\
	\midrule
	eil51 & 426   & \textbf{439} & 467   & 446   & 494   & 527   & 488   & 511   & 614 \\
	berlin52 & 7,542 & \textbf{7,542} & 8,307 & 7,788 & 9,013 & 8,822 & 9,004 & 8,980 & 10,402 \\
	st70  & 675   & \textbf{696} & 712   & 753   & 776   & 836   & 814   & 801   & 858 \\
	eil76 & 538   & \textbf{564} & 583   & 591   & 607   & 646   & 615   & 705   & 743 \\
	pr76  & 108,159 & \textbf{108,446} & 119,692 & 115,460 & 125,935 & 137,258 & 128,381 & 153,462 & 133,471 \\
	rat99 & 1,211 & \textbf{1,280} & 1,314 & 1,390 & 1,473 & 1,399 & 1,465 & 1,558 & 1,665 \\
	kroA100 & 21,282 & \textbf{21,897} & 23,356 & 22,876 & 24,309 & 26,578 & 25,787 & 26,854 & 30,516 \\
	kroB100 & 22,141 & \textbf{22,692} & 23,222 & 23,496 & 25,582 & 25,714 & 26,875 & 29,158 & 28,807 \\
	kroC100 & 20,749 & \textbf{21,074} & 21,699 & 23,445 & 25,264 & 24,582 & 25,640 & 26,327 & 27,636 \\
	kroD100 & 21,294 & 22,102 & \textbf{22,034} & 23,967 & 25,204 & 27,863 & 25,213 & 26,947 & 28,599 \\
	kroE100 & 22,068 & 22,913 & 23,516 & \textbf{22,800} & 25,900 & 27,452 & 27,313 & 27,585 & 30,979 \\
	rd100 & 7,910 & \textbf{8,159} & 8,944 & 8,757 & 8,980 & 10,002 & 9,485 & 9,938 & 10,467 \\
	eil101 & 629   & \textbf{659} & 673   & 702   & 693   & 728   & 720   & 817   & 847 \\
	lin105 & 14,379 & \textbf{15,023} & 15,193 & 15,536 & 16,930 & 16,568 & 18,592 & 20,356 & 21,167 \\
	pr107 & 44,303 & \textbf{45,113} & 45,905 & 47,058 & 52,816 & 49,192 & 52,765 & 48,521 & 55,956 \\
	pr124 & 59,030 & \textbf{61,623} & 65,945 & 64,765 & 65,316 & 64,591 & 68,178 & 69,297 & 82,761 \\
	bier127 & 118,282 & \textbf{121,576} & 129,495 & 128,103 & 141,354 & 135,134 & 145,516 & 129,333 & 153,658 \\
	ch130 & 6,110 & \textbf{6,270} & 6,498 & 6,470 & 7,279 & 7,367 & 7,434 & 7,578 & 8,280 \\
	pr136 & 96,772 & \textbf{99,474} & 105,361 & 110,531 & 109,586 & 116,069 & 105,778 & 120,769 & 142,438 \\
	pr144 & 58,537 & \textbf{59,436} & 61,974 & 60,321 & 73,032 & 74,684 & 73,613 & 61,652 & 77,704 \\
	ch150 & 6,528 & \textbf{6,985} & 7,210 & 7,232 & 7,995 & 7,641 & 7,914 & 8,191 & 9,203 \\
	kroA150 & 26,524 & \textbf{27,888} & 28,658 & 29,666 & 29,963 & 32,631 & 31,341 & 33,612 & 38,763 \\
	kroB150 & 26,130 & \textbf{27,209} & 27,404 & 29,517 & 31,589 & 33,260 & 31,616 & 32,825 & 35,289 \\
	pr152 & 73,682 & \textbf{75,283} & 75,396 & 77,206 & 88,531 & 82,118 & 86,915 & 85,699 & 90,292 \\
	u159  & 42,080 & \textbf{45,433} & 46,789 & 47,664 & 49,986 & 48,908 & 52,009 & 53,641 & 54,399 \\
	rat195 & 2,323 & \textbf{2,581} & 2,609 & 2,605 & 2,806 & 2,906 & 2,935 & 2,753 & 3,163 \\
	d198  & 15,780 & 16,453 & \textbf{16,138} & 16,596 & 17,632 & 19,002 & 17,975 & 18,805 & 19,339 \\
	kroA200 & 29,368 & \textbf{30,965} & 31,949 & 32,760 & 35,340 & 37,487 & 36,025 & 35,794 & 40,234 \\
	kroB200 & 29,437 & 31,692 & \textbf{31,522} & 33,107 & 35,412 & 34,490 & 36,532 & 36,976 & 40,615 \\
	ts225 & 126,643 & \textbf{136,302} & 140,626 & 138,101 & 160,014 & 145,283 & 151,887 & 152,493 & 188,008 \\
	tsp225 & 3,916 & \textbf{4,154} & 4,280 & 4,278 & 4,470 & 4,733 & 4,780 & 4,749 & 5,344 \\
	pr226 & 80,369 & \textbf{81,873} & 84,130 & 89,262 & 91,023 & 98,101 & 100,118 & 94,389 & 114,373 \\
	gil262 & 2,378 & \textbf{2,537} & 2,623 & 2,597 & 2,800 & 2,963 & 2,908 & 3,211 & 3,336 \\
	pr264 & 49,135 & \textbf{52,364} & 54,462 & 54,547 & 57,602 & 55,955 & 65,819 & 58,635 & 66,400 \\
	a280  & 2,579 & \textbf{2,867} & 3,001 & 2,914 & 3,128 & 3,125 & 2,953 & 3,302 & 3,492 \\
	pr299 & 48,191 & \textbf{51,895} & 51,903 & 54,914 & 58,127 & 58,660 & 59,740 & 61,243 & 65,617 \\
	lin318 & 42,029 & 45,375 & 45,918 & \textbf{45,263} & 49,440 & 51,484 & 52,353 & 54,019 & 60,939 \\
	linhp318 & 41,345 & 45,444 & 45,918 & \textbf{45,263} & 49,440 & 51,484 & 52,353 & 54,019 & 60,939 \\
	\midrule
	Approx. ratio & 1     & \textbf{1.05}  & 1.08  & 1.09  & 1.18  & 1.2   & 1.21  & 1.24  & 1.37 \\
	\bottomrule
\end{tabular}%
}
  \label{tab:real-tsp}%
\end{table}%

\subsection{Set Covering Problem}		
The SCP is also related to the diffusion optimization problem on graphs; for instance, the proof of hardness in the classical~\cite{KemKleTar03} paper uses SCP for the reduction. As in MVC, we leverage the MemeTracker graph, albeit differently.

We use the same cascade model as in MVC to assign the edge probabilities, and sample graphs from it in the same way. Let $\mathcal{R}^{G}(u)$ be the set of nodes reachable from $u$ in a sampled graph $G$. For every node $u$ in $G$, there are two corresponding nodes in the SCP instance, $u_{\mathcal{C}}\in\mathcal{C}$ and $u_{\mathcal{U}}\in\mathcal{U}$. An edge exists between $u_{\mathcal{C}}\in\mathcal{C}$ and $v_{\mathcal{U}}\in\mathcal{U}$ if and only if $v\in\mathcal{R}^{G}(u)$. In other words, each node in the sampled graph $G$ has a set consisting of the other nodes that it can reach in $G$. As such, the SCP reduces to finding the smallest set of nodes whose union can reach all other nodes. We generate training and testing graphs according to this same process, with $\alpha=0.1$.

Experimentally, we test S2V-DQN and the other baseline algorithms on a set of 1000 test graphs. S2V-DQN achieves an average approximation ratio of $1.001$, only slightly behind LP, which achieves $1.0009$, and well ahead of Greedy at $1.03$.
		
		\section{Experiment Details}
		
		\subsection{Problem instance generation}
		\label{app:instances}
		\subsubsection{Minimum Vertex Cover}
		For the Minimum Vertex Cover (MVC) problem, we generate random Erd\H{o}s-Renyi (edge probability 0.15) and Barabasi-Albert (average degree 4) graphs of various sizes, and use the integer programming solver CPLEX 12.6.1 with a time cutoff of 1 hour to compute optimal solutions for the generated instances. When CPLEX fails to find an optimal solution, we report the best one found within the time cutoff as ``optimal". All graphs were generated using the NetworkX~\footnote{\url{https://networkx.github.io/}} package in Python.
		%\begin{equation*}
		%\begin{aligned}
		%&\text{min}\sum_{i\in V}{x_i}\\
		%\text{subject to}\;\;
		%&x_i + x_j \geqslant 1, \;\; (i,j)\in E\\
		%&x_i \text{ binary}
		%\end{aligned}
		%\end{equation*}
		\subsubsection{Maximum Cut}
		For the Maximum Cut (MAXCUT) problem, we use the same graph generation process as in MVC, and augment each edge with a weight drawn uniformly at random from $[0,1]$. We use a quadratic formulation of MAXCUT with CPLEX 12.6.1. and a time cutoff of 1 hour to compute optimal solutions, and report the best solution found as ``optimal".
		%\begin{equation*}
		%\begin{aligned}
		%&\text{max}\sum_{(i,j)\in E}{w(i,j)z_{ij}}\\
		%\text{subject to}\;\;
		%&z_{ij}\leqslant x_i + x_j, \;\; (i,j)\in E\\
		%&z_{ij}\leqslant 2 - (x_i + x_j), \;\; (i,j)\in E\\
		%&x_i \text{ binary}
		%\end{aligned}
		%\end{equation*}
		\subsubsection{Traveling Salesman Problem}
%		For the Graphical TSP problem (GTSP)~\citep{CorFonNad85}, we use the same graph generators as in MVC and MAXCUT, but also guarantee that the generated graphs are connected (i.e. there exists a path between any two nodes of a graph). The connectivity is guaranteed by using the largest connected component of the generated graph as the final graph. Integer edge weights are drawn uniformly at random from $[1,10000]$, as the solver Concorde can only handle integer distances.
		
		For the (symmetric) 2-dimensional TSP, we use the instance generator of the 8th DIMACS Implementation Challenge~\footnote{\url{http://dimacs.rutgers.edu/Challenges/TSP/}}~\citep{JohnsonMcgeoch07} to generate two types of Euclidean instances: ``random" instances consist of $n$ points scattered uniformly at random in the $[10^6,10^6]$ square, while ``clustered" instances consist of $n$ points that are clustered into $n/100$ clusters; generator details are described in page 373 of~\citep{JohnsonMcgeoch07}. 
		
		To compute optimal TSP solutions for both  TSP, we use the state-of-the-art solver, Concorde~\footnote{\url{http://www.math.uwaterloo.ca/tsp/concorde/}}~\citep{AppBixChvCoo06}, with a time cutoff of 1 hour.  

		\subsubsection{Set Covering Problem}
		For the SCP, given a number of node $n$, roughly $0.2n$ nodes are in node-set $\mathcal{C}$, and the rest in node-set $\mathcal{U}$. An edge between nodes in $\mathcal{C}$ and $\mathcal{U}$ exists with probability either $0.05$ or $0.1$, which can be seen as ``density" values, and commonly appear for instances used in optimization papers on SCP~\cite{BalasHo80}. We guarantee that each node in $\mathcal{U}$ has at least 2 edges, and each node in $\mathcal{C}$ has at least one edge, a standard measure for SCP instances~\cite{BalasHo80}. We also use CPLEX 12.6.1. with a time cutoff of 1 hour to compute a near-optimal or optimal solution to a SCP instance.
		
		\subsection{ Full results on solution quality}
\begin{figure*}[th!]
				\centering
				\setlength{\tabcolsep}{3pt}
				\begin{tabular}{cccc}
					\includegraphics[width=0.45\textwidth]{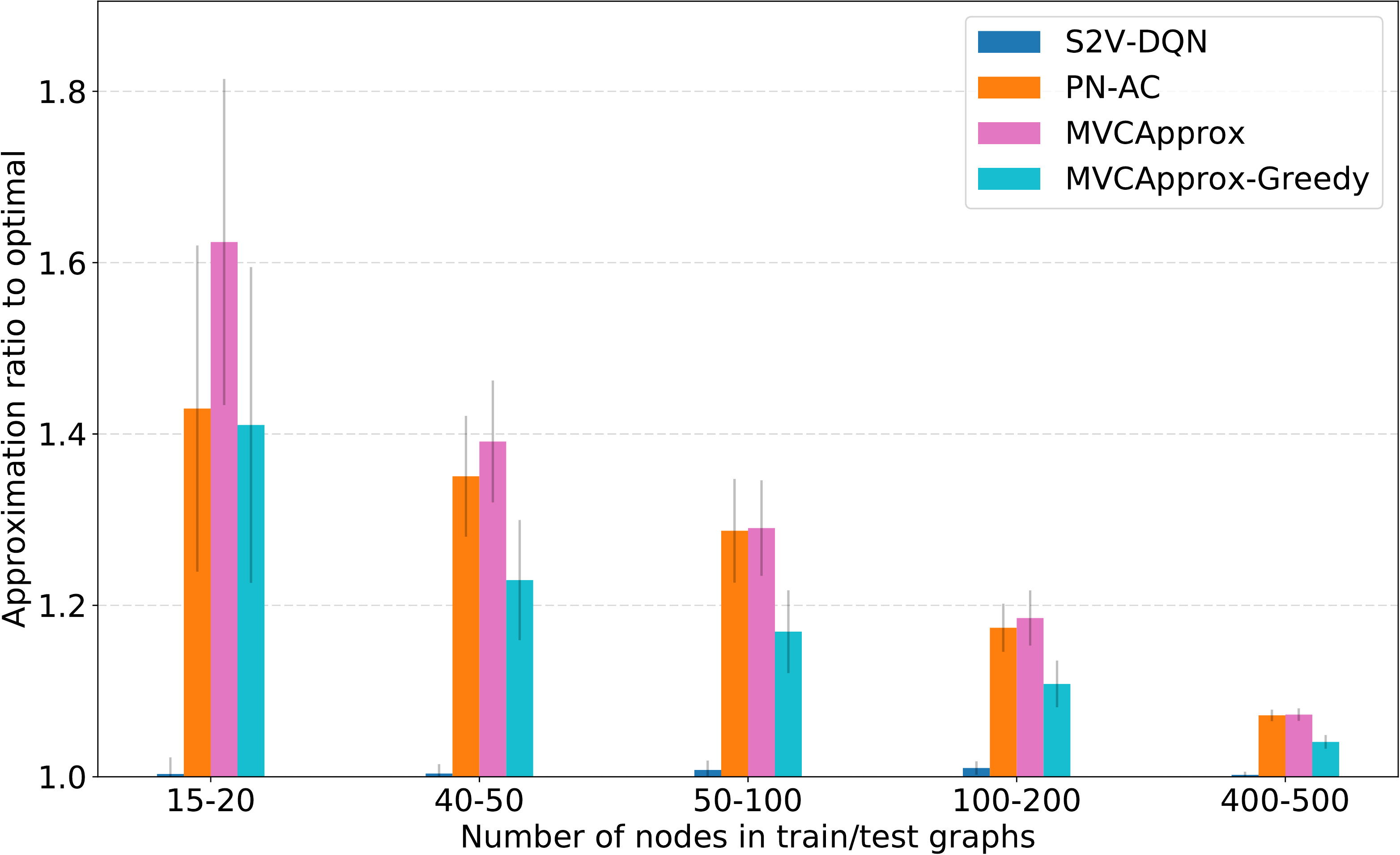} & 
					\includegraphics[width=0.45\textwidth]{results/task-mvc-gtype-barabasi_albert-crop} \\
					(a) MVC ER & (b) MVC BA \\
					\includegraphics[width=0.45\textwidth]{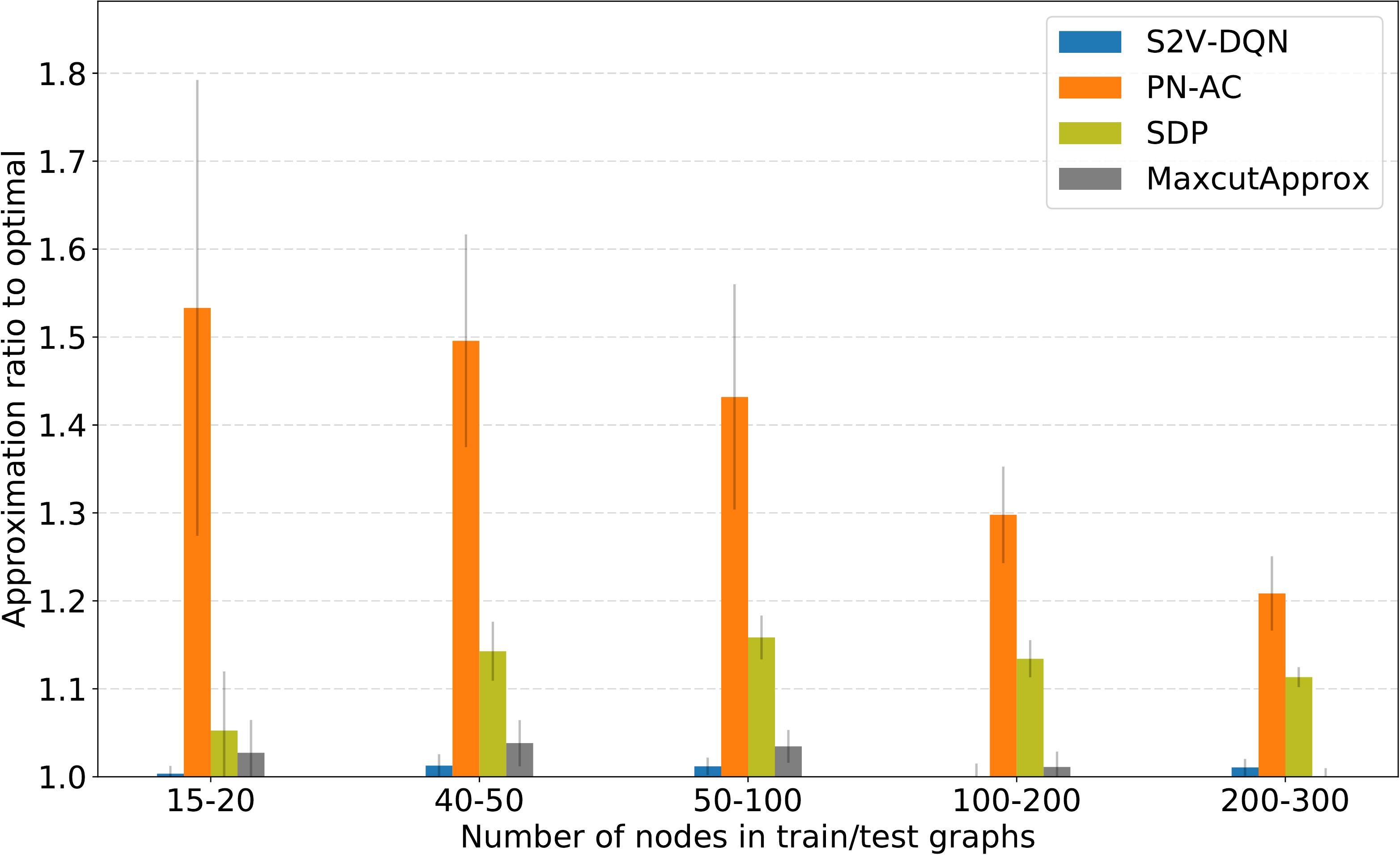} &   
					\includegraphics[width=0.45\textwidth]{results/task-maxcut-gtype-barabasi_albert-crop} \\
					(c) MAXCUT ER & (d) MAXCUT BA\\
		
					\includegraphics[width=0.45\textwidth]{results/task-tsp2d-gtype-random-crop} &   
					\includegraphics[width=0.45\textwidth]{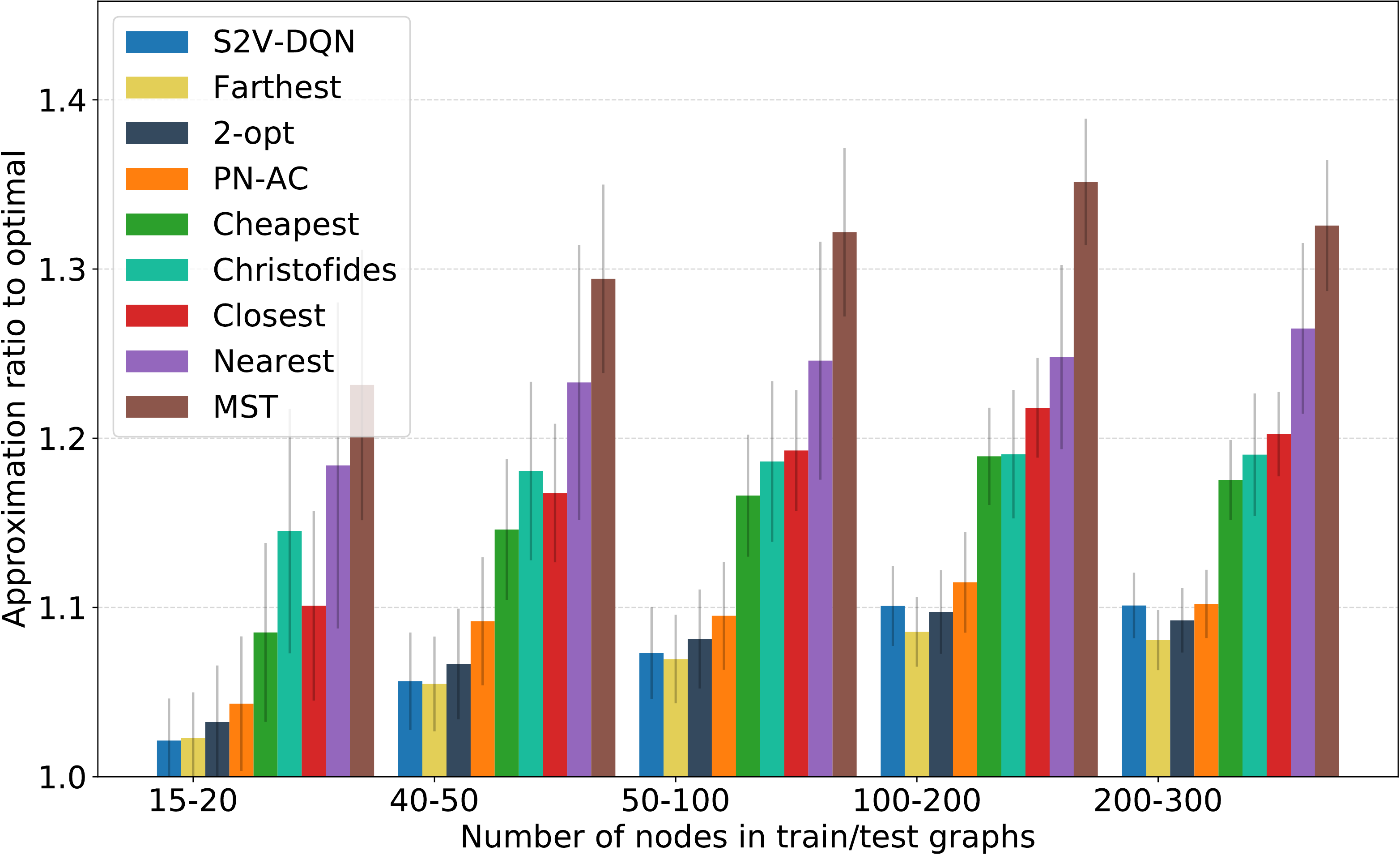} \\
					(e) TSP random & (f) TSP clustered \\
					\includegraphics[width=0.45\textwidth]{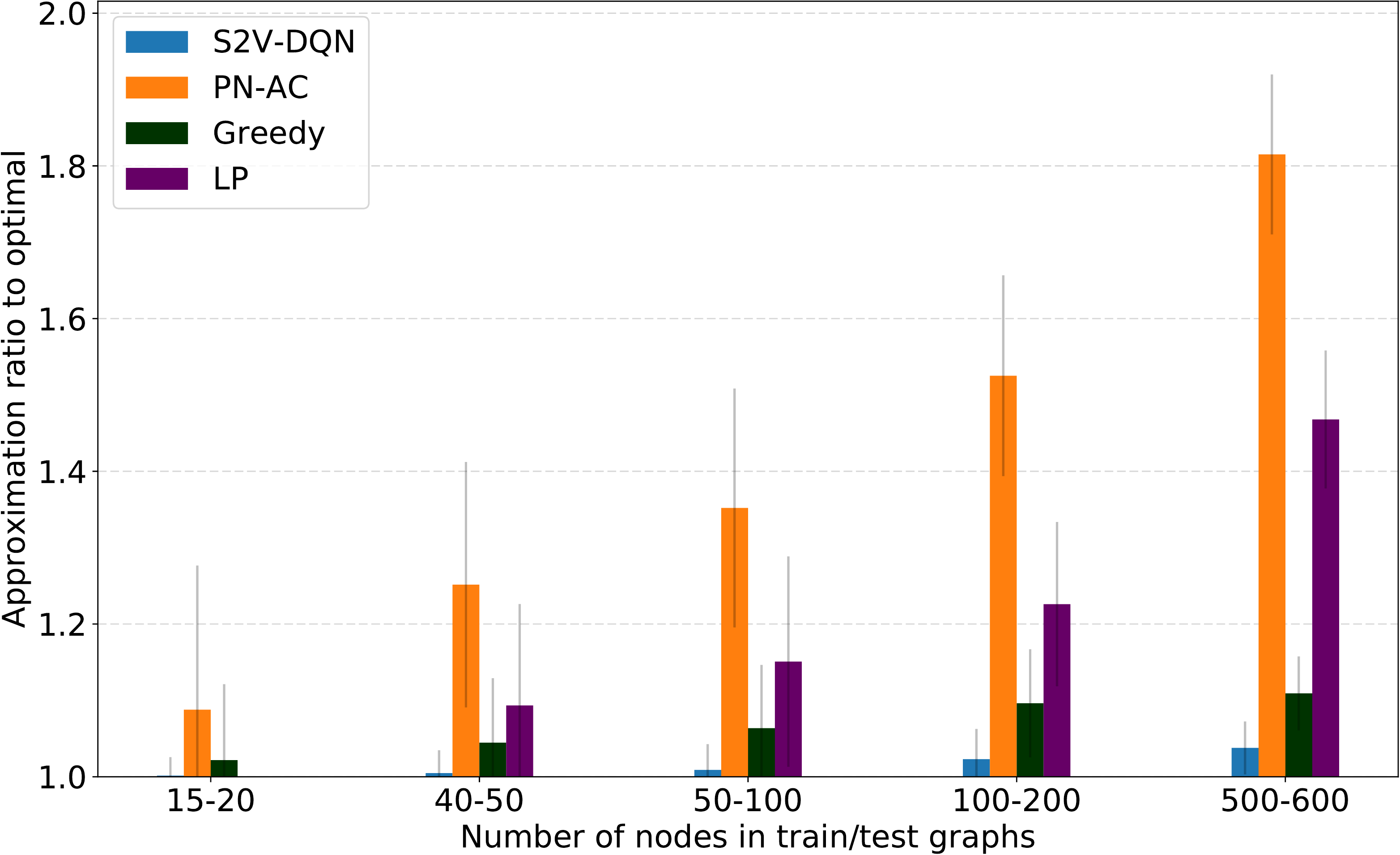} &   
					\includegraphics[width=0.45\textwidth]{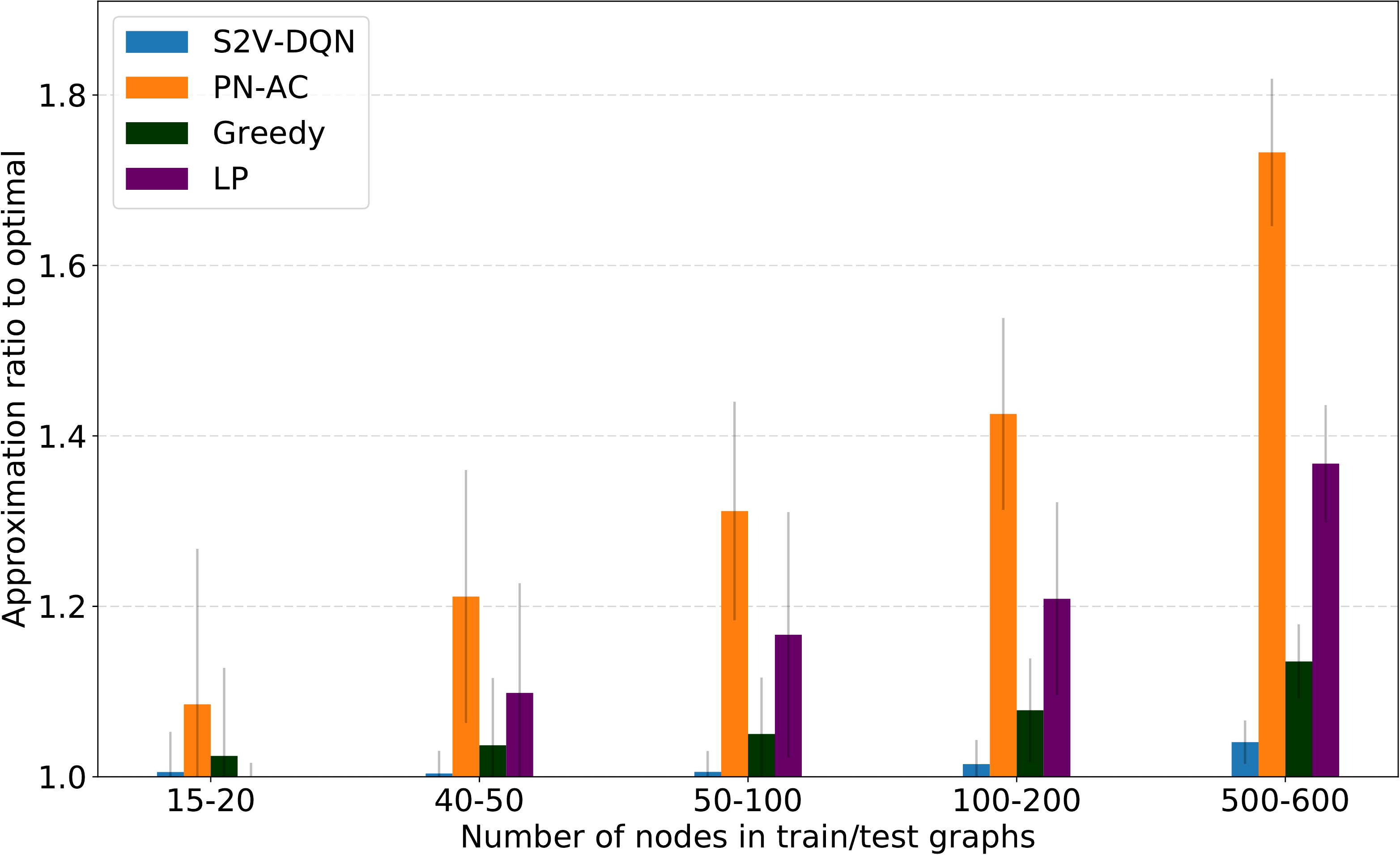}\\			
					 (g) SCP 0.1 & (h) SCP 0.05
		%			\\
		%			\includegraphics[width=0.45\textwidth]{results/task-tspmult-gtype-erdos_renyi-crop} &   
		%			\includegraphics[width=0.45\textwidth]{results/task-tspmult-gtype-barabasi_albert-crop} \\
		%			(g) GTSP ER & (h) GTSP BA 
				\end{tabular}   
				%  \vspace{-4mm}
				\caption{Approximation ratio on 1000 test graphs. Note that on MVC, our performance is pretty close to optimal. In this figure, training and testing graphs are generated according to the same distribution. 
				}
				\label{fig:sol_qual_full}
				%  \vspace{-3mm}
			\end{figure*}

		Table~\ref{fig:sol_qual_full} is a complete version of Table~\ref{fig:sol_qual} that appears in the main text.
		\subsection{Full results on generalization}
		\label{app:full_result}
		
		The full generalization results can be found in Table~\ref{tab:s2vgen_mvc_er},~\ref{tab:s2vgen_mvc_ba},~\ref{tab:s2vgen_MAXCUT_er},~\ref{tab:s2vgen_MAXCUT_ba},~\ref{tab:s2vgen_TSP2D_er}, ~\ref{tab:s2vgen_TSP2D_ba} ,~\ref{tab:s2vgen_SCP_0.05} and~\ref{tab:s2vgen_SCP_0.1}.
%		~\ref{tab:s2vgen_GTSP_er} and~\ref{tab:s2vgen_GTSP_ba}. 
		
		% 
		% \begin{figure*}[h]
		% \centering
		% 
		% \begin{subfigure}[t]{0.49\textwidth}
		% \centering
		%   \includegraphics[width=\textwidth]{results/task-mvc-gtype-barabasi_albert-crop}
		%   \caption{MVC BA}
		% \end{subfigure}
		% \begin{subfigure}[t]{0.49\textwidth}
		% \centering
		%   \includegraphics[width=\textwidth]{results/task-maxcut-gtype-erdos_renyi-crop}
		%   \caption{MAXCUT ER}
		% \end{subfigure}
		% \begin{subfigure}[t]{0.49\textwidth}
		% \centering
		%   \includegraphics[width=\textwidth]{results/task-tspmult-gtype-erdos_renyi-crop}
		%   \caption{GTSP ER}
		% \end{subfigure}
		% \begin{subfigure}[t]{0.49\textwidth}
		% \centering
		%   \includegraphics[width=\textwidth]{results/task-tsp2d-gtype-random-crop}
		%   \caption{TSP2D random}
		% \end{subfigure}
		% %\begin{subfigure}[t]{0.49\textwidth}
		% %\centering
		% %  \includegraphics[width=\textwidth]{results/task-tsp2d-gtype-clustered-crop}
		% %  \caption{TSP2D clustered}
		% %\end{subfigure}
		% 	\caption{Approximation ratios on 1000 test graphs on various tasks and graph types.}	
		% 	\label{fig:app_sol_more}
		% \end{figure*}
		% 
		% Here we present the full test results, including generalization performance, on all of the tasks. 
		% 
		% Generally, on MVC and MaxCut, our algorithm's performance is quite stable, and can generalize to quite large sizes. For TSP2D and GTSP, it becomes harder to generalize. However, we can still achieve a 10\% approximation ratio. 
		
		\begin{table*}[htbp]
			\centering
			\resizebox{1.0\textwidth}{!}{%	
			\begin{tabular}{c|c|c|c|c|c|c|c|c|c}
				\toprule
				\diagbox{Train}{Test} & 15-20 & 40-50 & 50-100 & 100-200 & 200-300 & 300-400 & 400-500 & 500-600 & 1000-1200 \\
				\hline
				15-20 & 1.0032 & 1.0883 & 1.0941 & 1.0710 & 1.0484 & 1.0365 & 1.0276 & 1.0246 & 1.0111 \\
				\hline
				40-50 & \diagbox{}{} & 1.0037 & 1.0076 & 1.1013 & 1.0991 & 1.0800 & 1.0651 & 1.0573 & 1.0299 \\
				\hline
				50-100 & \diagbox{}{} & \diagbox{}{} & 1.0079 & 1.0304 & 1.0570 & 1.0532 & 1.0463 & 1.0427 & 1.0238  \\
				\hline
				100-200 & \diagbox{}{} & \diagbox{}{} & \diagbox{}{} & 1.0102 & 1.0095 & 1.0136 & 1.0142 & 1.0125 & 1.0103 \\
				\hline
				400-500 & \diagbox{}{} & \diagbox{}{} & \diagbox{}{} & \diagbox{}{} & \diagbox{}{} & \diagbox{}{} & 1.0021 & 1.0027 & 1.0057 \\
				\hline
			\end{tabular}%
			}
			\caption{S2V-DQN's generalization on MVC problem in ER graphs.}
			\label{tab:s2vgen_mvc_er}%
		\end{table*}
		
		\begin{table*}[htbp]
			\centering
			\resizebox{1.0\textwidth}{!}{%	
			\begin{tabular}{c|c|c|c|c|c|c|c|c|c}
				\toprule
				\diagbox{Train}{Test} & 15-20 & 40-50 & 50-100 & 100-200 & 200-300 & 300-400 & 400-500 & 500-600 & 1000-1200 \\
				\hline
				15-20 & 1.0016 & 1.0027 & 1.0039 & 1.0066 & 1.0093 & 1.0106 & 1.0125 & 1.0150 & 1.0491 \\
				\hline
				40-50 & \diagbox{}{} & 1.0027 & 1.0051 & 1.0092 & 1.0130 & 1.0144 & 1.0161 & 1.0170 & 1.0228 \\
				\hline
				50-100 & \diagbox{}{} & \diagbox{}{} & 1.0033 & 1.0041 & 1.0045 & 1.0040 & 1.0045 & 1.0048 & 1.0062 \\
				\hline
				100-200 & \diagbox{}{} & \diagbox{}{} & \diagbox{}{} & 1.0016 & 1.0020 & 1.0019 & 1.0021 & 1.0026 & 1.0060 \\
				\hline
				400-500 & \diagbox{}{} & \diagbox{}{} & \diagbox{}{} & \diagbox{}{} & \diagbox{}{} & \diagbox{}{} & 1.0025 & 1.0026 & 1.0030 \\
				\hline
			\end{tabular}%
			}
			\caption{S2V-DQN's generalization on MVC problem in BA graphs.}
			\label{tab:s2vgen_mvc_ba}%
		\end{table*}
		
		\begin{table*}[htbp]
			\centering
			\resizebox{1.0\textwidth}{!}{%	
			\begin{tabular}{c|c|c|c|c|c|c|c|c|c}
				\toprule
				\diagbox{Train}{Test} & 15-20 & 40-50 & 50-100 & 100-200 & 200-300 & 300-400 & 400-500 & 500-600 & 1000-1200 \\
				\hline
				15-20 & 1.0034 & 1.0167 & 1.0407 & 1.0667 & 1.1067 & 1.1489 & 1.1885 & 1.2150 & 1.1488 \\
				\hline
				40-50 & \diagbox{}{} & 1.0127 & 1.0154 & 1.0089 & 1.0198 & 1.0383 & 1.0388 & 1.0384 & 1.0534 \\
				\hline
				50-100 & \diagbox{}{} & \diagbox{}{} & 1.0112 & 1.0024 & 1.0109 & 1.0467 & 1.0926 & 1.1426 & 1.1297 \\
				\hline
				100-200 & \diagbox{}{} & \diagbox{}{} & \diagbox{}{} & 1.0005 & 1.0021 & 1.0211  & 1.0373 & 1.0612 & 1.2021 \\
				\hline
				200-300 & \diagbox{}{} & \diagbox{}{} & \diagbox{}{} & \diagbox{}{} & 1.0106 & 1.0272 & 1.0487 & 1.0700 & 1.1759 \\
				\hline
			\end{tabular}%
			}
			\caption{S2V-DQN's generalization on MAXCUT problem in ER graphs.}
			\label{tab:s2vgen_MAXCUT_er}%
		\end{table*}
		
		\begin{table*}[htbp]
			\centering
			\resizebox{1.0\textwidth}{!}{%	
			\begin{tabular}{c|c|c|c|c|c|c|c|c|c}
				\toprule
				\diagbox{Train}{Test} & 15-20 & 40-50 & 50-100 & 100-200 & 200-300 & 300-400 & 400-500 & 500-600 & 1000-1200 \\
				\hline
				15-20 & 1.0055 & 1.0119 & 1.0176 & 1.0276 & 1.0357 & 1.0386 & 1.0335 & 1.0411 & 1.0331 \\
				\hline
				40-50 & \diagbox{}{} & 1.0107 & 1.0119 & 1.0139 & 1.0144 & 1.0119 & 1.0039 & 1.0085 & 0.9905 \\
				\hline
				50-100 & \diagbox{}{} & \diagbox{}{} & 1.0150 & 1.0181 & 1.0202 & 1.0188 & 1.0123 & 1.0177 & 1.0038 \\
				\hline
				100-200 & \diagbox{}{} & \diagbox{}{} & \diagbox{}{} & 1.0166 & 1.0183 & 1.0166 & 1.0104 & 1.0166 & 1.0156 \\
				\hline
				200-300 & \diagbox{}{} & \diagbox{}{} & \diagbox{}{} & \diagbox{}{} & 1.0420 & 1.0394 & 1.0290 & 1.0319 & 1.0244\\
				\hline
			\end{tabular}%
			}
			\caption{S2V-DQN's generalization on MAXCUT problem in BA graphs.}
			\label{tab:s2vgen_MAXCUT_ba}%
		\end{table*}
		
		\begin{table*}[htbp]
			\centering
			\resizebox{1.0\textwidth}{!}{%	
			\begin{tabular}{c|c|c|c|c|c|c|c|c|c}
				\toprule
				\diagbox{Train}{Test} & 15-20 & 40-50 & 50-100 & 100-200 & 200-300 & 300-400 & 400-500 & 500-600 & 1000-1200 \\
				\hline
				15-20 & 1.0147 & 1.0511 & 1.0702 & 1.0913 & 1.1022 & 1.1102 & 1.1124 & 1.1156 & 1.1212 \\
				\hline
				40-50 & \diagbox{}{} & 1.0533 & 1.0701 & 1.0890 & 1.0978 & 1.1051 & 1.1583 & 1.1587 & 1.1609 \\
				\hline
				50-100 & \diagbox{}{} & \diagbox{}{} & 1.0701 & 1.0871 & 1.0983 & 1.1034 & 1.1071 & 1.1101 & 1.1171  \\
				\hline
				100-200 & \diagbox{}{} & \diagbox{}{} & \diagbox{}{} & 1.0879 & 1.0980 & 1.1024 & 1.1056 & 1.1080 & 1.1142 \\
				\hline
				200-300 & \diagbox{}{} & \diagbox{}{} & \diagbox{}{} & \diagbox{}{} & 1.1049 & 1.1090 & 1.1084 & 1.1114 & 1.1179 \\
				\hline
			\end{tabular}%
			}
			\caption{S2V-DQN's generalization on TSP in random graphs.}
			\label{tab:s2vgen_TSP2D_er}%
		\end{table*}
		
		\begin{table*}[htbp]
			\centering
			\resizebox{1.0\textwidth}{!}{%	
			\begin{tabular}{c|c|c|c|c|c|c|c|c|c}
				\toprule
				\diagbox{Train}{Test} & 15-20 & 40-50 & 50-100 & 100-200 & 200-300 & 300-400 & 400-500 & 500-600 & 1000-1200 \\
				\hline
				15-20 & 1.0214 & 1.0591 & 1.0761 & 1.0958 & 1.0938 & 1.0966 & 1.1009 & 1.1012 & 1.1085 \\
				\hline
				40-50 & \diagbox{}{} & 1.0564 & 1.0740 & 1.0939 & 1.0904 & 1.0951 & 1.0974 & 1.1014 & 1.1091 \\
				\hline
				50-100 & \diagbox{}{} & \diagbox{}{} & 1.0730 & 1.0895 & 1.0869 & 1.0918 & 1.0944 & 1.0975 & 1.1065  \\
				\hline
				100-200 & \diagbox{}{} & \diagbox{}{} & \diagbox{}{} & 1.1009 & 1.0979 & 1.1013 & 1.1059 & 1.1048 & 1.1091 \\
				\hline
				200-300 & \diagbox{}{} & \diagbox{}{} & \diagbox{}{} & \diagbox{}{} & 1.1012 & 1.1049 & 1.1080 & 1.1067 & 1.1112 \\
				\hline
			\end{tabular}%
			}
			\caption{S2V-DQN's generalization on TSP in clustered graphs.}
			\label{tab:s2vgen_TSP2D_ba}%
		\end{table*}

				\begin{table*}[htbp]
	\centering
	\resizebox{1.0\textwidth}{!}{%	
		\begin{tabular}{c|c|c|c|c|c|c|c|c|c}
			\toprule
			\diagbox{Train}{Test}            & 15-20 & 40-50 & 50-100 & 100-200 & \multicolumn{1}{l|}{200-300} & \multicolumn{1}{l|}{300-400} & \multicolumn{1}{l|}{400-500} & \multicolumn{1}{l|}{500-600} & \multicolumn{1}{l}{1000-1200} \\
			\midrule
			15-20 & \multicolumn{1}{r|}{1.0055} & \multicolumn{1}{r|}{1.0170} & \multicolumn{1}{r|}{1.0436} & \multicolumn{1}{r|}{1.1757} & 1.3910 & 1.6255 & 1.8768 & 2.1339 & 3.0574 \\
			\midrule
			40-50 & \diagbox{}{} & \multicolumn{1}{r|}{1.0039} & \multicolumn{1}{r|}{1.0083} & \multicolumn{1}{r|}{1.0241} & 1.0452 & 1.0647 & 1.0792 & 1.0858 & 1.0775 \\
			\midrule
			50-100 & \diagbox{}{} & \diagbox{}{} & \multicolumn{1}{r|}{1.0056} & \multicolumn{1}{r|}{1.0199} & 1.0382 & 1.0614 & 1.0845 & 1.0821 & 1.0620 \\
			\midrule
			100-200 & \diagbox{}{} & \diagbox{}{} & \diagbox{}{} & \multicolumn{1}{r|}{1.0147} & 1.0270 & 1.0417 & 1.0588 & 1.0774 & 1.0509 \\
			\midrule
			200-300 & \diagbox{}{} & \diagbox{}{} & \diagbox{}{} & \diagbox{}{} & 1.0273 & 1.0415 & 1.0828 & 1.1357 & 1.2349 \\
			\hline
		\end{tabular}%
	}
	\caption{S2V-DQN's generalization on SCP with edge probability 0.05.}
	\label{tab:s2vgen_SCP_0.05}%
\end{table*}		

				\begin{table*}[htbp]
			\centering
			\resizebox{1.0\textwidth}{!}{%	
				\begin{tabular}{c|c|c|c|c|c|c|c|c|c}
					\toprule
					\diagbox{Train}{Test}           & 15-20 & 40-50 & 50-100 & 100-200 & \multicolumn{1}{l|}{200-300} & \multicolumn{1}{l|}{300-400} & \multicolumn{1}{l|}{400-500} & \multicolumn{1}{l|}{500-600} & \multicolumn{1}{l}{1000-1200} \\
					\midrule
					15-20 & \multicolumn{1}{r|}{1.0015} & \multicolumn{1}{r|}{1.0200} & \multicolumn{1}{r|}{1.0369} & \multicolumn{1}{r|}{1.0795} & 1.1147 & 1.1290 & 1.1325 & 1.1255 & \multicolumn{1}{r}{1.0805} \\
					\midrule
					40-50 & \diagbox{}{} & \multicolumn{1}{r|}{1.0048} & \multicolumn{1}{r|}{1.0137} & \multicolumn{1}{r|}{1.0453} & 1.0849 & 1.1055 & 1.1052 & 1.0958 & \multicolumn{1}{r}{1.0618} \\
					\midrule
					50-100 & \diagbox{}{} & \diagbox{}{} & \multicolumn{1}{r|}{1.0090} & \multicolumn{1}{r|}{1.0294} & 1.0771 & 1.1180 & 1.1456 & 1.2161 & \multicolumn{1}{r}{1.0946} \\
					\midrule
					100-200 & \diagbox{}{} & \diagbox{}{} & \diagbox{}{} & \multicolumn{1}{r|}{1.0231} & 1.0394 & 1.0564 & 1.0702 & 1.0747 & \multicolumn{1}{r}{2.5055} \\
					\midrule
					200-300 & \diagbox{}{} & \diagbox{}{} & \diagbox{}{} & \diagbox{}{} & 1.0378 & 1.0517 & 1.0592 & 1.0556 & \multicolumn{1}{r}{1.3192} \\
					\hline
				\end{tabular}%
			}
			\caption{S2V-DQN's generalization on SCP with edge probability 0.1.}
			\label{tab:s2vgen_SCP_0.1}%
		\end{table*}

		\subsection{Experiment Configuration of S2V-DQN}
		\label{app:s2v_config}
		
		The node/edge representations and hyperparameters used in our experiments is shown in Table~\ref{tab:hyper_param}. For our method, we simply tune the hyperparameters on small graphs (i.e., the graphs with less than 50 nodes), and fix them for larger graphs. 
		
		\begin{table*}[htbp]
			\centering
			\resizebox{\textwidth}{!}{%
				\begin{tabular}{c|c|c|c|c|c|c}
					\toprule
					\textbf{Problem} & \textbf{Node tag} & \textbf{Edge feature} & \textbf{Embedding size $p$} & \textbf{$T$} & \textbf{Batch size} & \textbf{n-step} \\
					\hline
					Minimum Vertex Cover & 0/1 tag & N/A & 64 & 5 & 128  & 5 \\
					\hline
					Maximum Cut & 0/1 tag & edge length; end node tag & 64 & 3 & 64  & 1 \\
					\hline
					Traveling Salesman Problem & coordinates; 0/1 tag; start/end node & edge length; end node tag & 64 & 4 & 64 & 1  \\
					\hline
					Set Covering Problem & 0/1 tag & N/A  & 64 & 5 & 64 &  2  \\
					\bottomrule
				\end{tabular}%
				
			}
			\caption{S2V-DQN's configuration used in Experiment. }
			\label{tab:hyper_param}%
			
		\end{table*}%
		
		\subsection{Stabilizing the training of S2V-DQN}
		
		For the learning rate, we use exponential decay after a certain number of steps, where the decay factor is fixed to 0.95. We also anneal the exploration probability $\epsilon$ from 1.0 to 0.05 in a linear way. For the discounting factor used in MDP, we use 1.0 for MVC, MAXCUT and SCP. For TSP, we use 0.1. 
		
		We also normalize the intermediate reward by the maximum number of nodes. For Q-learning, it is also important to disentangle the actual $Q$ with obsolete $\tilde{Q}$, as mentioned in~\cite{MniKavSilRusetal15}. 

		Also for TSP with insertion helper function, we find it works better with \textit{negative} version of designed reward function. This sounds counter intuitive at the beginning. However, since typically the RL agent will bias towards most recent rewards, flipping the sign of reward function suggests a focus over future rewards. This is especially useful with the insertion construction. But it shows that designing a good reward function is still challenging for learning combinatorial algorithm, which we will investigate in our future work. 
		
\begin{figure*}[t!]
	\centering
	\setlength{\tabcolsep}{3pt}
	\begin{tabular}{cccc}
		\includegraphics[width=0.45\textwidth]{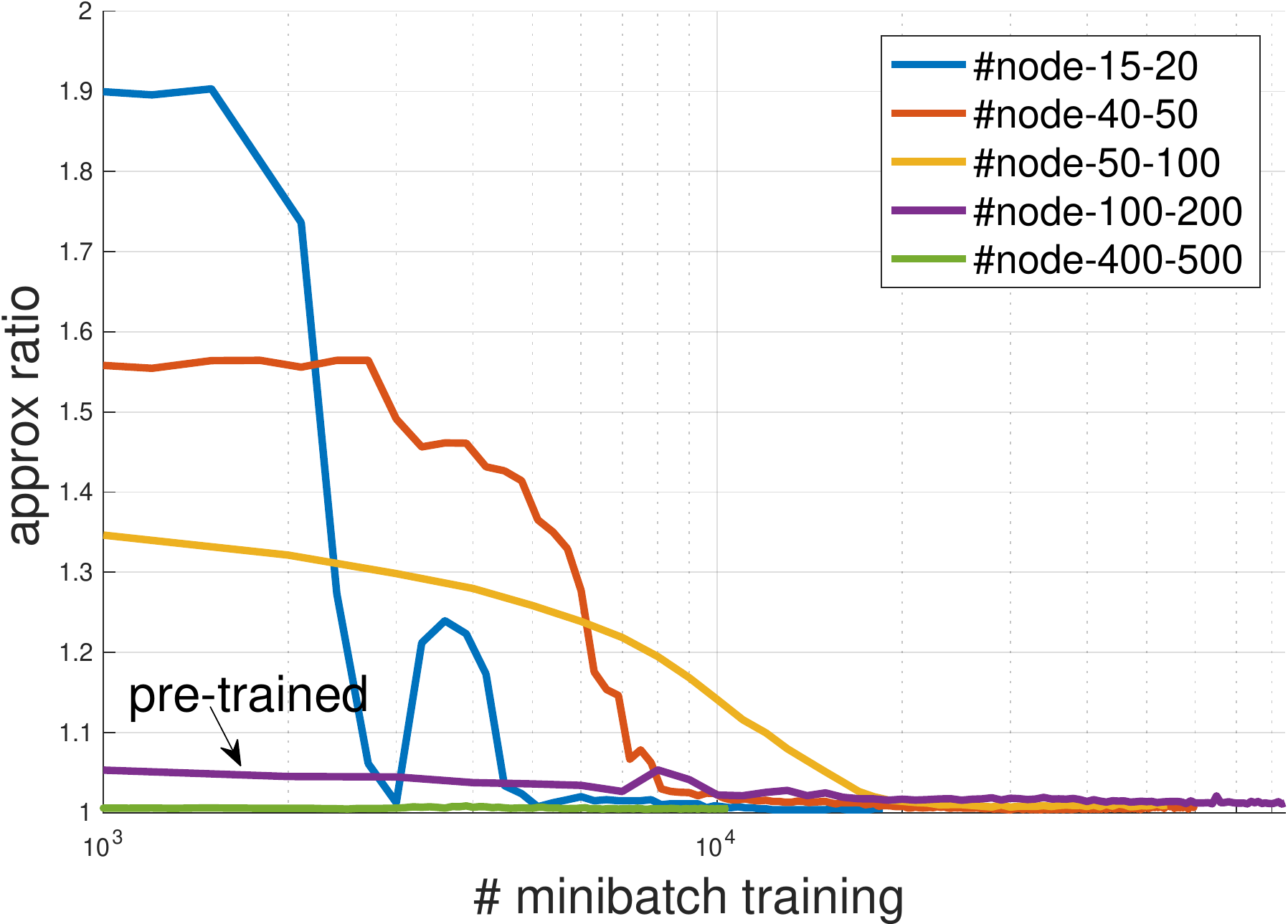} & 
		\includegraphics[width=0.45\textwidth]{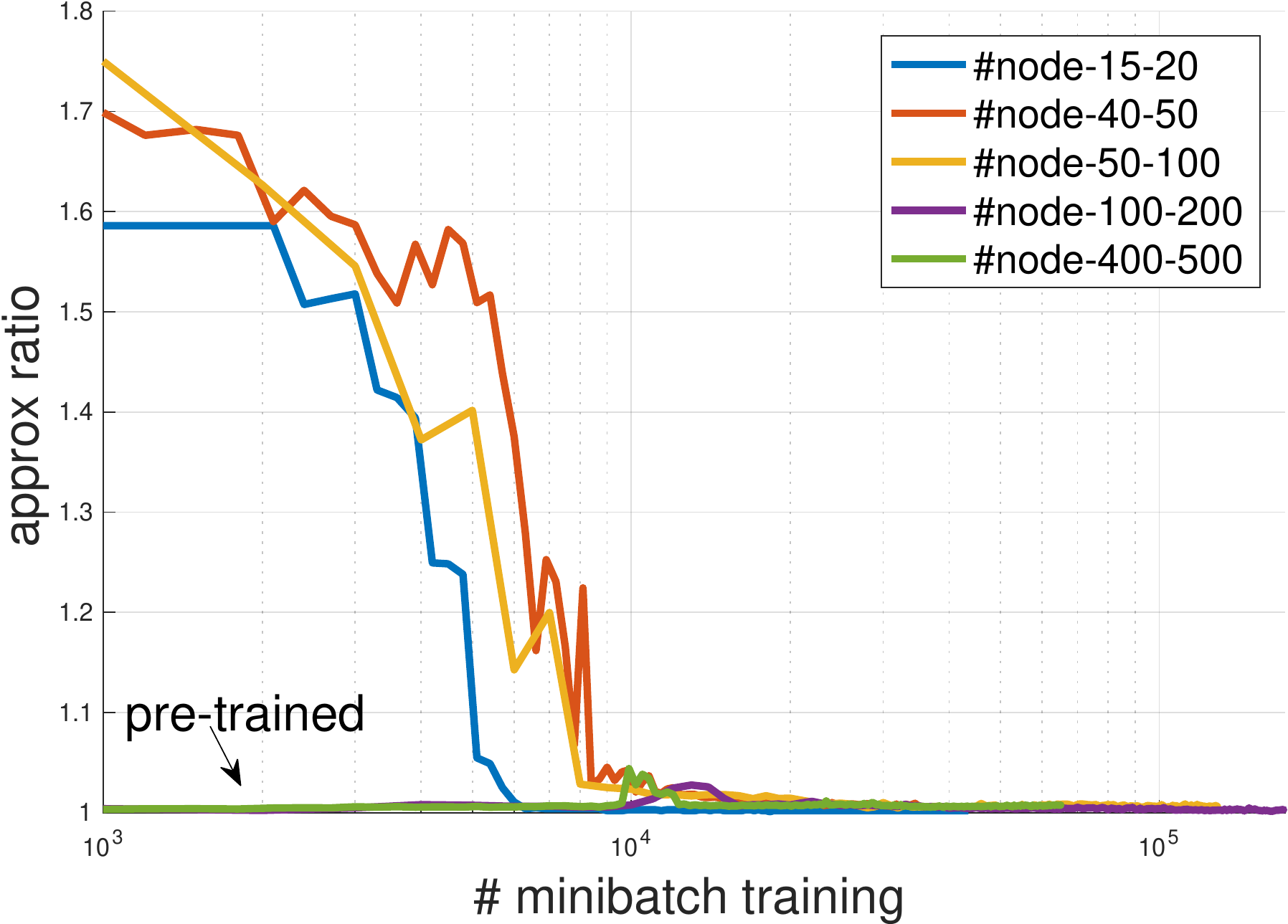} \\
		(a) MVC ER & (b) MVC BA \\
		\includegraphics[width=0.45\textwidth]{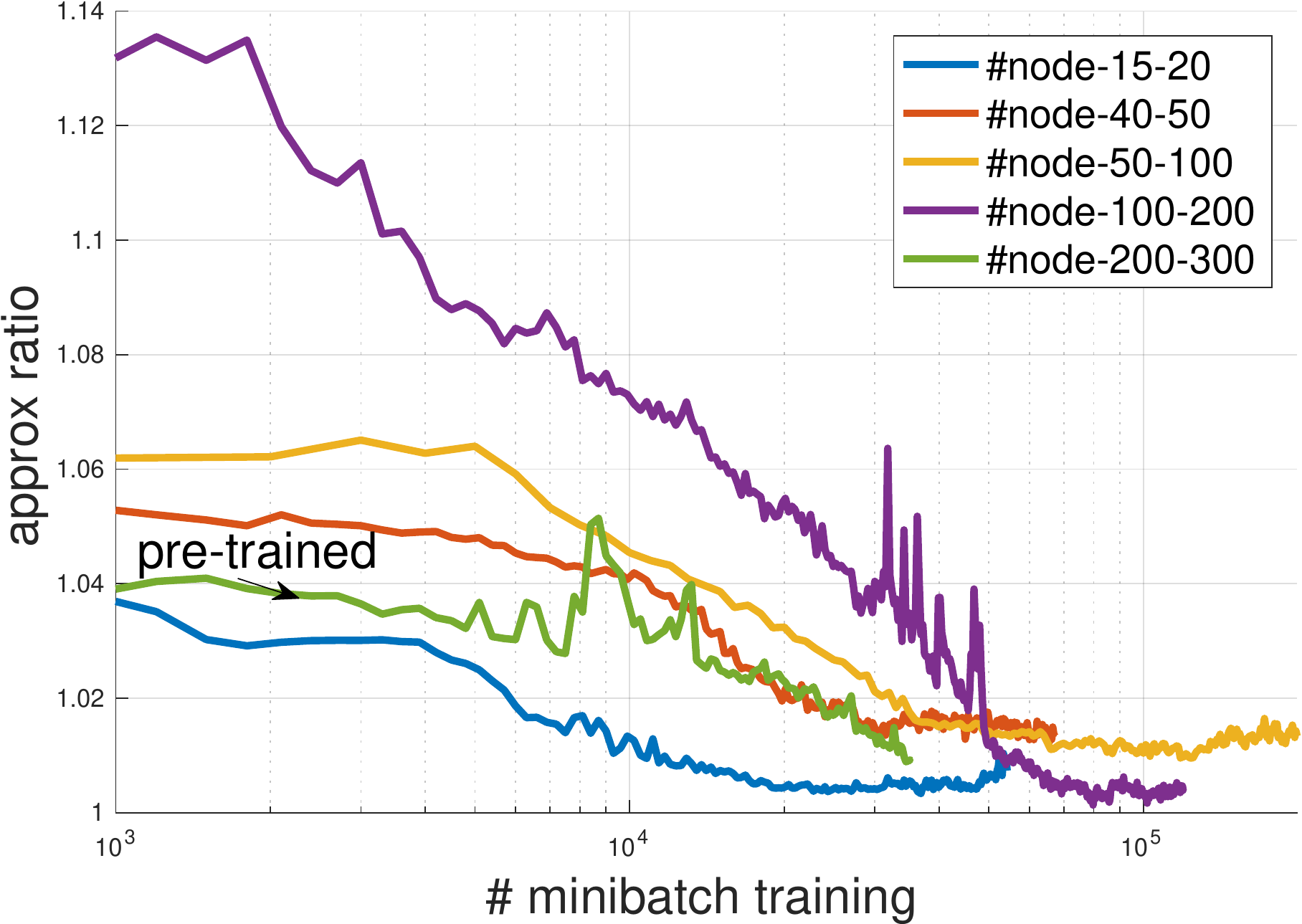} &   
		\includegraphics[width=0.45\textwidth]{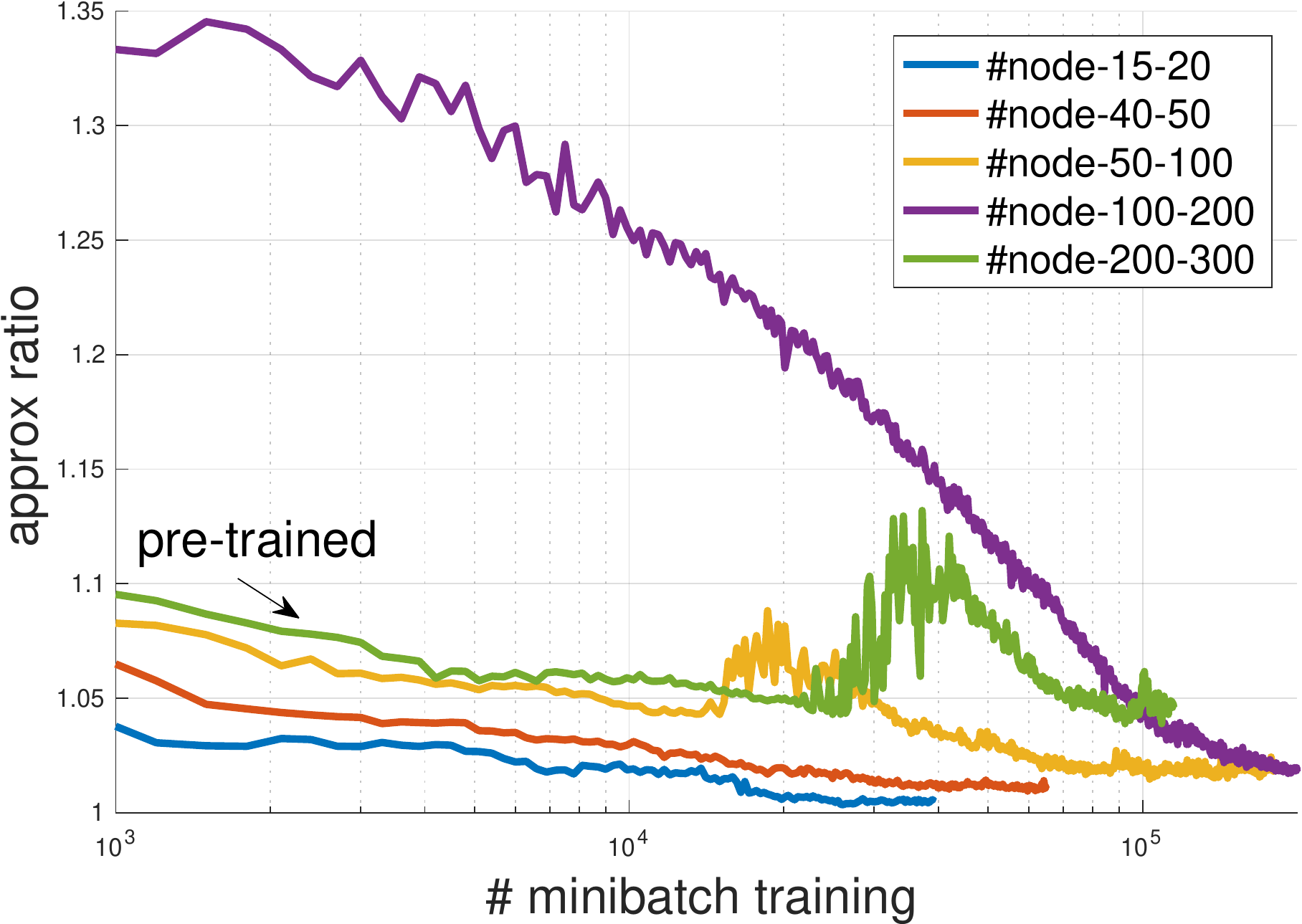} \\
		(c) MAXCUT ER & (d) MAXCUT BA \\
		\includegraphics[width=0.45\textwidth]{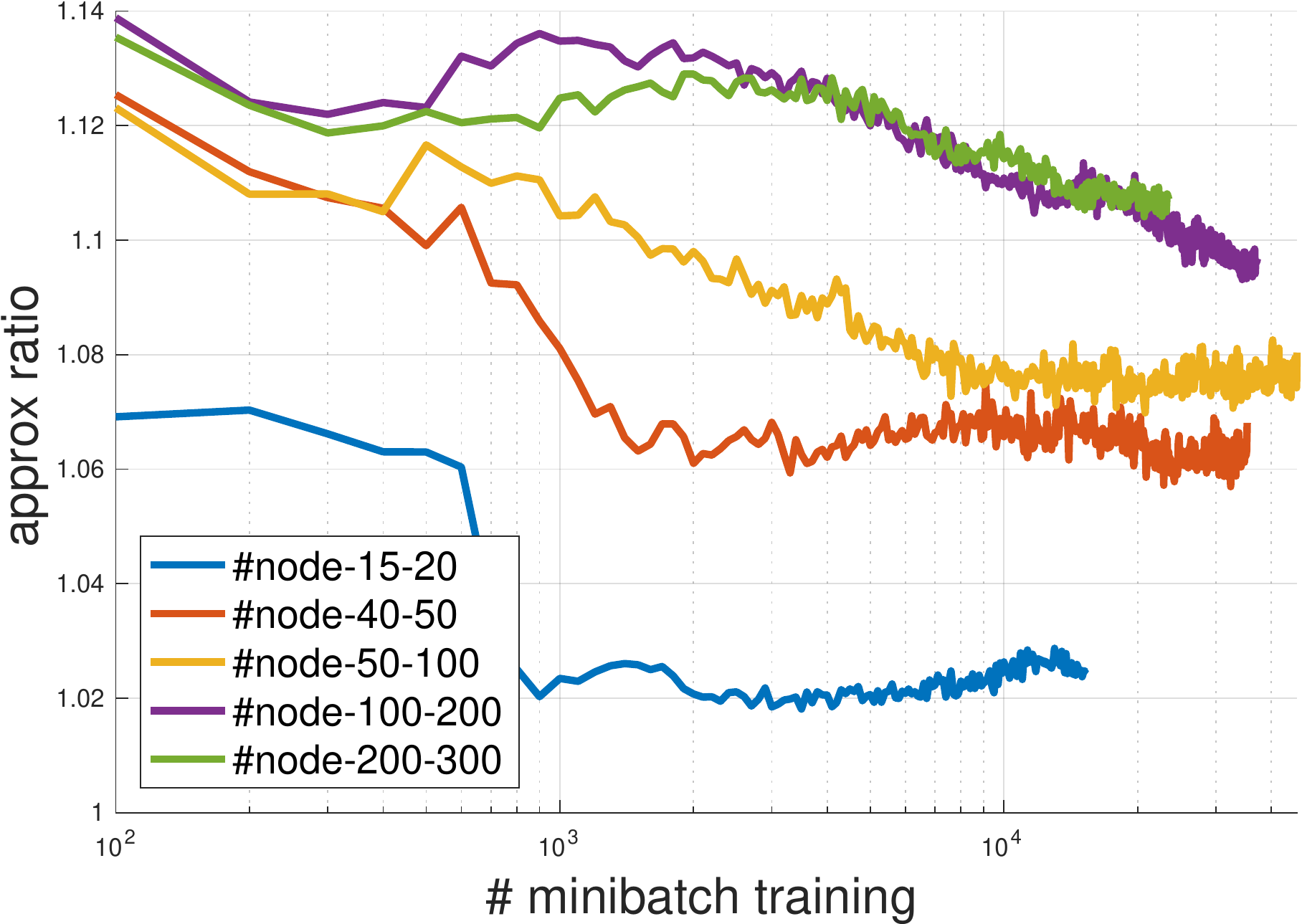} &   
		\includegraphics[width=0.45\textwidth]{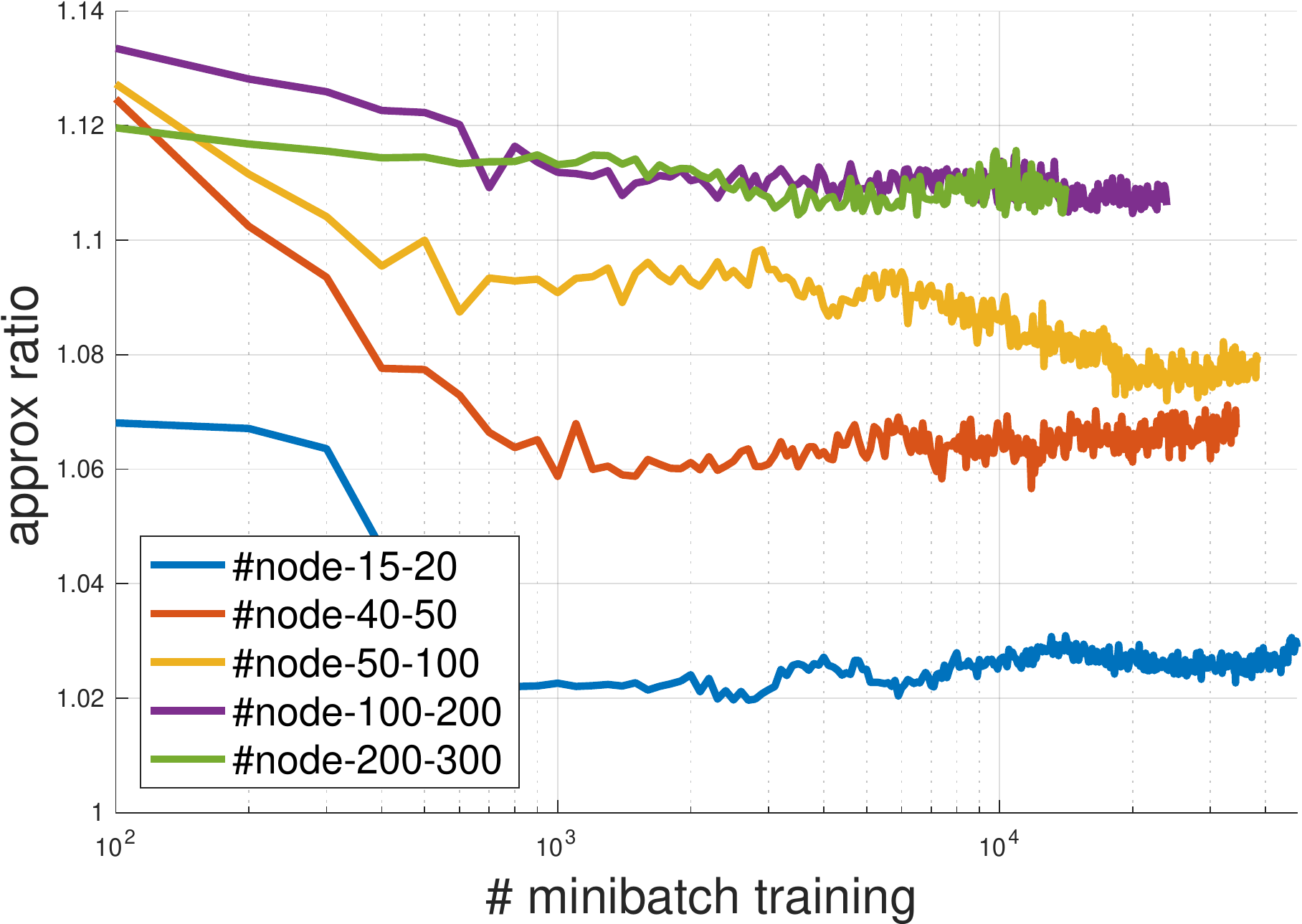} \\
		(e) TSP random & (f) TSP clustered \\
		\includegraphics[width=0.45\textwidth]{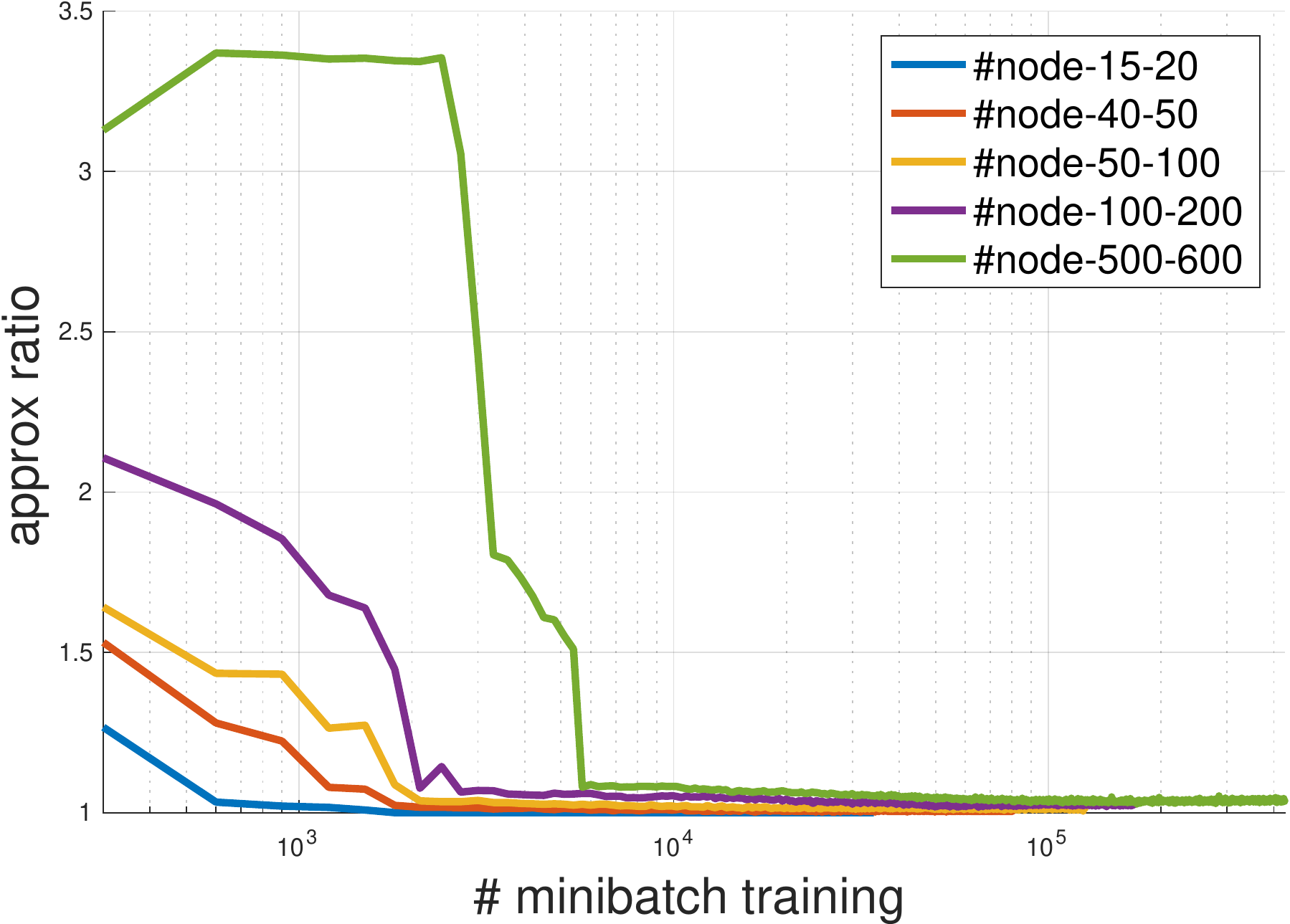} &   
		\includegraphics[width=0.45\textwidth]{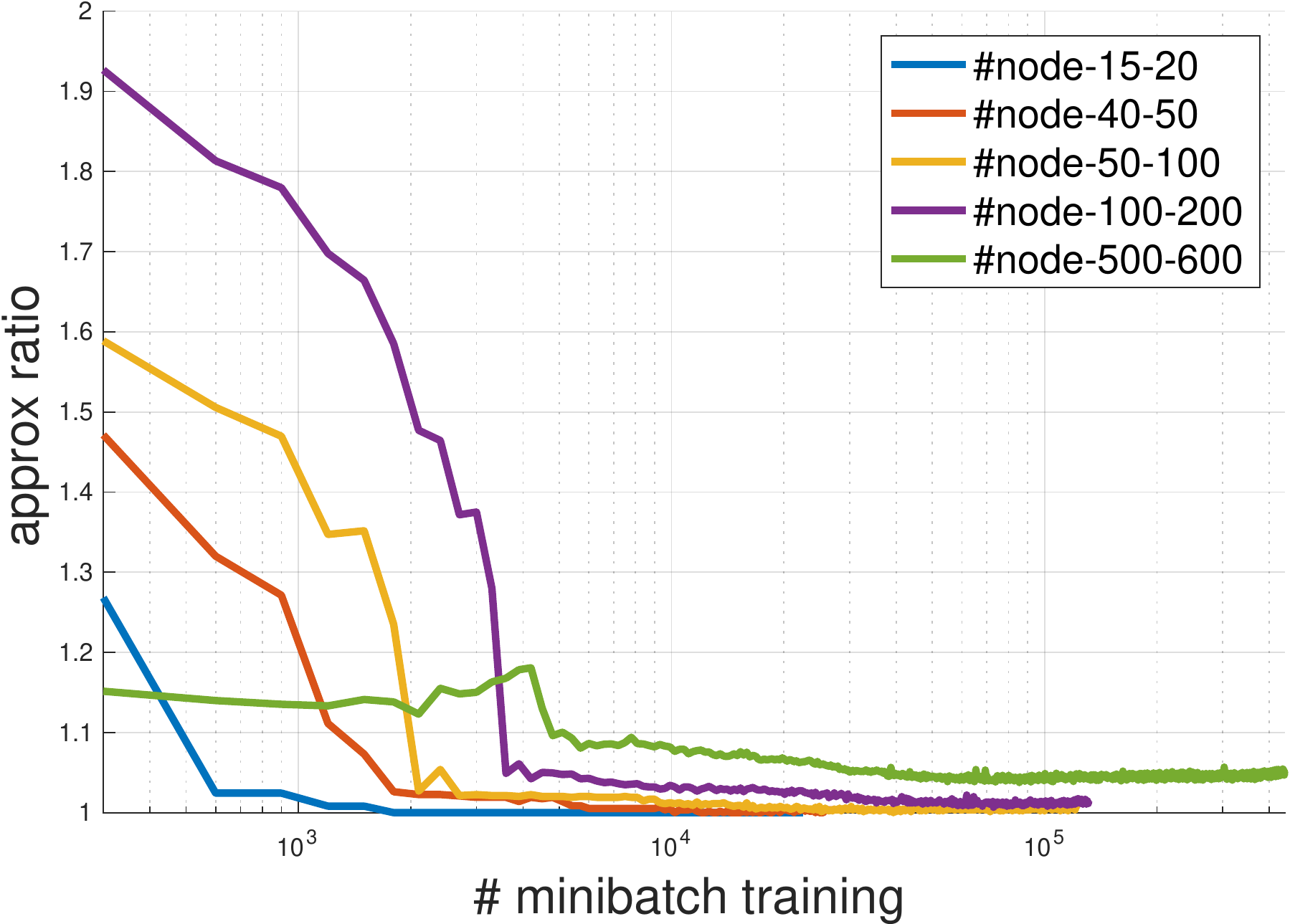} \\
		 (g) SCP 0.1 & (h) SCP 0.05
		%			\\
		%			\includegraphics[width=0.40\textwidth]{curve-tspmult-erdos_renyi} &   
		%			\includegraphics[width=0.40\textwidth]{curve-tspmult-barabasi_albert} \\
		%			(g) GTSP ER & (h) GTSP BA 
		% 
		% 
		%   \includegraphics[width=0.45\textwidth]{curve-mvc-barabasi_albert} & 
		%   \includegraphics[width=0.45\textwidth]{curve-maxcut-erdos_renyi} \\
		%   (a) MVC BA & (b) MAXCUT ER \\ 
		%   \includegraphics[width=0.45\textwidth]{curve-tspmult-erdos_renyi} & 
		%   \includegraphics[width=0.45\textwidth]{curve-tsp2d-random} \\
		%   (c) GTSP ER & (d) TSP2D random 
	\end{tabular}
	% 
	% \begin{subfigure}[t]{0.24\textwidth}
	% \centering
	%   \includegraphics[width=\textwidth]{curve-mvc-barabasi_albert}
	%   \caption{MVC BA \label{fig:comp_embed}}
	% \end{subfigure}
	% \begin{subfigure}[t]{0.24\textwidth}
	% \centering
	%   \includegraphics[width=\textwidth]{curve-maxcut-erdos_renyi}
	%   \caption{Maxcut ER \label{fig:comp_embed}}
	% \end{subfigure}
	% \begin{subfigure}[t]{0.24\textwidth}
	% \centering
	%   \includegraphics[width=\textwidth]{curve-tspmult-erdos_renyi}
	%   \caption{GTSP ER \label{fig:comp_embed}}
	% \end{subfigure}
	% \begin{subfigure}[t]{0.24\textwidth}
	% \centering
	%   \includegraphics[width=\textwidth]{curve-tsp2d-random}
	%   \caption{TSP2D random \label{fig:comp_embed}}
	% \end{subfigure}
	%\vspace{-4mm}
	\caption{S2V-DQN convergence measured by the held-out validation performance.}
	\label{fig:convergence}
	%	\vspace{-4mm}	
\end{figure*}

\subsection{Convergence of S2V-DQN} 
\label{app:convergence}
In Figure~\ref{fig:convergence}, we plot our algorithm's convergence with respect to the held-out validation performance. We first obtain the convergence curve for each type of problem under every graph distribution. To visualize the convergence at the same scale, we plot the approximate ratio. 
% For the convergence on other graphs, please see Appendix~\ref{app:convergence}. 

Figure~\ref{fig:convergence} shows that our algorithm converges nicely on the MVC, MAXCUT and SCP problems. For the MVC, we use the model trained on small graphs to initialize the model for training on larger ones. Since our model also generalizes well to problems with different sizes, the curve looks almost flat. 
For TSP, where the graph is essentially fully connected, it is harder to learn a good model based on graph structure. Nevertheless, as shown in previous section, the graph embedding can still learn good feature representations with multiple embedding iterations.  

\subsection{Complete time v/s approximation ratio plots}
\begin{figure*}[th!]
	\centering
	\setlength{\tabcolsep}{3pt}
	\begin{tabular}{ccc}
		\includegraphics[width=0.32\textwidth]{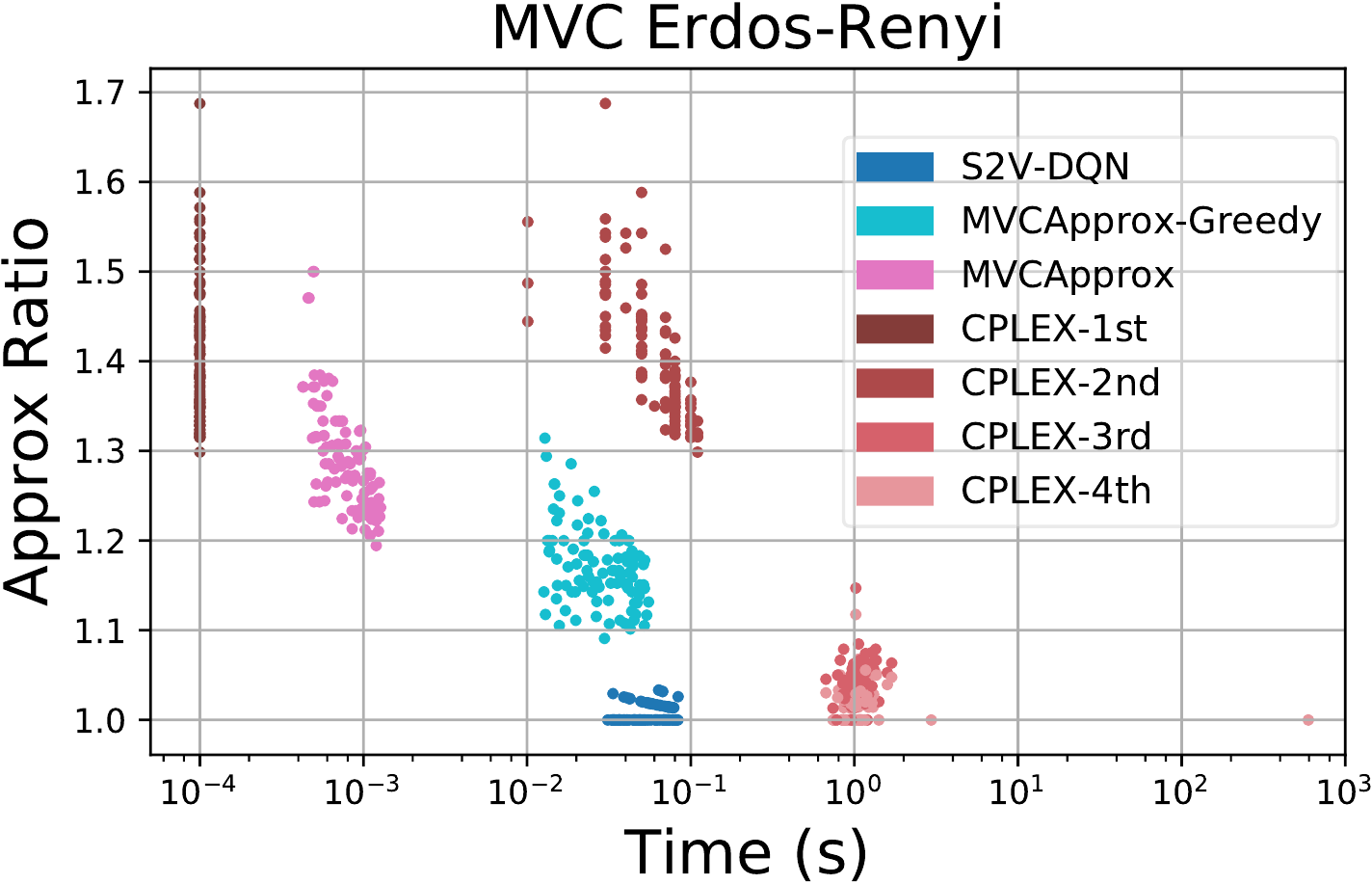} & 
		\includegraphics[width=0.32\textwidth]{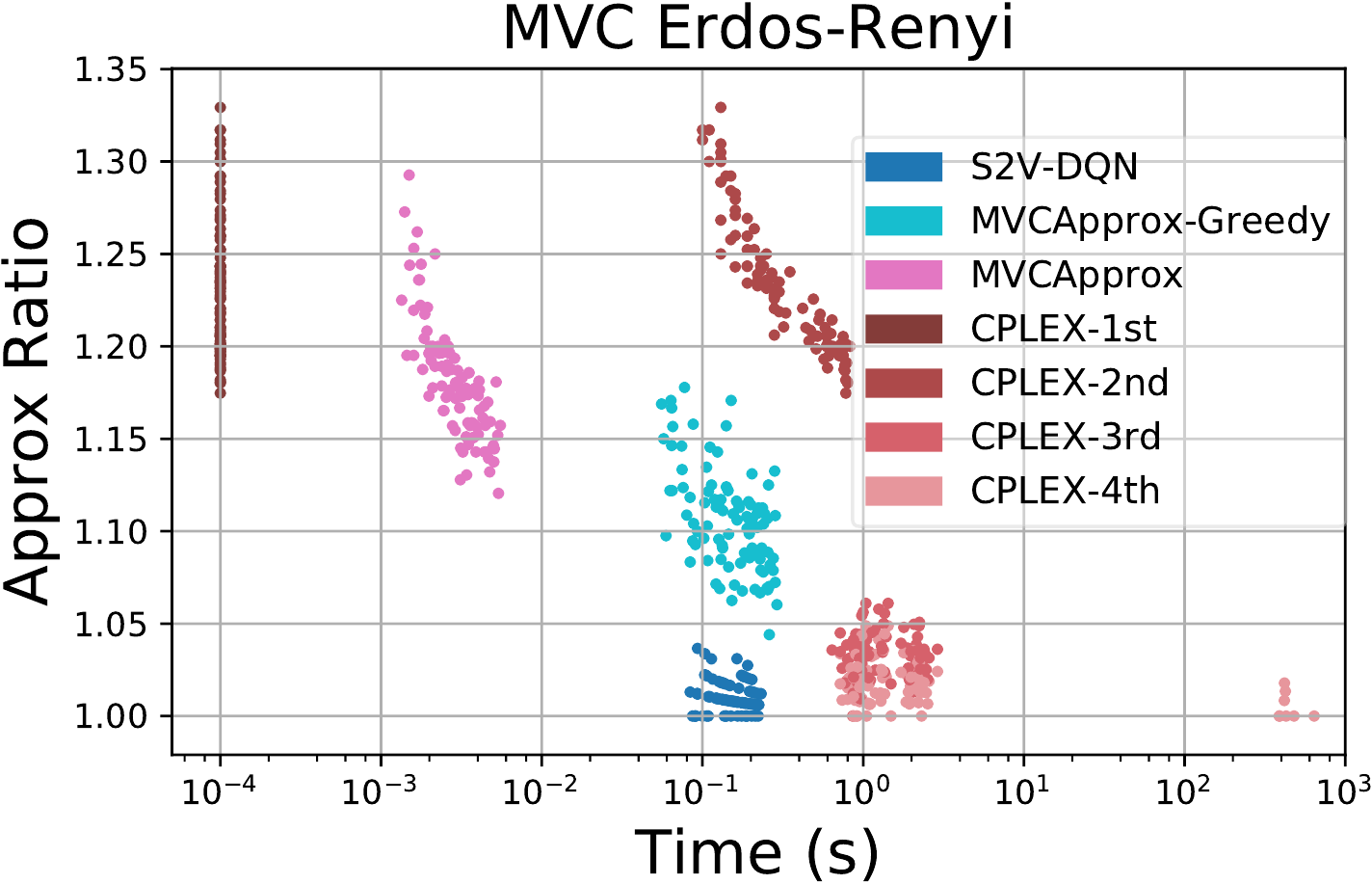} &
		\includegraphics[width=0.32\textwidth]{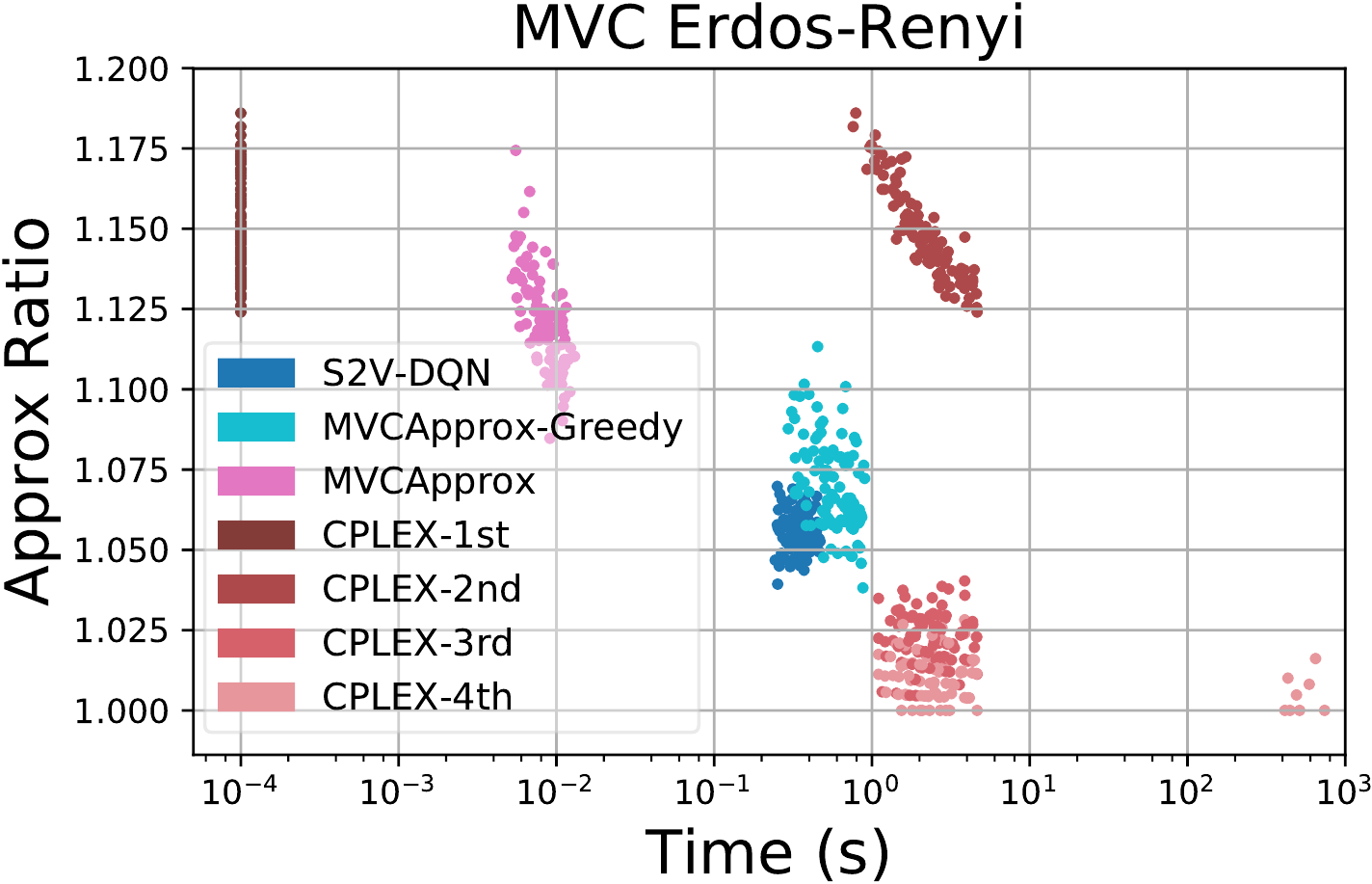}
		\\
		(a) MVC ER 50-100 & (b) MVC ER 100-200 & (c) MVC ER 200-300 \\
		\includegraphics[width=0.32\textwidth]{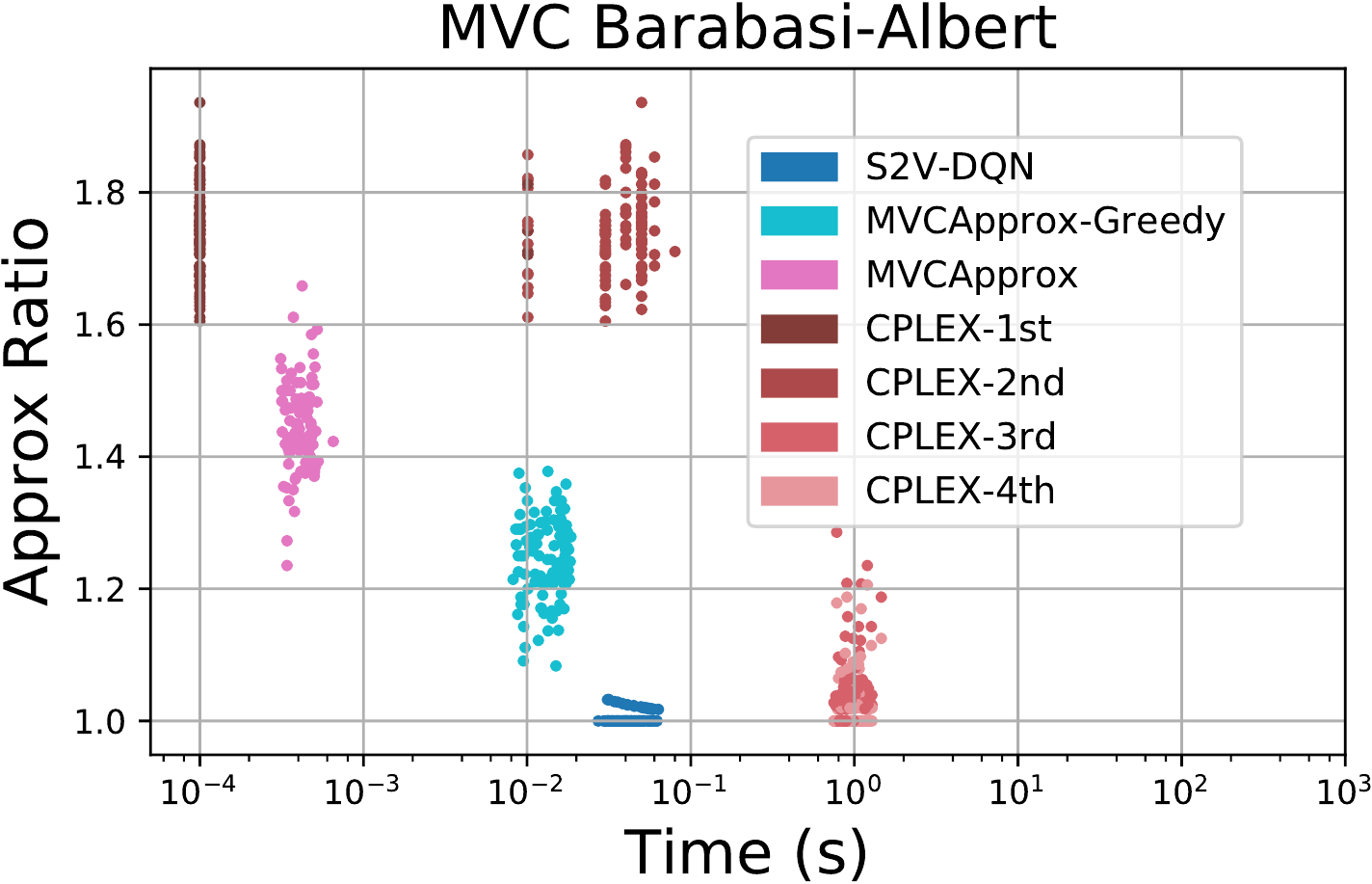} & 
		\includegraphics[width=0.32\textwidth]{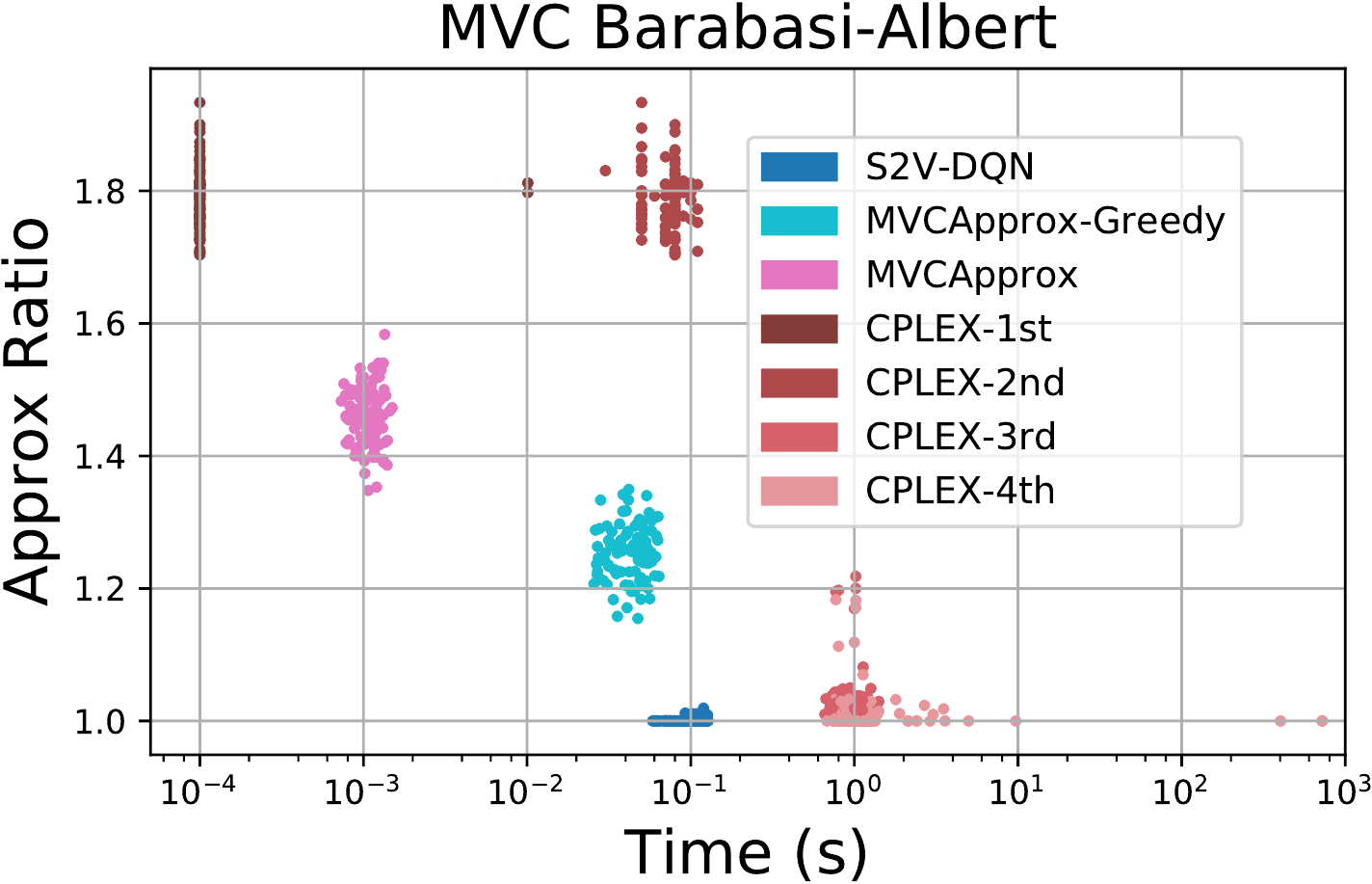} &
		\includegraphics[width=0.32\textwidth]{time-mvc-ba-200-300-crop}
		\\
		(d) MVC BA 50-100 & (e) MVC BA 100-200 & (f) MVC BA 200-300 \\
		\includegraphics[width=0.32\textwidth]{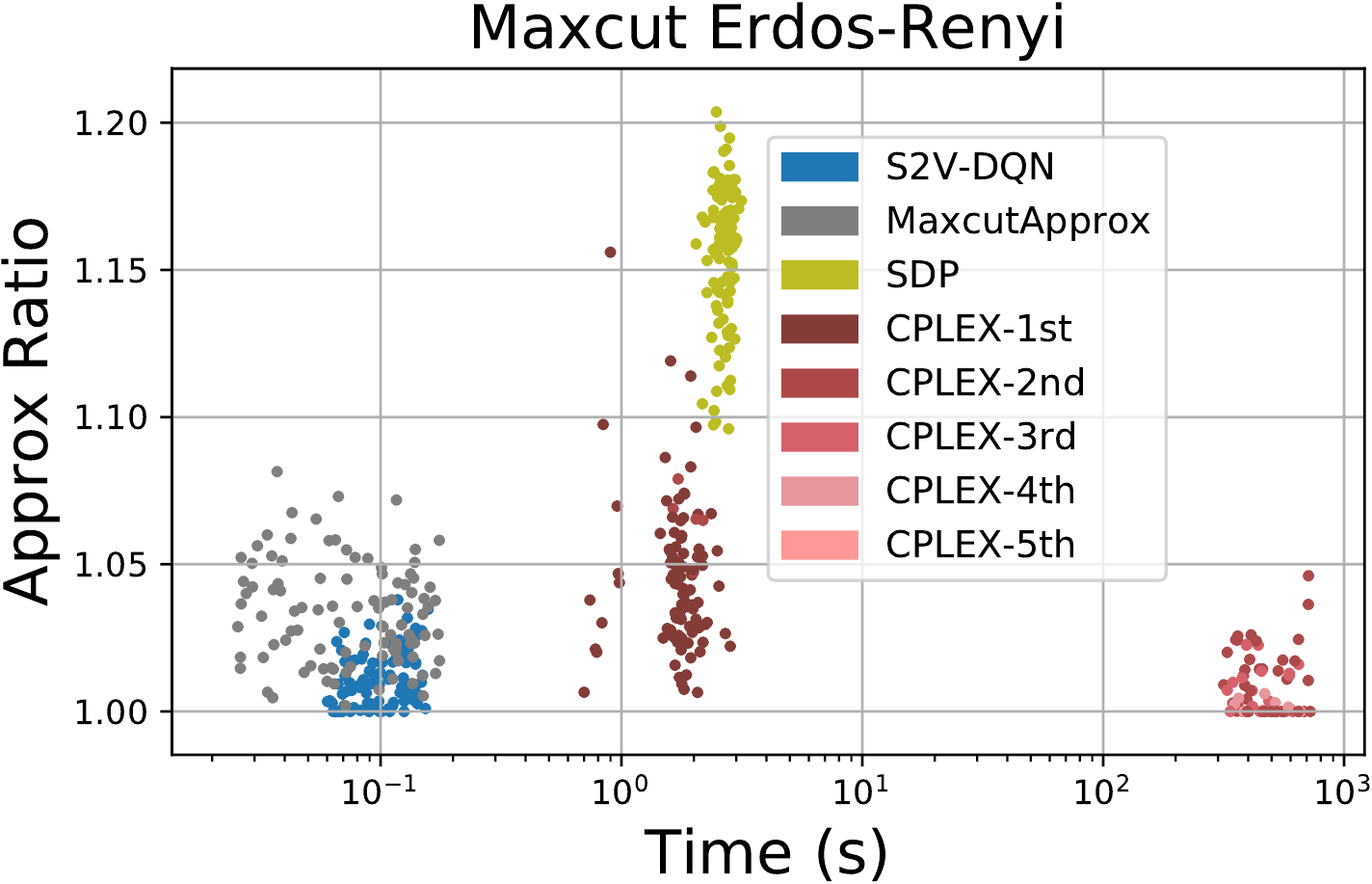} & 
		\includegraphics[width=0.32\textwidth]{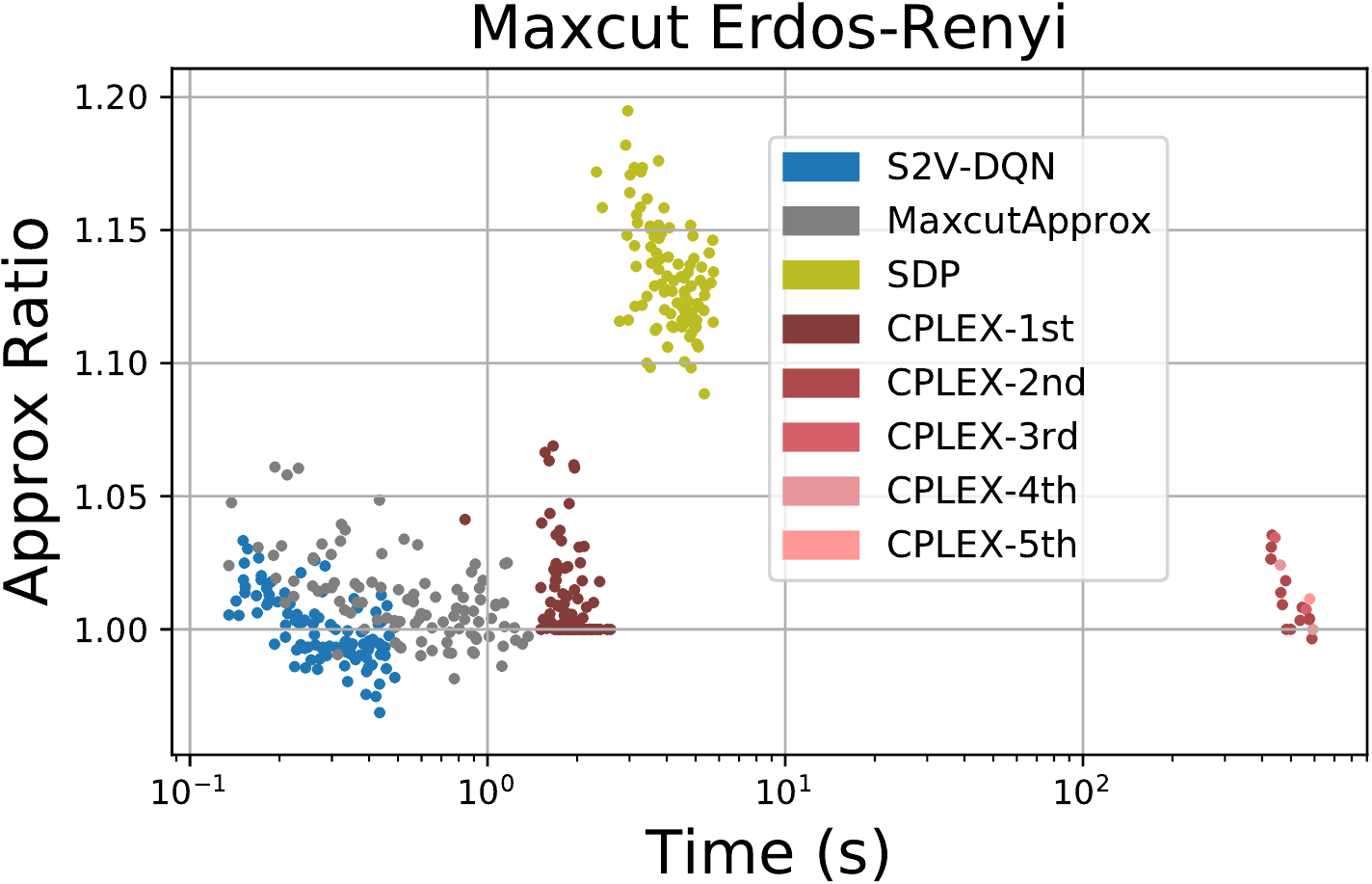} &
		\includegraphics[width=0.32\textwidth]{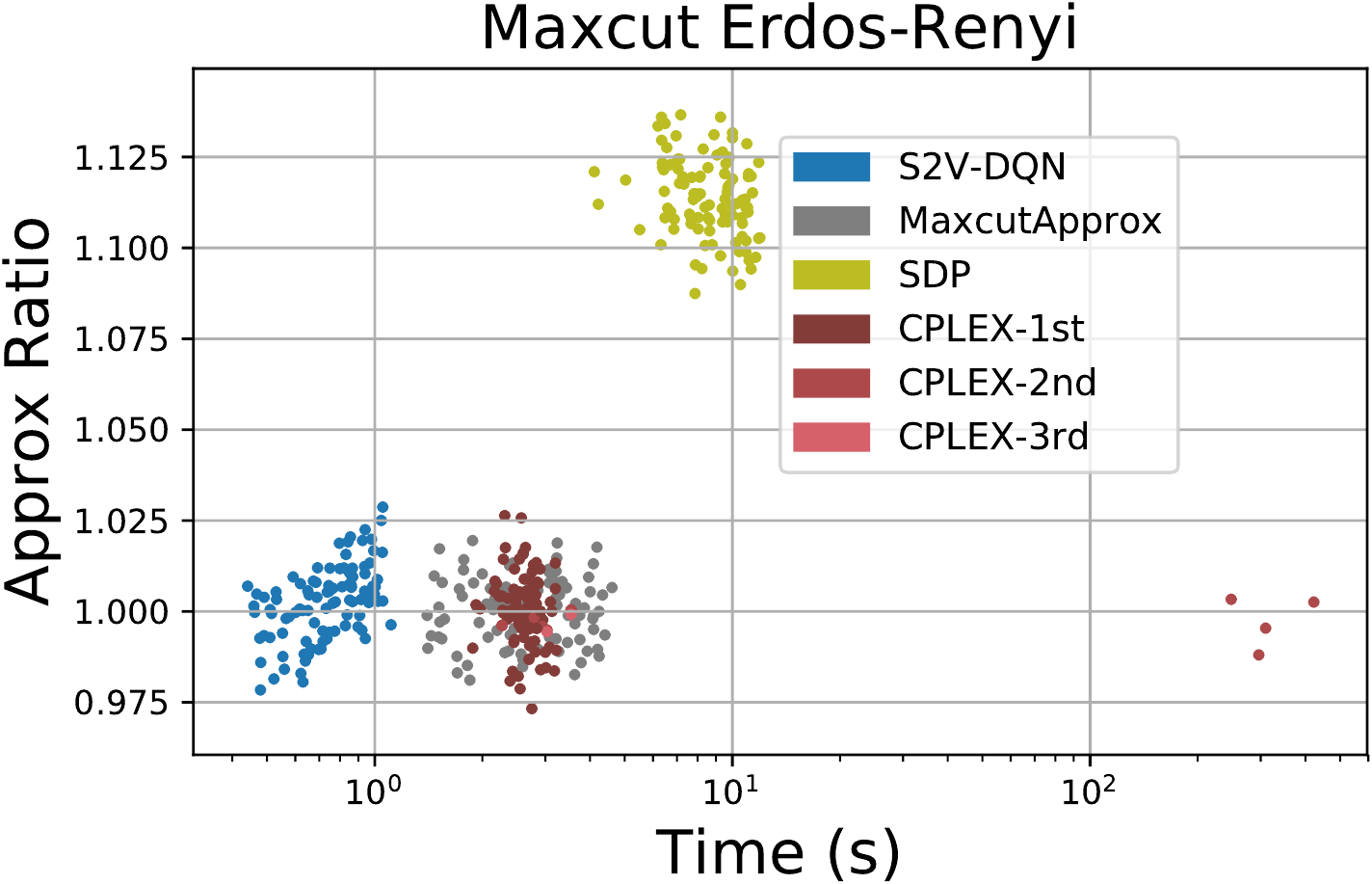}
		\\
		(g) MAXCUT ER 50-100 & (h) MAXCUT ER 100-200 & (i) MAXCUT ER 200-300 \\
		\includegraphics[width=0.32\textwidth]{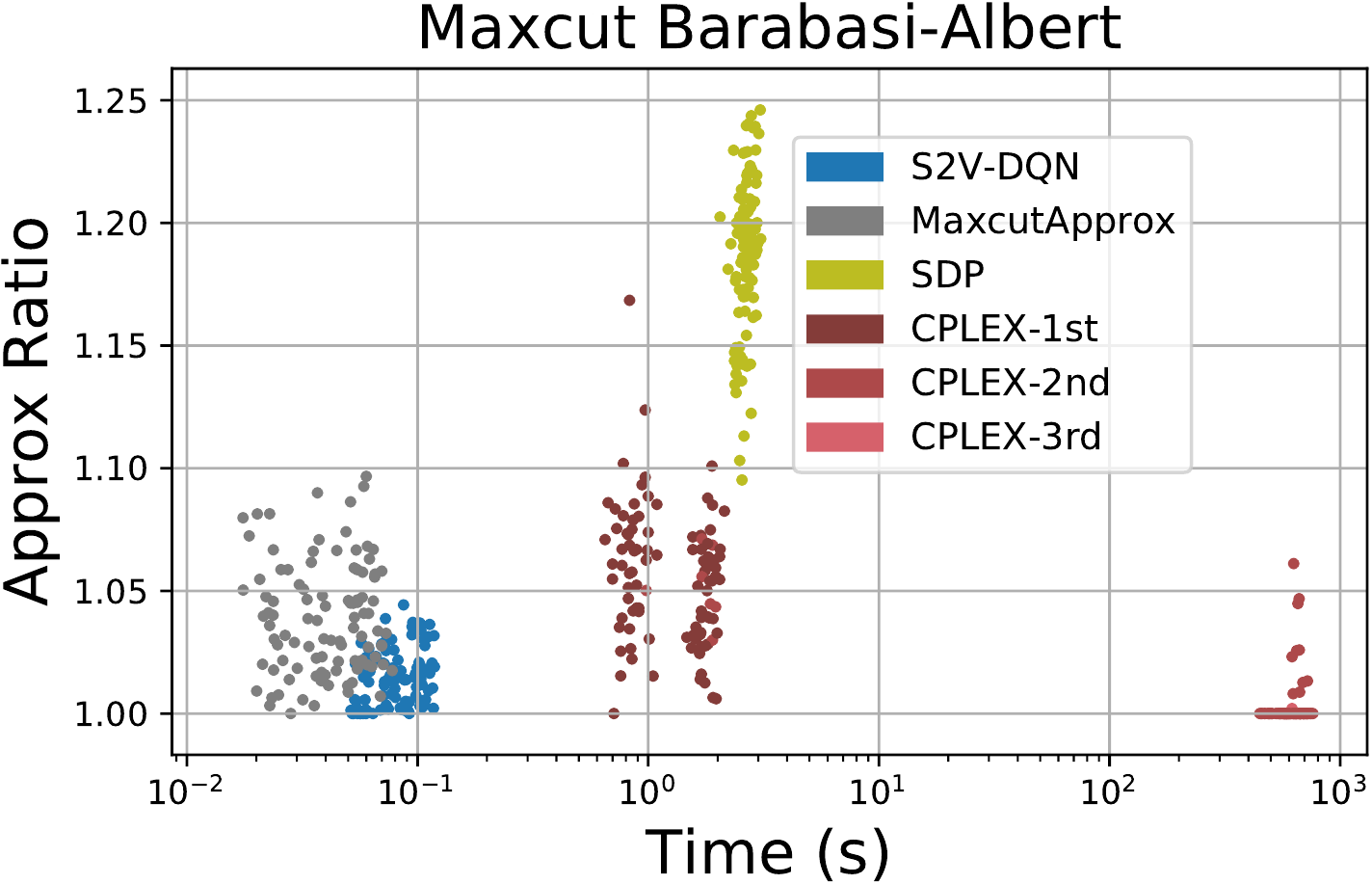} & 
		\includegraphics[width=0.32\textwidth]{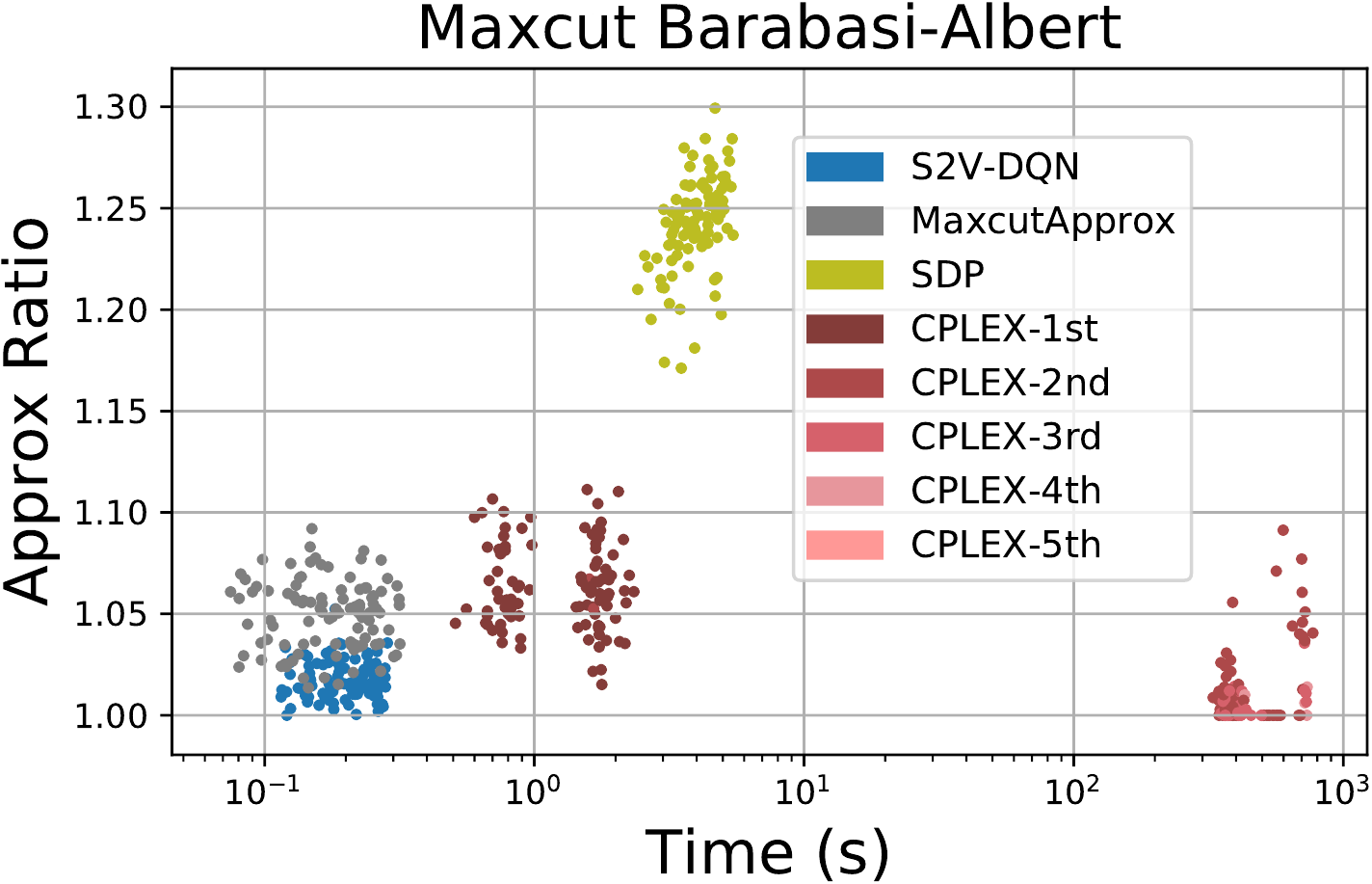} &
		\includegraphics[width=0.32\textwidth]{time-maxcut-ba-200-300-crop}
		\\
		(j) MAXCUT BA 50-100 & (k) MAXCUT BA 100-200 & (l) MAXCUT BA 200-300 \\
	    \includegraphics[width=0.32\textwidth]{{{time-scp-0.05-50-100-crop}}} & 
	    \includegraphics[width=0.32\textwidth]{{{time-scp-0.05-100-200-crop}}} &
	    \includegraphics[width=0.32\textwidth]{{{time-scp-0.05-200-300-crop}}}
	    \\
	    (m) SCP 0.05 50-100 & (n) SCP 0.05 100-200 & (o) SCP 0.05 200-300 \\
		\includegraphics[width=0.32\textwidth]{{{time-scp-0.1-50-100-crop}}} & 
		\includegraphics[width=0.32\textwidth]{{{time-scp-0.1-100-200-crop}}} &
		\includegraphics[width=0.32\textwidth]{{{time-scp-0.1-200-300-crop}}}
		\\
		(p) SCP 0.1 50-100 & (q) SCP 0.1 100-200 & (r) SCP 0.1 200-300 \\

	\end{tabular}
	%  \vspace{-4mm}
	\caption{Time-approximation trade-off for MVC, MAXCUT and SCP. In this figure, each dot represents a solution found for a single problem instance. For CPLEX, we also record the time and quality of each solution it finds. For example, CPLEX-1st means the first feasible solution found by CPLEX.
	}
	\label{fig:tradeoff}
	%  \vspace{-3mm}
\end{figure*}

Figure~\ref{fig:tradeoff} is a superset of Figure~\ref{fig:tradeoff_sub}, including both graph types and three graph size ranges for MVC, MAXCUT and SCP.
\label{app:fulltradeoff}
		
\subsection{Additional analysis of the trade-off between time and approx. ratio}
Tables~\ref{tab:tradeoff_mvc} and~\ref{tab:tradeoff_maxcut} offer another perspective on the trade-off between the running time of a heuristic and the quality of the solution it finds. We ran CPLEX for MVC and MAXCUT for 10 minutes on the 200-300 node graphs, and recorded the time and value of all the solutions found by CPLEX within the limit; results shown next carry over to smaller graphs. Then, for a given method M that terminates in T seconds on a graph G and returns a solution with approximation ratio R, we asked the following 2 questions:

\begin{enumerate}
	\item If CPLEX is given the same amount of time T for G, how well can CPLEX do?
	\item How long does CPLEX need to find a solution of same or better quality than the one the heuristic has found?
\end{enumerate}
For the first question, the column ``Approx. Ratio of Best Solution" in Tables~\ref{tab:tradeoff_mvc} and~\ref{tab:tradeoff_maxcut} shows the following:
\begin{itemize}
	\item[--] MVC (Table~\ref{tab:tradeoff_mvc}): The larger values for S2V-DQN imply that solutions we find quickly are of higher quality, as compared to the MVCApprox/Greedy baselines.
	\item[--] MAXCUT (Table~\ref{tab:tradeoff_maxcut}): On  most of the graphs, CPLEX cannot find any solution at all if given the same time as S2V-DQN or MaxcutApprox. SDP (solved with state-of-the-art CVX solver) is so slow that CPLEX finds solutions that are 10\% better than those of SDP if given the same time as SDP (on ER graphs), which confirms that SDP is not time-efficient. One possible interpretation of the poor performance of SDP is that its theoretical guaranteed of 0.87 is \textit{in expectation} over the solutions it can generate, and so the variance in the approximation ratios of these solutions may be very large.
\end{itemize}
For the second question, the column ``Additional Time Needed" in Tables~\ref{tab:tradeoff_mvc} and~\ref{tab:tradeoff_maxcut} shows the following:
\begin{itemize}
	\item[--] MVC (Table~\ref{tab:tradeoff_mvc}): The larger values for S2V-DQN imply that solutions we find are harder to improve upon, as compared to the MVCApprox/Greedy baselines.
	\item[--] MAXCUT (Table~\ref{tab:tradeoff_maxcut}): On ER (BA) graphs, CPLEX (10 minute-cutoff) cannot find a solution that is better than those of S2V-DQN or MaxcutApprox on many instances (e.g. the value (59) for S2V-DQN on ER graphs means that on $41=100-59$ graphs, CPLEX could not find a solution that is as good as S2V-DQN's). When we consider only those graphs for which CPLEX could find a better solution, S2V-DQN’s solutions take significantly more time for CPLEX to beat, as compared to MaxcutApprox and SDP. The negative values for SDP indicate that CPLEX finds a solution better than SDP’s in a shorter time.
\end{itemize}

% Table generated by Excel2LaTeX from sheet 'Sheet1'
\begin{table}[htbp]
	\centering
	\caption{Minimum Vertex Cover (100 graphs with 200-300 nodes): Trade-off between running time and approximation ratio. An ``Approx. Ratio of Best Solution"  value of 1.x\% means that the solution found by CPLEX if given the same time as a certain heuristic (in the corresponding row) is x\% worse, on average. ``Additional Time Needed" in seconds is the additional amount of time needed by CPLEX to find a solution of value at least as good as the one found by a given heuristic; negative values imply that CPLEX finds such solutions faster than the heuristic does. Larger values are better for both metrics. The values in parantheses are the number of instances (out of 100) for which CPLEX finds some solution in the given time (for ``Approx. Ratio of Best Solution"), or finds some solution that is at least as good as the heuristic's (for ``Additional Time Needed").}
	\begin{tabular}{lcc|cc}
		\toprule
		& \multicolumn{2}{c|}{Approx. Ratio of Best Solution} & \multicolumn{2}{c}{Additional Time Needed} \\
		& ER    & BA    & ER    & BA \\
		\midrule
		S2V-DQN & 1.09 (100) & 1.81 (100) & 2.14 (100) & 137.42 (100) \\
		\midrule
		MVCApprox-Greedy & 1.07 (100) & 1.44 (100) & 1.92 (100) & 0.83 (100) \\
		\midrule
		MVCApprox   & 1.03 (100) & 1.24 (98) & 2.49 (100) & 0.92 (100) \\
		\bottomrule
	\end{tabular}%
	\label{tab:tradeoff_mvc}%
\end{table}%

% Table generated by Excel2LaTeX from sheet 'Sheet1'
\begin{table}[htbp]
	\centering
	\caption{Maximum Cut (100 graphs with 200-300 nodes): please refer to the caption of Table~\ref{tab:tradeoff_mvc}.}
	\begin{tabular}{lcc|cc}
		\toprule
		& \multicolumn{2}{c|}{Approx. Ratio of Best Solution} & \multicolumn{2}{c}{Additional Time Needed} \\
		& ER    & BA    & ER    & BA \\
		\midrule
		S2V-DQN & N/A (0) & 1081.45 (1) & 8.99 (59) & 402.05 (34) \\
		\midrule
		MaxcutApprox\;\;\;\;\;\;\;\;\; & 1.00 (48) & 340.11 (3) & -0.23 (50) & 218.19 (57) \\
		\midrule
		SDP   & 0.90 (100) & 0.84 (100) & -6.06 (100) & -5.54 (100) \\
		\bottomrule
	\end{tabular}%
	\label{tab:tradeoff_maxcut}%
\end{table}%
		
	\subsection{Visualization of solutions}

\begin{figure*}[th!]
	\centering
	\setlength{\tabcolsep}{3pt}
	\begin{tabular}{cc}
		\includegraphics[width=1.0\textwidth]{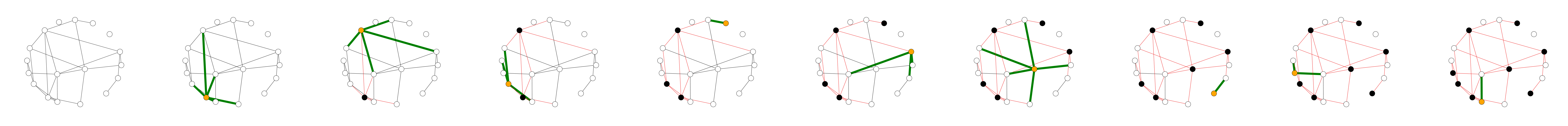} \\ 
		\includegraphics[width=1.0\textwidth]{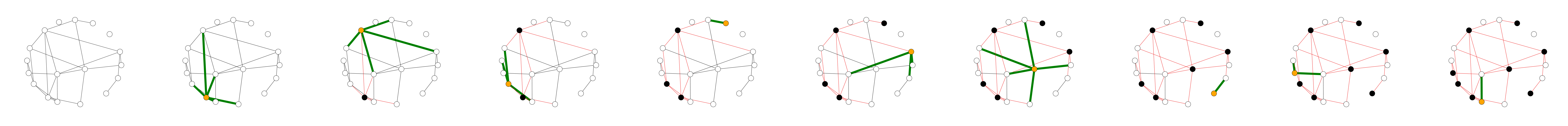}
	\end{tabular}   
	%  \vspace{-4mm}
	\caption{Minimum Vertex Cover: an optimal solution to an ER graph instance found by S2V-DQN. Selected node in each step is colored in orange, and nodes in the partial solution up to that iteration are colored in black. Newly covered edges are in thick green, previously covered edges are in red, and uncovered edges in black. We show that the agent is not only picking the node with large degree, but also trying to maintain the connectivity after removal of the covered edges. For more detailed analysis, please see Appendix~\ref{app:mvc_graphviz}. 
	}
	\label{fig:viz_mvc}
	%  \vspace{-3mm}
\end{figure*}

\begin{figure*}[th!]
	\centering
	\setlength{\tabcolsep}{3pt}
	\begin{tabular}{cc}
		\includegraphics[width=1.0\textwidth]{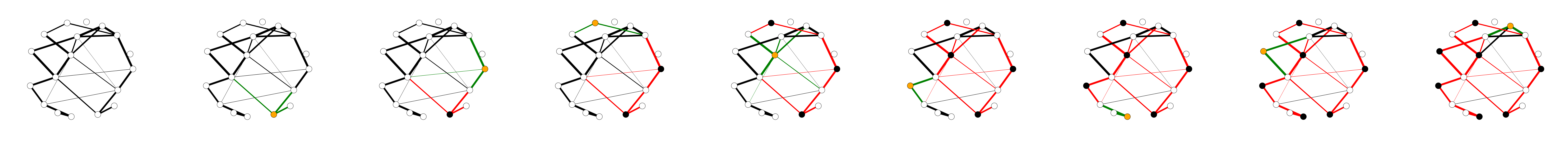} \\ 
		\includegraphics[width=0.8\textwidth]{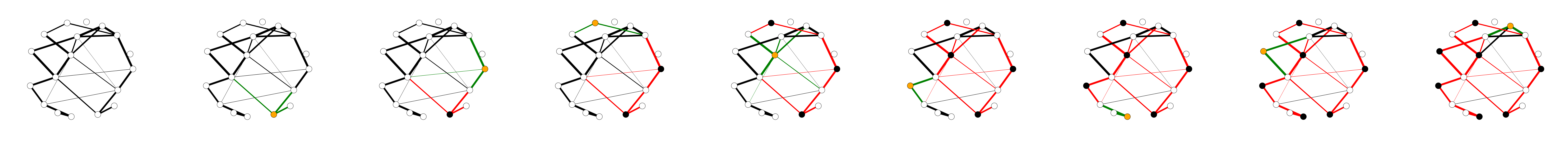}
	\end{tabular}   
	%  \vspace{-4mm}
	\caption{Maximum Cut: an optimal solution to ER graph instance found by S2V-DQN. Nodes are partitioned into two sets: white or black nodes. At each iteration, the node selected to join the set of black nodes is highlighted in orange, and the new cut edges it produces are in green. Cut edges from previous iteration are in red (Best viewed in color). It seems the agent will try to involve the nodes that won't cancel out the edges in current cut set. 
	}
	\label{fig:viz_maxcut}
	%  \vspace{-3mm}
\end{figure*}

\begin{figure*}[th!]
	\centering
	\setlength{\tabcolsep}{3pt}
	\begin{tabular}{c|c}
		\includegraphics[width=.4\textwidth]{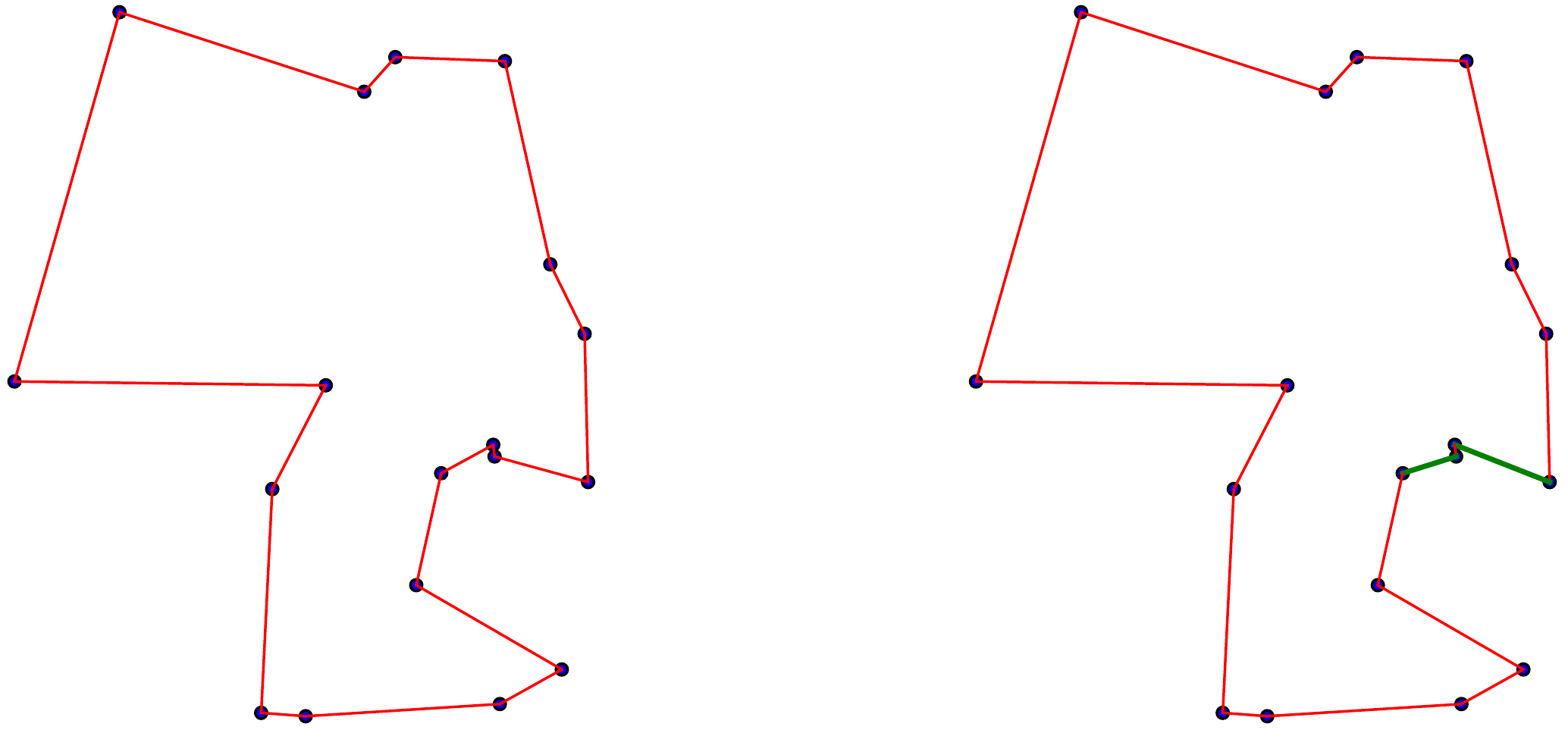} &
		\includegraphics[width=.4\textwidth]{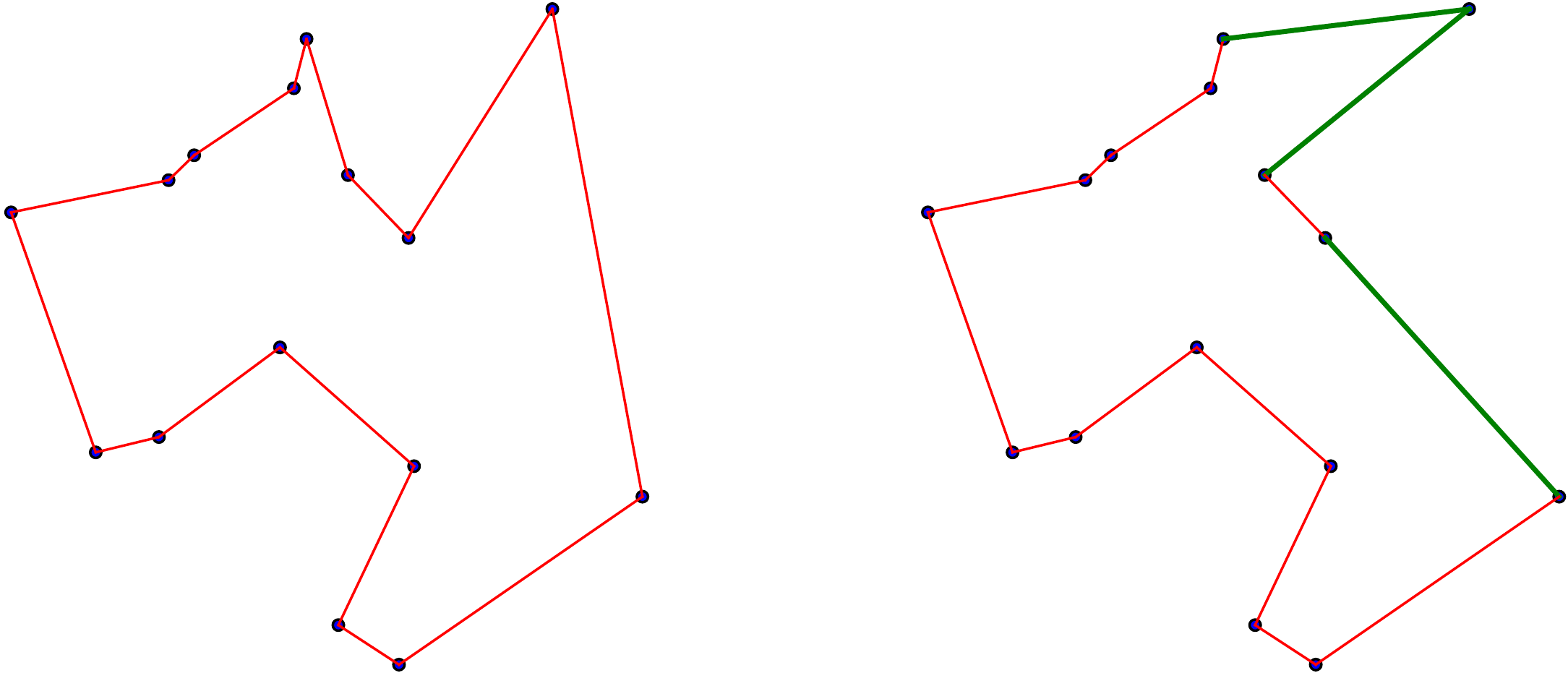}
	\end{tabular}   
	%  \vspace{-4mm}
	\caption{Traveling Salesman Problem. Left: optimal tour to a ``random" instance with 18 points (all edges are red), compared to a tour found by our method next to it. For our tour, edges that are not in the optimal tour are shown in green. Our tour is $0.07\%$ longer than an optimal tour. Right: a ``clustered" instance with 15 points; same color coding as left figure. Our tour is $0.5\%$ longer than an optimal tour. (Best viewed in color). 
		%%tsp2d_random_1: 1: ours=3727284, opt=3724625
		%%tsp2d_clustered_10: 10: ours=3742521, opt=3724010
	}
	\label{fig:viz_tsp2d}
	%  \vspace{-3mm}
\end{figure*}

In Figure~\ref{fig:viz_mvc}, \ref{fig:viz_maxcut} and \ref{fig:viz_tsp2d}, we visualize solutions found by our algorithm for MVC, MAXCUT and TSP problems, respectively. For the ease of presentation, we only visualize small-size graphs. For MVC and MAXCUT, the graph is of the ER type and has 18 nodes. For TSP, we show solutions for a ``random" instance (18 points) and a ``clustered" one (15 points).

For MVC and MAXCUT, we show two step by step examples where S2V-DQN finds the optimal solution. For MVC, it seems we are picking the node which covers the most edges in the current state. However, in a more detailed visualization in Appendix~\ref{app:mvc_graphviz}, we show that our algorithm learns a smarter greedy or dynamic programming like strategy. While picking the nodes, it also learns how to keep the connectivity of graph by scarifying the intermediate edge coverage a little bit.

In the example of MAXCUT, it is even more interesting to see that the algorithm did not pick the node which gives the largest intermediate reward at the beginning. Also in the intermediate steps, the agent seldom chooses a node which would cancel out the edges that are already in the cut set. This also shows the effectiveness of graph state representation, which provides useful information to support the agent's node selection decisions. For TSP, we visualize an optimal tour and one found by S2V-DQN for two instances. While the tours found by S2V-DQN differ slightly from the optimal solutions visualized, they are of comparable cost and look qualitatively acceptable. The cost of the tours found by S2V-DQN is within $0.07\%$ and $0.5\%$ of optimum, respectively.

		\subsection{Detailed visualization of learned MVC strategy}
		\label{app:mvc_graphviz}
	
	\begin{figure}
		\centering
		\setlength{\tabcolsep}{3pt}
		\begin{tabular}{c|c}
			\includegraphics[width=0.45\textwidth]{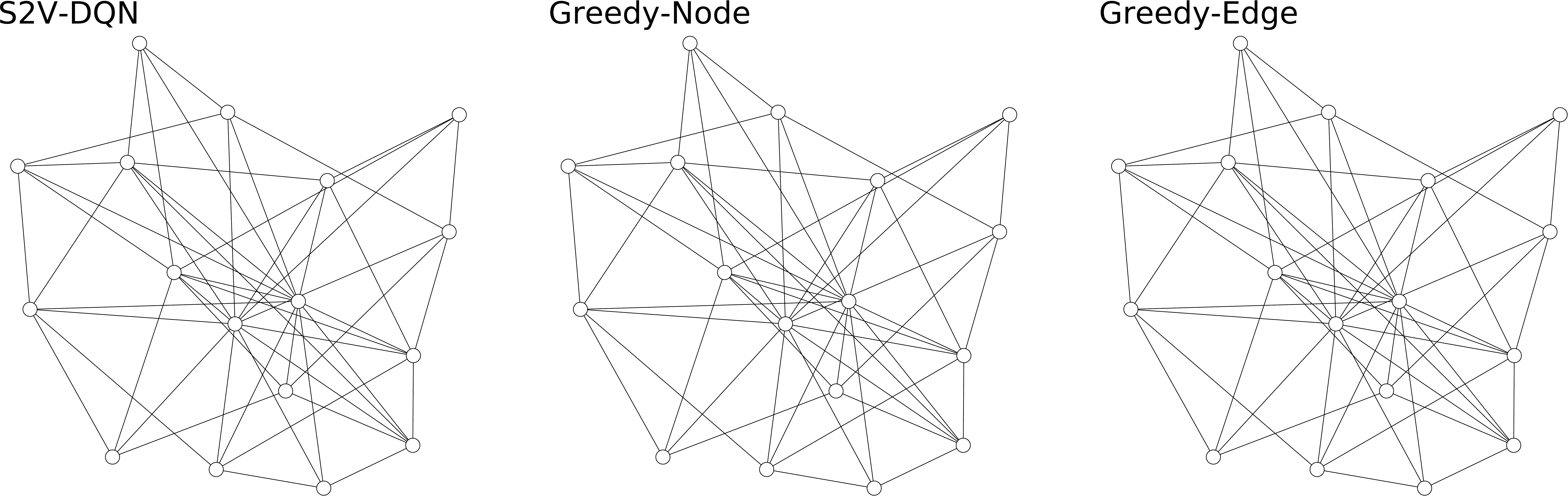} & 
			\includegraphics[width=0.45\textwidth]{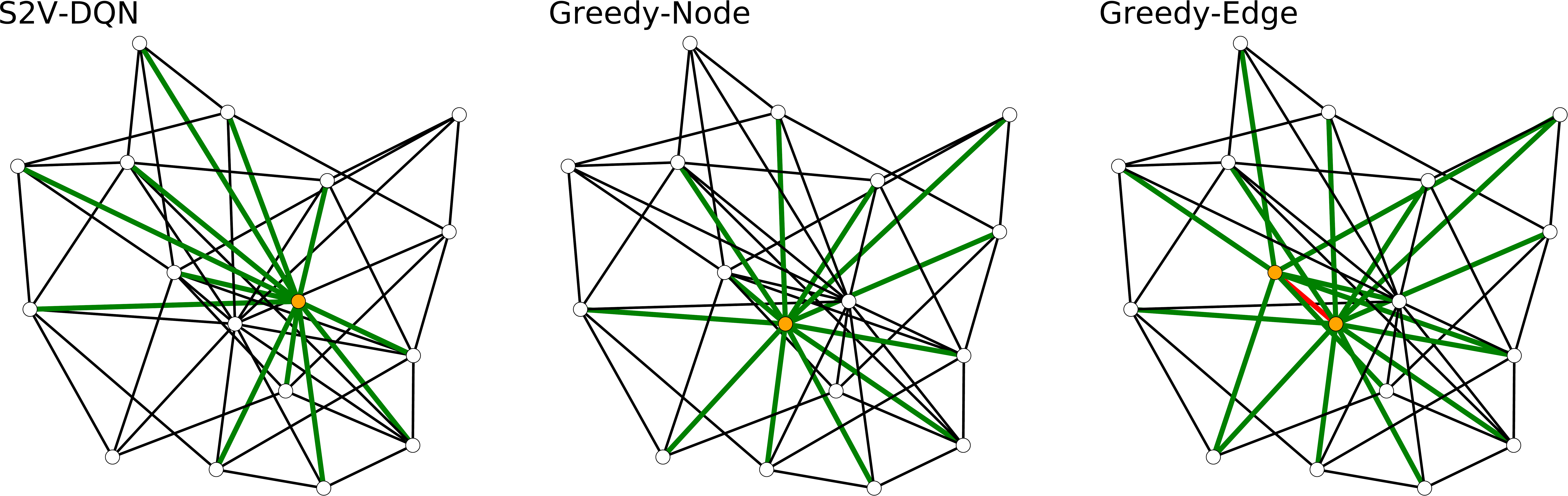} \\
			step (0) & step (1) \\
			\includegraphics[width=0.45\textwidth]{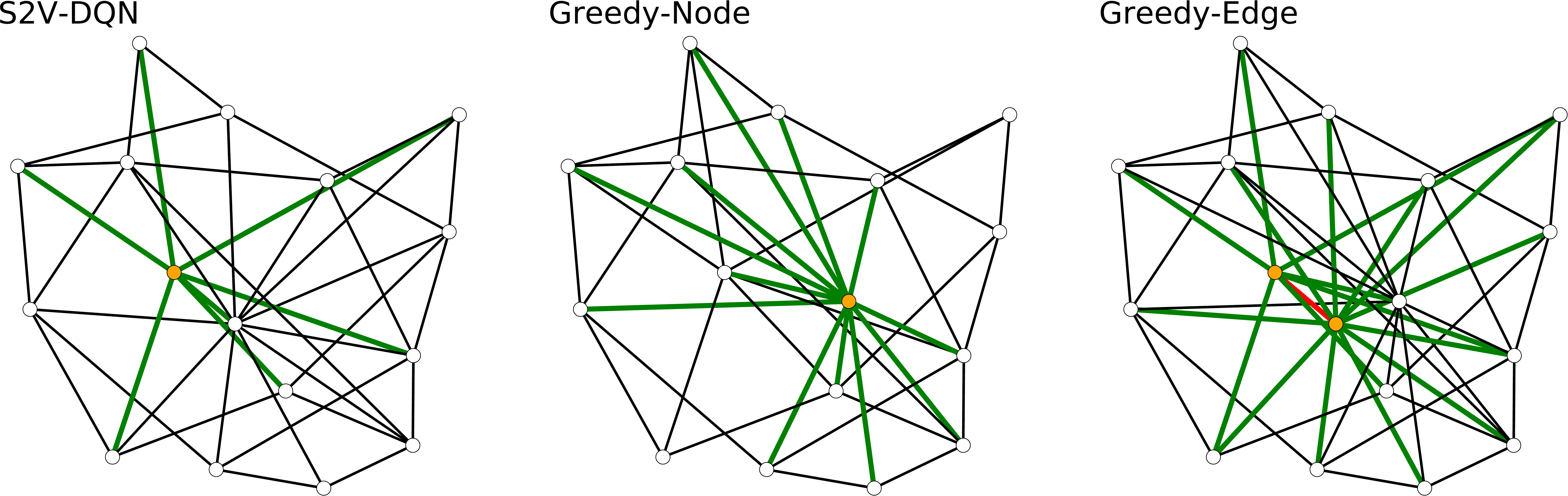} & 
			\includegraphics[width=0.45\textwidth]{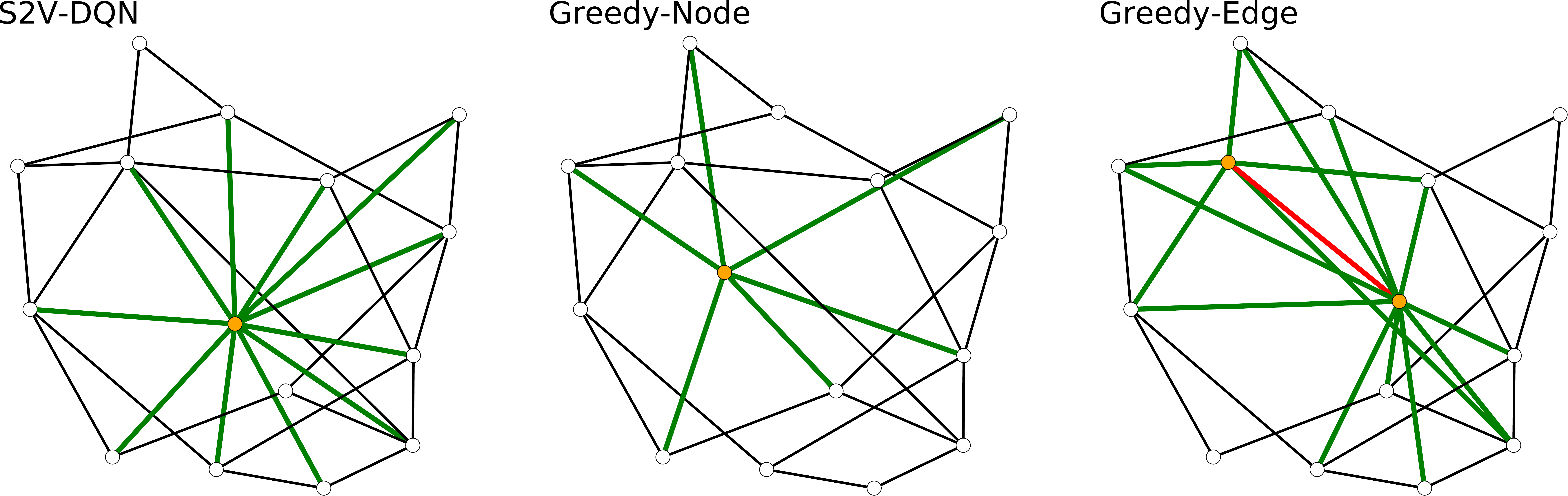} \\
			step (2) & step (3) \\
			\includegraphics[width=0.45\textwidth]{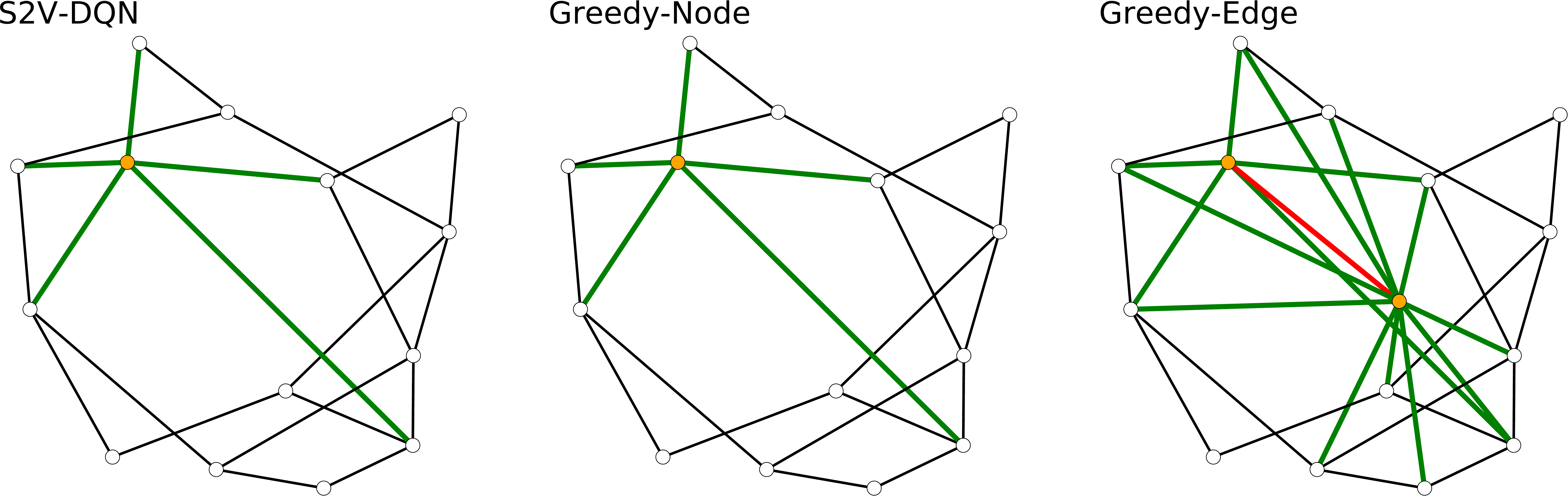} & 
			\includegraphics[width=0.45\textwidth]{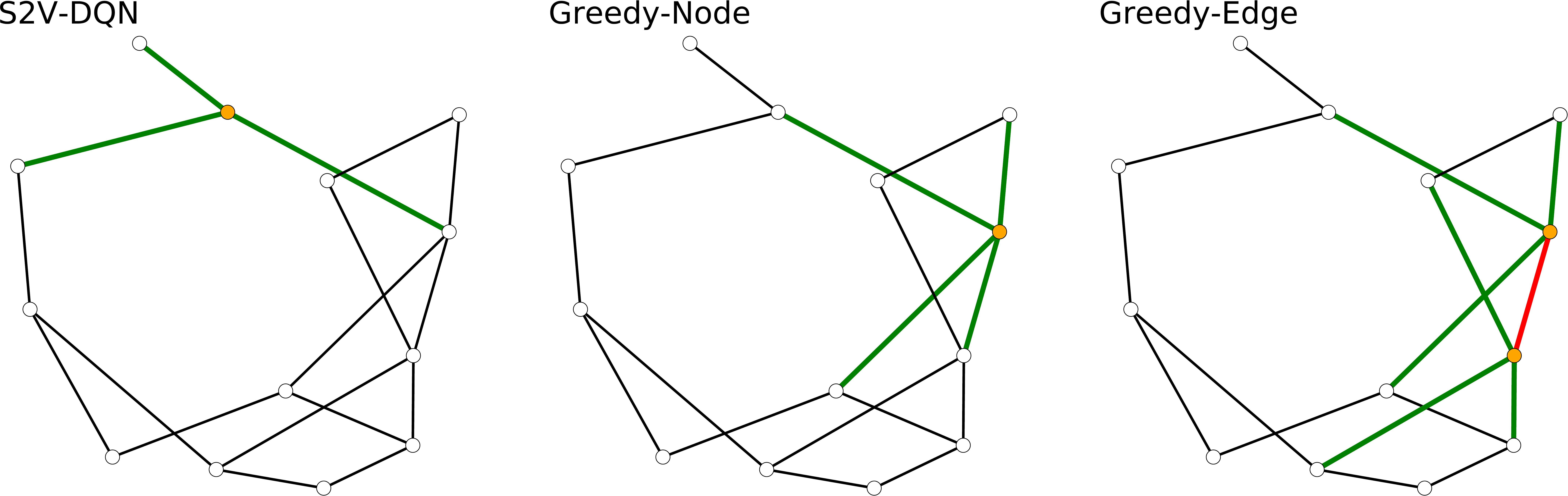} \\
			step (4) & step (5) \\
			\includegraphics[width=0.45\textwidth]{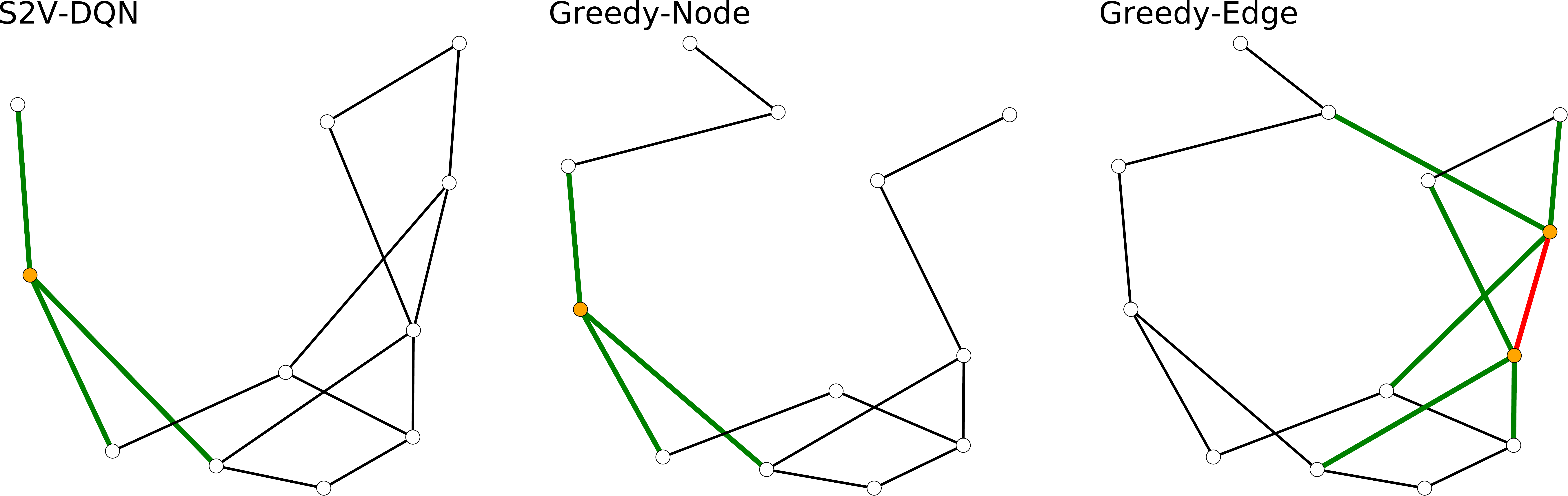} & 
			\includegraphics[width=0.45\textwidth]{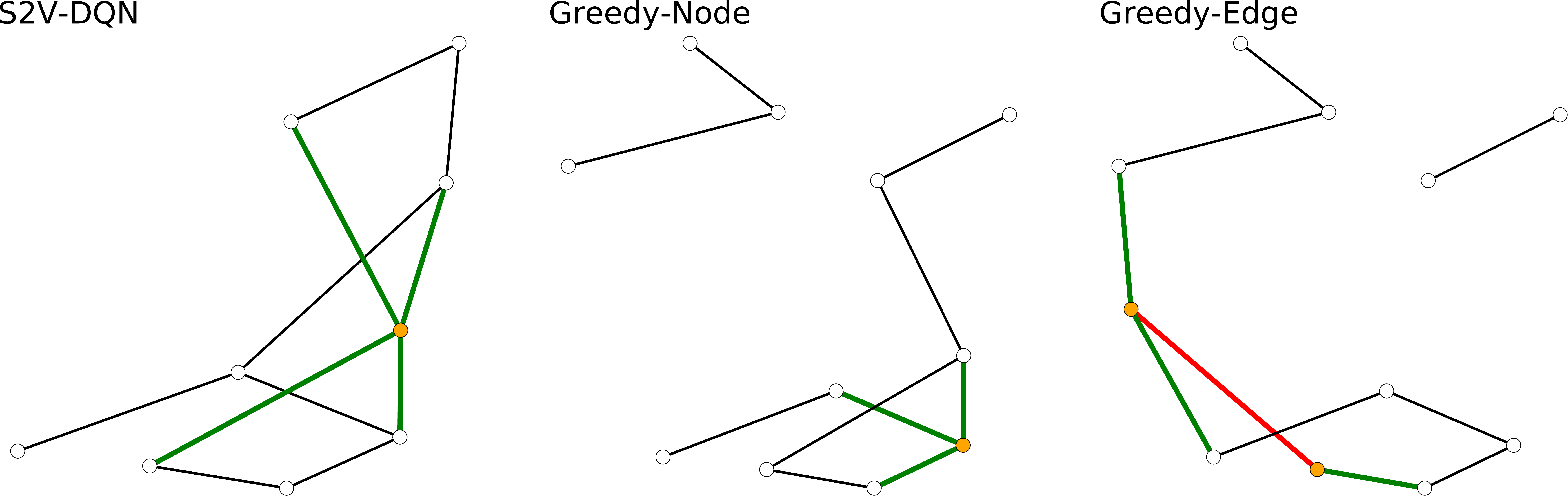} \\
			step (6) & step (7) \\
			\includegraphics[width=0.45\textwidth]{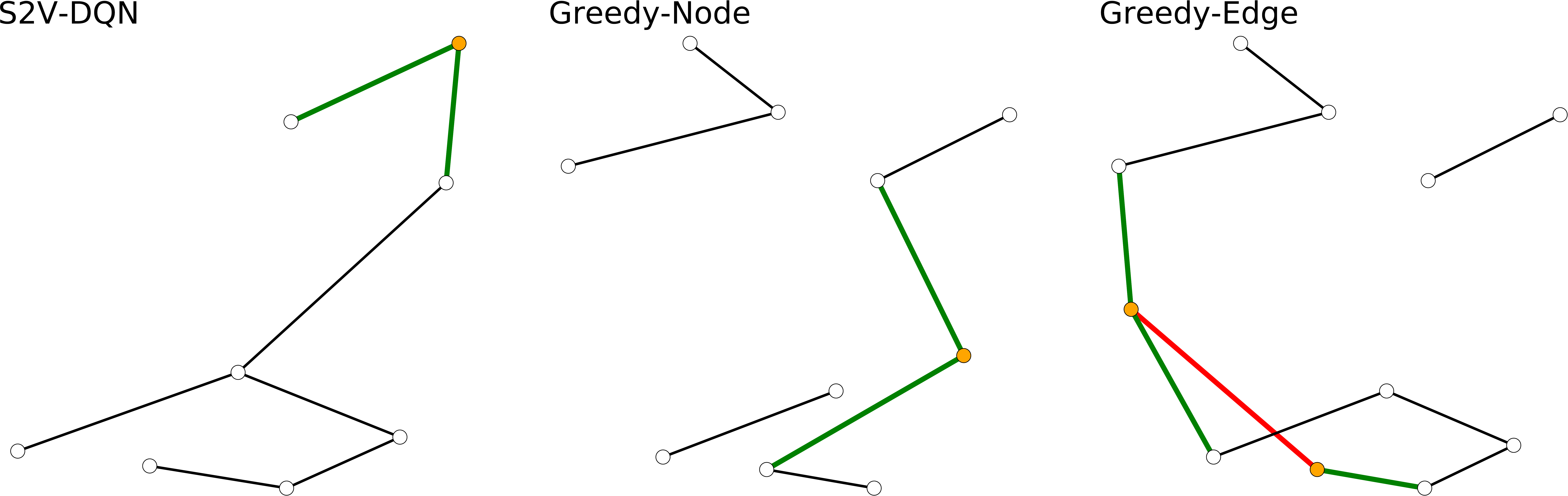} & 
			\includegraphics[width=0.45\textwidth]{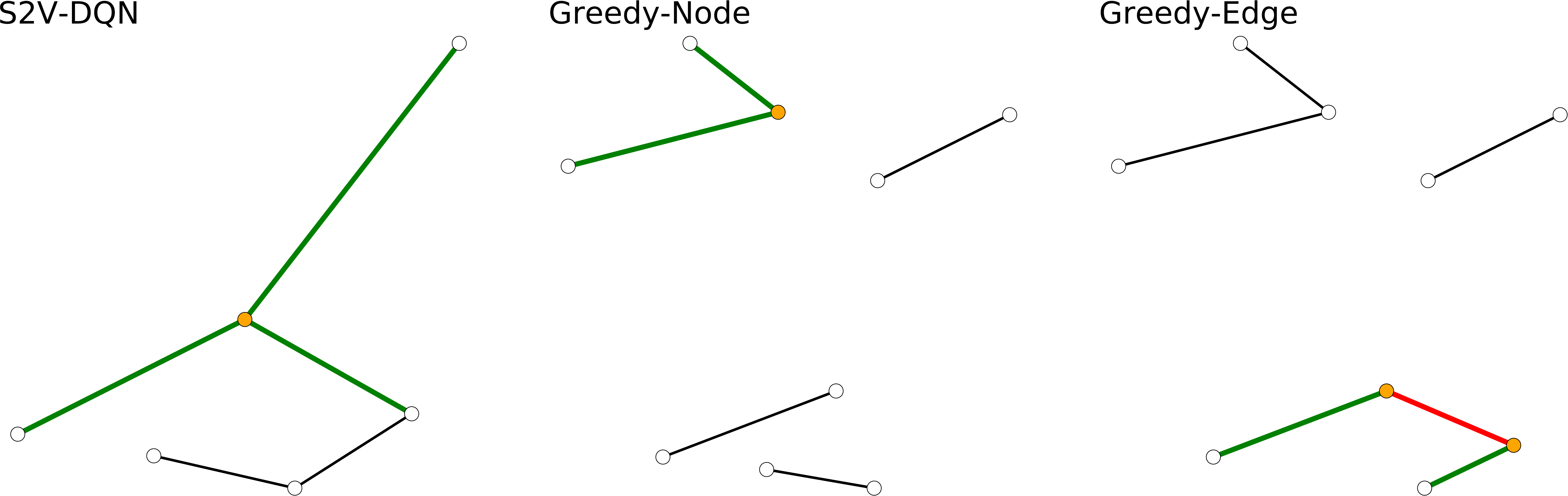} \\
			step (8) & step (9) \\
			\includegraphics[width=0.45\textwidth]{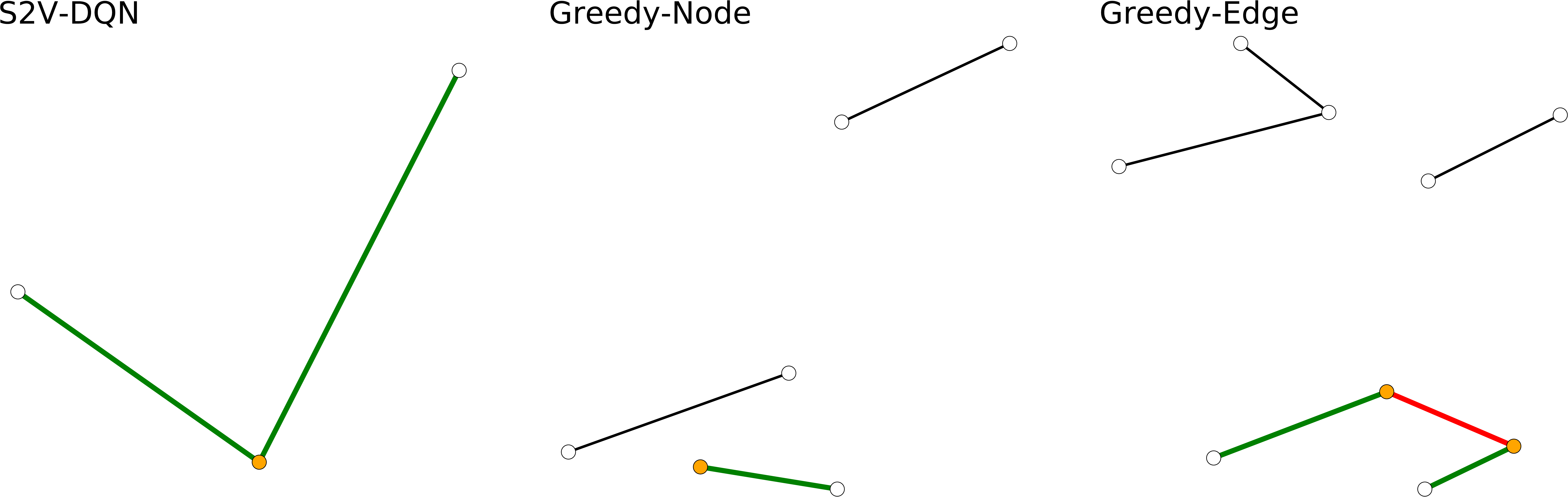} & 
			\includegraphics[width=0.45\textwidth]{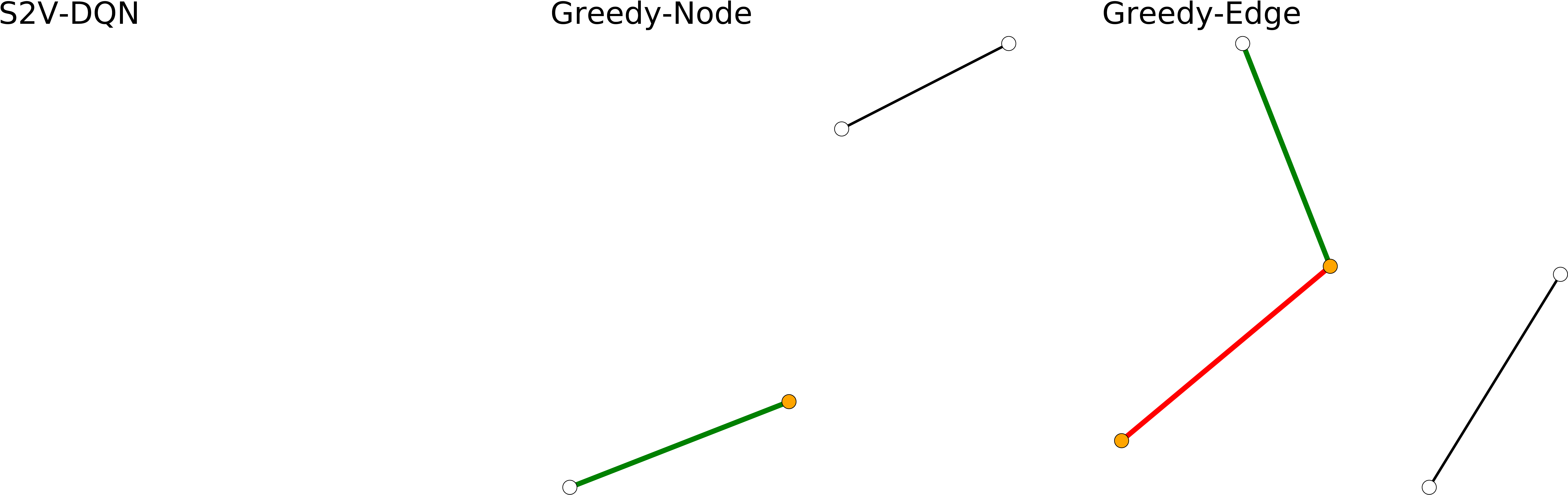} \\
			step (10) & step (11) \\
		\end{tabular}
		\caption{Step-by-step comparison between our S2V-DQN and two greedy heuristics. We can see our algorithm will also favor the large degree nodes, but it will also do something smartly: instead of breaking the graph into several disjoint components, our algorithm will try the best to keep the graph connected. }
		\label{fig:mvc_comp_viz}
	\end{figure}
	
		In Figure~\ref{fig:mvc_comp_viz}, we show a detailed comparison with our learned strategy and two other simple heuristics. We find that the S2V-DQN can learn a much smarter strategy, where the agent is trying to maintain the connectivity of graph during node picking and edge removal. 
		
		\subsection{Experiment Configuration of PN-AC}
		
		We implemented PN-AC to the best of our capabilities. Note that it is quite possible that there are minor differences between our implementation and ~\citet{BelPhaLeNoretal16} that might have resulted in performance not as good as reported in that paper.
		
		For experiments of PN-AC across all tasks, we follow the configurations provided in~\cite{BelPhaLeNoretal16}:
		\begin{inlineenum}
			\item For the input data, we use mini-batches of 128 sequences with 0-paddings to the maximal input length (which is the maximal number of nodes) in the training data.
			\item For node representation, we use coordinates for TSP, so the input dimension is 2. For MVC, MAXCUT and SCP, we represent nodes based on the adjacency matrix of the graph. To get a fixed dimension representation for each node, we use SVD to get a low-rank approximation of the adjacency matrix. We set the rank as 8, so that each node in the input sequence is represented by a 8-dimensional vector.
			\item For the network structure, we use standard single-layer LSTM cells with 128 hidden units for both encoder and decoder parts of the pointer networks.
			\item For the optimization method, we train the PN-AC model with the Adam optimizer~\citep{KinBa14} and use an initial learning rate of $10^{-3}$ that decay every 5000 steps by a factor of 0.96.
			\item For the glimpse trick, we exactly use one-time glimpse in our implementation, as described in the original PN-AC paper.
			\item We initialize all the model parameters uniformly random within $[-0.08, 0.08]$ and clip the $L2$ norm of the gradients to 1.0.
			\item For the baseline function in the actor-critic algorithm, we tried the critic network in our implementation, but it hurts the performance according to our experiments. So we use the exponential moving average performance of the sampled solution from the pointer network as the baseline.
		\end{inlineenum}
		
		\textbf{Consistency with the results from ~\citet{BelPhaLeNoretal16}} Though our TSP experiment setting is not exactly the same as ~\citet{BelPhaLeNoretal16}, we still include some of the results directly here, for the sake of completeness. We applied the insertion heuristic to PN-AC as well, and all the results reported in our paper are with the insertion heuristic. We compare the approximation ratio reported by~\citet{BelPhaLeNoretal16} verses which reported by our implementation. For TSP20: 1.02 vs 1.03 (reported in our paper); TSP50: 1.05 vs 1.07 (reported in our paper); TSP100: 1.07 vs 1.09 (reported in our paper). Note that we have variable graph size in each setting (where the original PN-AC is only reported on fixed graph size), which makes the task more difficult. Therefore, we think the performance gap here is pretty reasonable.

		% \subsection{Convergence on other graphs}
		% \label{app:convergence}
		% 
		% \begin{figure*}[t]
		% \centering
		% \begin{subfigure}[t]{0.24\textwidth}
		% \centering
		%   \includegraphics[width=\textwidth]{curve-mvc-erdos_renyi}
		%   \caption{MVC ER \label{fig:comp_embed}}
		% \end{subfigure}
		% \begin{subfigure}[t]{0.24\textwidth}
		% \centering
		%   \includegraphics[width=\textwidth]{curve-maxcut-barabasi_albert}
		%   \caption{Maxcut BA \label{fig:comp_embed}}
		% \end{subfigure}
		% \begin{subfigure}[t]{0.24\textwidth}
		% \centering
		%   \includegraphics[width=\textwidth]{curve-tspmult-barabasi_albert}
		%   \caption{GTSP BA \label{fig:comp_embed}}
		% \end{subfigure}
		% \begin{subfigure}[t]{0.24\textwidth}
		% \centering
		%   \includegraphics[width=\textwidth]{curve-tsp2d-clustered}
		%   \caption{TSP2D clustered \label{fig:comp_embed}}
		% \end{subfigure}
		% \caption{S2V-DQN convergence regarding held-out validation performance. }
		% 	\label{fig:app_convergence}
		% \end{figure*}
		% 
		% Figure~\ref{fig:app_convergence} shows the convergence on the four tasks with different types of graphs. The observations are consistent with other types of graphs shown in the main  paper. 
		
	\end{appendix}

\end{document}